\documentclass[11pt]{article}
\usepackage[T1]{fontenc}
\usepackage[utf8]{inputenc}
\usepackage[letterpaper,margin=1in]{geometry}
\usepackage{amsmath,amssymb,mathtools,booktabs,array,tabularx,longtable,graphicx,caption}
\usepackage{lmodern}
\usepackage{microtype}
\usepackage{natbib}
\usepackage[section]{placeins}
\usepackage[hidelinks]{hyperref}
\graphicspath{{figures/}}
\newcolumntype{L}[1]{>{\raggedright\arraybackslash}p{#1}}
\newcolumntype{R}[1]{>{\raggedleft\arraybackslash}p{#1}}
\newcommand{\papertablesize}{\fontsize{8.5}{10.5}\selectfont}
\newcommand{\reservetable}[1]{\par\ifdim\dimexpr\pagegoal-\pagetotal\relax<#1\relax\newpage\fi}

\newcommand{\notetext}[1]{\par\vspace{3pt}\begin{minipage}{\linewidth}\footnotesize\emph{Notes:} #1\end{minipage}}
\title{Estimation History and Retransformation in Machine-Learning Range-Variance Forecasting}
\author{Nicholas Appiah\thanks{Corresponding author: Nicholas Appiah, Department of Mathematics and Statistics, Texas Tech University, Lubbock, TX, USA. Email: \href{mailto:niappiah@ttu.edu}{niappiah@ttu.edu}.}
\and A. Alexandre Trindade\thanks{Department of Mathematics and Statistics, Texas Tech University, Lubbock, TX, USA.}
\and Svetlozar T. Rachev\thanks{Department of Mathematics and Statistics, Texas Tech University, Lubbock, TX, USA.}}
\date{}
\begin{document}
\maketitle
\begin{abstract}
Longer training histories can improve a machine-learning forecast without making it more accurate than a linear alternative. We examine this distinction using linear regressions and extreme gradient boosting (XGBoost) with matched predictors, forecast dates, and refitting schedules. Daily forecasts for 21 selected instruments target five-session Garman--Klass range variance over 1,715 dates from 2 January 2020 through 28 October 2026. With 18 common supplied features, expanding rather than rolling estimation produces an 85.45\% reduction in the realized XGBoost-18-minus-OLS-18 quasi-likelihood (QLIKE) gap, while linear regression retains lower loss. A three-predictor comparison gives the same ranking. Correcting log-scale forecasts with previously observed forecast errors attenuates the realized history interaction, but the change is imprecisely estimated. Moreover, 2020 accounts for 76.5\% of the signed history interaction, while smaller negative interactions remain in exploratory post-2020 comparisons. The history-specific procedure interaction is descriptive and its full-period precision is sensitive to serial dependence: tests using longer HAC bandwidths do not reject zero, and the basic 22-date block interval includes zero. The findings show why predictor matching, operational retransformation, and temporal stability should be assessed together when interpreting gains from longer estimation histories.
\end{abstract}
\section{Introduction}\label{sec:intro}

An apparent gain from a machine-learning volatility forecast can reflect its information set, training history, or conversion from a fitted log variance to a variance forecast. These choices also affect linear forecasts. A useful comparison therefore asks how much relative forecast performance changes when the same predictors and estimation histories are used, and whether the change persists when forecasts are returned to the variance scale using only information available at issuance.

The heterogeneous autoregressive (HAR) model provides a parsimonious benchmark for persistent volatility dynamics \citep{corsi2009,audrino2016}. Recent evidence shows why the comparison with more flexible methods depends on the forecasting design. \citet{chassot2026} document the importance of training-window and re-estimation choices when comparing HAR with machine-learning forecasts in a large stock panel. \citet{branco2024} find that additional predictors can improve realized-volatility forecasts without a systematic advantage for nonlinear models; \citet{christensen2023} likewise evaluate machine-learning methods against expanded linear information sets. These studies motivate a joint examination of estimation history, matched predictors, and the treatment of log-scale forecasts.

Our contribution is a matched empirical comparison for the stated five-session Garman--Klass range-variance target. We build on the training-window comparison of \citet{chassot2026} and the strong linear benchmarks of \citet{branco2024,christensen2023}. Within each matched learner pair and history, we hold the supplied features, target, forecast origins, scoring mask and refit schedule common. We then compare the change in the paired loss gap between rolling and expanding procedures under both raw exponentiation and sequential smearing, and trace when that change occurs. We use both three-feature and 18-supplied-feature comparisons to distinguish information-set changes from changes in fitted procedures. Raw-versus-smeared forecasts measure how operational retransformation changes the realized interaction, while separate inference assesses its uncertainty. These retrospective comparisons describe a selected instrument panel and a fixed forecast design; they do not establish a general advantage of machine learning or an optimal training window.

The empirical panel contains 21 selected instruments, comprising 16 exchange-traded funds and trusts and five individual stocks. Daily forecasts target the next five-session annualized mean of Garman--Klass (GK) within-session variance. The evaluation covers 1,715 common trading dates from 2 January 2020 through 28 October 2026, with 36,015 instrument-date outcomes. Models are refit monthly. Ordinary least squares (OLS) with 18 supplied features, denoted OLS-18, is compared with XGBoost-18; XGBoost-3 uses the same three predictors as HAR. A positive log-link model estimated directly under quasi-likelihood (QLIKE) loss, denoted DirectQLIKE-18, provides a comparison that avoids log-target retransformation.

Expanding estimation narrows the XGBoost-18 minus OLS-18 QLIKE difference from 0.0878 to 0.0128, an 85.45\% reduction in the realized XGBoost-18-minus-OLS-18 QLIKE gap, while the linear model retains lower loss. The three-predictor comparison gives the same ranking, and expanding OLS-18 has lower realized QLIKE than HAR. Sequential smearing attenuates the history interaction, although that change is imprecisely estimated. The interaction is also concentrated in 2020. For forecasting practice, these results show why a gain from enlarging a training sample should be interpreted alongside predictor-matched loss levels and the time profile of the gain. A separate conditional-variance exercise examines history and calibration within the generalized autoregressive conditional heteroskedasticity (GARCH) family, including its Glosten--Jagannathan--Runkle (GJR) asymmetric variant. It uses a one-session close-to-close target and is reported as auxiliary evidence.

The remainder of the paper is organized as follows. Section~\ref{sec:data} describes the data, variance measures, and predictor construction. Section~\ref{sec:models} sets out the forecasting models, estimation histories, and retransformation procedures. Section~\ref{sec:evaluation} defines the loss functions and statistical inference. Section~\ref{sec:results} presents the forecasting results, temporal and historical sensitivities, and GARCH/GJR comparison. Section~\ref{sec:conclusion} discusses the implications and limitations and concludes the paper.

\section{Data and measurement}\label{sec:data}

\subsection{Sources, sample, and descriptive evidence}

The five individual stocks are AAPL, MSFT, AMZN, JPM, and XOM. The 16 exchange-traded funds and trusts are SPY, QQQ, DIA, IWM, EFA, EEM, EWJ, EWU, FXE, FXY, FXB, UUP, GLD, SLV, USO, and DBC. These instruments provide selected equity, currency-linked, and commodity-linked exposures. Appendix~\ref{app:data-diagnostics}, Table~\ref{tab:instrument-key}, identifies the tickers. They form a selected panel rather than a probability sample of securities. We use Yahoo Finance adjusted open, high, low, and close (OHLC) prices for the range measure, adjusted-close prices for returns, and reported volume for volume features. The Yahoo extract was retrieved in the R statistical environment using the \texttt{quantmod} package; we use the adjusted OHLC and adjusted-close series after the OHLC ordering and common-factor checks in Appendix~\ref{app:uso-source-audit}. Volume is retained as supplied.\footnote{The provider snapshot was retrieved on 4 November 2026. The last scored forecast was issued on 28 October 2026 and its target ended on 4 November 2026.} Missing observations are retained as unavailable, and valid extremes remain in the panel.

Table~\ref{tab:descriptive} summarizes 36,015 adjusted-close returns and forecast targets. Returns average 0.0428\% per session with a standard deviation of 1.5839\%, skewness of $-0.985$, and excess kurtosis of 23.355. The annualized five-session target has mean 0.0327 and standard deviation 0.0913 in decimal variance units, with skewness 21.019 and excess kurtosis 710.888. The pooled moments weight every instrument-date equally. Here $N$ is the observation count, SD denotes standard deviation, and excess kurtosis is $K-3$, where $K$ is the standardized fourth central moment. The Jarque--Bera (JB) statistics and their reference $p$-values describe distributional shape; they do not account for serial or cross-instrument dependence. Appendix~\ref{app:data-diagnostics}, Tables~\ref{tab:app-returns}--\ref{tab:app-target}, reports these statistics by instrument. Figures~\ref{fig:spy-gk}--\ref{fig:spy-acf} show the SPY variance path and serial-dependence diagnostics.

\begin{table}[!htbp]\centering\papertablesize\setlength{\tabcolsep}{2pt}
\caption{Sample and distributional characteristics of the primary evaluation panel}\label{tab:descriptive}
\begin{tabular}{@{}L{.22\linewidth}R{.07\linewidth}R{.085\linewidth}R{.085\linewidth}R{.085\linewidth}R{.085\linewidth}R{.085\linewidth}R{.085\linewidth}R{.13\linewidth}@{}}\toprule
Series & $N$ & Mean & SD & Median & \shortstack[r]{5th\\percentile} & \shortstack[r]{95th\\percentile} & Skewness & \shortstack[r]{Excess\\kurtosis} \\ \midrule
Adjusted-close log return (\%) & 36,015 & 0.0428 & 1.5839 & 0.0538 & -2.2758 & 2.2847 & -0.9855 & 23.3555 \\
Five-session GK target & 36,015 & 0.0327 & 0.0913 & 0.0147 & 0.0008 & 0.1025 & 21.0189 & 710.8885 \\
\bottomrule\end{tabular}
\notetext{$N$ is the observation count and SD is standard deviation. Each series contains 21 instruments over 1,715 common dates (36,015 instrument-dates). Returns are percent per session; the Garman--Klass (GK) five-session target is annualized variance in decimal units. Percentiles pool the equally represented instrument-dates and describe the realized sample. Excess kurtosis is the standardized fourth central moment minus three. Instrument-level moments and Jarque--Bera diagnostics remain in Appendix~\ref{app:data-diagnostics}.}
\end{table}

\subsection{Range-variance outcome}

Throughout the paper, $i$ indexes instruments, $t$ and $s$ index observed trading sessions, and $j$ counts session offsets. A hat denotes a fitted or forecast quantity, a prime denotes a transpose, and $\log$ is the natural logarithm. For instrument $i$ on session $t$, let $O_{i,t}$, $H_{i,t}$, $L_{i,t}$, and $C_{i,t}$ denote adjusted open, high, low, and close. The daily Garman--Klass variance measure $v_{i,t}$ is
\begin{equation}\label{eq:gk}
v_{i,t}=\frac{1}{2}\left[\log\left(\frac{H_{i,t}}{L_{i,t}}\right)\right]^2-(2\log 2-1)\left[\log\left(\frac{C_{i,t}}{O_{i,t}}\right)\right]^2.
\end{equation}
The target $Y_{i,t}^{(5)}$ for a forecast issued after the close at $t$ is the mean of the next five observed session measures, annualized using 252 trading sessions per year:
\begin{equation}\label{eq:target}
Y_{i,t}^{(5)}=\frac{252}{5}\sum_{j=1}^{5}v_{i,t+j}.
\end{equation}
The factor $252/5$ annualizes average daily within-session variance on the observed trading calendar. The outcome is nonnegative. A cumulative variance measure would sum the five session measures without taking their average. It is neither a close-to-close return variance nor an intraday realized-variance observation. The five future sessions overlap across adjacent forecast origins, which is one reason dependence in the loss sequence matters \citep{garman1980,patton2011}.

\subsection{Information set and feature construction}

Let $P^{adj}_{i,t}$ denote the adjusted-close price and $r_{i,t}=\log(P^{adj}_{i,t}/P^{adj}_{i,t-1})$ its log return and let $V_{i,t}$ denote reported volume. All features are available at the close at $t$. For trailing window length $k$, define the range means $\bar v_{i,t}^{(k)}=k^{-1}\sum_{j=0}^{k-1}v_{i,t-j}$ for $k\in\{1,5,22\}$. The feature vector $x_{i,t}\in\mathbb R^{18}$ contains (i) $\log\bar v_{i,t}^{(1)}$, $\log\bar v_{i,t}^{(5)}$, and $\log\bar v_{i,t}^{(22)}$; (ii) five lagged log range measures $\log v_{i,t-j}$, $j=1,\ldots,5$; (iii) $r_{i,t}$, $r_{i,t-1}$, and $r_{i,t-5}$; (iv) $\min(r_{i,t},0)$ and $|r_{i,t}|$; (v) the trailing return sums $\sum_{j=0}^{k-1}r_{i,t-j}$ for $k=5,22$; and (vi) log-volume changes $\log V_{i,t}-\log V_{i,t-k}$ for $k=1,5,22$. HAR and XGBoost-3 use only the three log trailing range means. OLS-18 and XGBoost-18 use the same 18 features. No identifier, calendar-time variable, or future outcome enters $x_{i,t}$.

The log floor is fixed by instrument before forecasting. Let $\epsilon_i=10^{-6}$ times the median of the first 252 positive observed daily $v_{i,t}$ values; the annualized floor is $252\epsilon_i$. We use $\log\max(v_{i,t},\epsilon_i)$ and $\log\max(Y_{i,t}^{(5)},252\epsilon_i)$ only where a log is required, and the corresponding positive forecast floor after exponentiation. The same fixed rule applies to both estimation histories. Raw exponentiation and direct QLIKE forecasts use the positive forecast floor; the post-smearing rule is specified in Section~\ref{sec:models}.

\subsection{Historical holdings sensitivity}\label{sec:holdings-data}

The historical exercise uses Center for Research in Security Prices (CRSP) prices and the institutionally available FTSE Russell sample table \texttt{ftsesamp.ftsesamp\_russell\_us}, whose records include Top 200 flags. This entitlement-limited product is not full historical Russell 3000 membership. Its 104 observed month-end snapshots span September 2008--September 2017. Date-valid eight-character Committee on Uniform Securities Identification Procedures (CUSIP) identifiers link to 313 CRSP permanent security identification numbers (PERMNOs) from 343 observed CUSIPs. The data were accessed through Wharton Research Data Services (WRDS). The five missing snapshots are April--August 2017.

A snapshot dated in calendar month $m$ defines candidates only in month $m+1$ and expires at its end; it is not carried into later missing months. Thus there are no candidate rosters for May--September 2017. Historical publication timestamps are unavailable: availability by the first session of the following month is an assumption, not authenticated daily point-in-time membership. We impose no end-of-sample survivor restriction. The same minimum of 1,008 complete, matured training labels applies at each monthly refit; insufficient-history securities receive no forecast that month. The scored historical window is 4 January 2010--24 October 2017, with 1,860 observed evaluation dates and 180--195 eligible securities per observed date (median 187). A date-by-date issuance and scoring census accompanies the code.

The table label \emph{CRSP} denotes native observations. \emph{Check} denotes an alternative-source sensitivity that substitutes three daily GK measures on 6 May 2010 using archived Yahoo Finance prices, while retaining native returns and volume. It is not an authenticated correction. Exploratory size, liquidity and volatility strata are within-date thirds; sector labels come from the lagged roster. Appendix~\ref{app:holdings-reproduction} gives the exact membership, price-screening and source-substitution rules, and Appendix~\ref{app:additional-results} retains both versions and every reported stratum.

\section{Forecast design}\label{sec:models}

\subsection{Training histories and refit timing}

Let $\mathcal I_{i,t}$ be the eligible training origins for instrument $i$ at forecast origin $t$. An origin $s$ is eligible only when the predictors and target are observed and all five target sessions have elapsed by $t$; equivalently, $s+5\leq t$. Training origins begin in 2004. The rolling history $\mathcal I_{i,t}^{R}$ contains the most recent 1,008 eligible origins; the expanding history $\mathcal I_{i,t}^{E}$ contains all eligible origins since 2004. Both require at least 1,008 mature labels. Each model is refit at the first eligible forecast origin in a calendar month and is held fixed between refits; forecasts are issued daily. The rolling and expanding comparisons therefore change the estimation history while retaining a common target, date schedule, and forecasting code.

\subsection{Log-linear and boosted-tree forecasts}

Let $z_{i,t}=\log\max(Y_{i,t}^{(5)},252\epsilon_i)$ be the floored log target, and let $h_{i,s}$ collect the three log trailing range means. In the following regressions, $\alpha_i$ is an intercept, $\boldsymbol\beta_i$ is the coefficient vector for the stated predictors, and $u_{i,s}$ is a regression error. The log-target HAR and OLS-18 specifications are
\begin{align}
z_{i,s}&=\alpha_i+\boldsymbol\beta_i' h_{i,s}+u_{i,s}, && h_{i,s}=\big(\log\bar v_{i,s}^{(1)},\log\bar v_{i,s}^{(5)},\log\bar v_{i,s}^{(22)}\big)', \label{eq:har}\\
z_{i,s}&=\alpha_i+\boldsymbol\beta_i'x_{i,s}+u_{i,s}, && \text{OLS-18}. \label{eq:ols}
\end{align}
The coefficients are estimated separately by instrument and refit. Predictors are centered and scaled using the current training sample; a population standard deviation no larger than $10^{-12}$ is replaced by one. The three return features satisfy the exact identity
\begin{equation}\label{eq:return-redundancy}
r_{i,t}=2\min(r_{i,t},0)+|r_{i,t}|.
\end{equation}
Accordingly, OLS-18 denotes 18 supplied features, not 18 linearly independent regressors. The intercept-augmented design has 19 columns and rank 18 in every saved OLS-18 and DirectQLIKE-18 fit; HAR has rank four including its intercept. The singular value decomposition (SVD) least-squares solver discards singular values below the largest singular value multiplied by the maximum design dimension and double-precision machine accuracy. It returns the minimum-norm coefficient solution on the standardized design; individual redundant coefficients are not identified. Appendix~\ref{app:redundancy-check} reports a retrospective, exploratory rerun removing the negative-return feature from both OLS and XGBoost, retaining the same histories, masks, seed and other settings. This matters for trees because column subsampling can make a redundant supplied feature affect their fit. Model suffixes count supplied features.

XGBoost estimates an additive tree function on the same log outcome. Let $M$ be the number of boosting rounds, $T_{i,m}$ the regression tree added at round $m$, $\eta$ its learning-rate multiplier, and $\widehat z_{i,t}^{(0)}$ the initial constant prediction. The fitted log forecast is
\begin{equation}\label{eq:xgb}
\widehat z_{i,t}^{(M)}=\widehat z_{i,t}^{(0)}+\eta\sum_{m=1}^{M}T_{i,m}(x_{i,t}),\qquad \eta=0.01,
\end{equation}
where each $T_{i,m}$ is a regression tree of maximum depth three. In XGBoost version 3.4.1, the initial constant $\widehat z_{i,t}^{(0)}$ is the training mean of the log target, estimated automatically in the library's numerical precision with \texttt{boost\_from\_average=1}; it is not a fixed 0.5. The squared-error objective is used on $z$. Row sampling and column sampling \emph{by tree} are both 0.7; column sampling by level and node are one. The minimum child weight is 20, the squared-weight (L2) leaf penalty is one, the absolute-weight (L1) and split penalties are zero, and the maximum number of boosting rounds is 2,000. Central processing unit (CPU) histogram trees use 256 bins and seed 1729.

Let $v_0$ be the first origin among the most recent 200 eligible training origins. These 200 origins form chronological validation. Inner-fitting origins must satisfy $s+5<v_0$; this removes exactly five origins immediately preceding $v_0$ in every saved primary fit. A rolling fit consequently has 803 inner-fitting and 200 validation origins, with five purged origins. The latest inner label ends before validation begins. The validation criterion is root mean square error (RMSE). Fifty-round early stopping selects $M$, after which the learner is refit for $M$ rounds on the full eligible training history, including those temporarily purged validation-boundary origins. Every label in that final refit is mature at issuance. XGBoost-3 uses $h_{i,t}$ and XGBoost-18 uses $x_{i,t}$ \citep{chen2016}. Appendix~\ref{app:numerical-record} supplies the numerical configuration and fit census.

The history interaction describes two complete forecasting procedures under common rules. Expanding and rolling variants share the predictors, refit schedule, hyperparameters, 200-origin validation rule and horizon purge. Their training samples, estimated coefficients or trees, automatically estimated initial scores, selected boosting rounds and subsequent calibration errors can differ. The interaction is a paired-loss contrast across those procedures. The raw and smeared paths show the loss consequences of the operational adjustment for each procedure; they do not identify a structural or causal mechanism.

\subsection{Direct QLIKE forecast and retransformation}

DirectQLIKE-18 estimates a positive variance forecast $f_{i,s}=\exp(\alpha_i+\boldsymbol\beta_i'x_{i,s})$ directly under QLIKE loss. Here $a_{i,s}$ stacks an intercept and the training-standardized predictors, $\boldsymbol\theta$ stacks their coefficients, and $|\mathcal I_{i,t}|$ is the number of eligible training origins. Coefficients are estimated by
\begin{equation}\label{eq:direct}
\widehat{\boldsymbol\theta}_i=\arg\min_{\boldsymbol\theta}\frac{1}{|\mathcal I_{i,t}|}\sum_{s\in\mathcal I_{i,t}}\left[\frac{Y_{i,s}^{(5)}}{\exp(a_{i,s}'\boldsymbol\theta)}+a_{i,s}'\boldsymbol\theta\right],
\end{equation}
The objective is QLIKE up to terms that do not depend on the forecast. A damped Newton procedure accepts a fit only after objective and gradient checks. It initializes at the training mean target after normalization and solves on the identifiable SVD basis. Failed model-month fits are recorded and left unavailable, without fallback forecasts. Appendix~\ref{app:numerical-record} gives the exact stopping, acceptance and line-search rules; all 4,452 saved primary DirectQLIKE-18 fits passed.

The training objective uses the observed nonnegative targets without flooring either the target or the fitted conditional mean. The positive forecast floor is applied only at issuance: the reported DirectQLIKE forecast is the maximum of the exponentiated prediction and $252\epsilon_i$.

For the log-target models, raw variance forecasts are $\widehat Y^{raw}_{i,t}=\max\{252\epsilon_i,\exp(\widehat z_{i,t})\}$. Let $\mathcal H_{i,t}$ contain the available earlier forecast origins with matured outcomes, $e_{i,s}$ denote their log forecast errors, and $\widehat S_{i,t}$ denote the sequential smearing factor. The adjusted forecast $\widehat Y^{smear}_{i,t}$ is
\begin{equation}\label{eq:smear}
\widehat S_{i,t}=\frac{1}{|\mathcal H_{i,t}|}\sum_{s\in\mathcal H_{i,t}}\exp(e_{i,s}),\quad e_{i,s}=z_{i,s}-\widehat z_{i,s},\quad \widehat Y^{smear}_{i,t}=\widehat Y^{raw}_{i,t}\widehat S_{i,t},
\end{equation}
The set $\mathcal H_{i,t}$ retains at most the 504 most recent such errors; $|\mathcal H_{i,t}|$ is its size. Each error uses the log prediction generated at its own historical issuance origin, with that origin's fitted model and information set; later refits do not replace it with a fitted value. The sequences therefore combine errors from successive fitted models and are operational calibration histories, not residuals from one fixed regression. Factors update at every daily issuance, including between monthly model refits. The exact maturity rule is $s+5\leq t$ on the observed trading-session index.

The saved reconstruction begins simulated issuance on 2 January 2018. The first 126-error factor is available on 10 July 2018; before that minimum the multiplier is one. The 2018--2019 feedback warm-up contains 503 issued origins, of which 499 errors have matured by the first scored origin, 2 January 2020, for every model and history. Thus the minimum is satisfied then, although the 504-error cap is not yet filled. These are reconstructed historical forecasts, not a claim of prospective operation in 2018.

The positive floor is applied to the raw exponentiated forecast before multiplication by the factor; it is not reapplied after smearing. A factor below one could therefore produce a forecast below the raw floor, but no saved primary-evaluation forecast does so. All saved forecasts remain positive. Smearing is an operational retransformation adjustment, not a guarantee that the resulting forecast equals a conditional arithmetic mean \citep{duan1983,bollerslev2016}. Appendix~\ref{app:loss-calibration-diagnostics}, Table~\ref{app:smearing-summary}, summarizes the factors.

The two historical ratios are
\[\exp(e_{i,s})=\max(Y_{i,s},252\epsilon_i)/\exp(\widehat z_{i,s}),\qquad Y_{i,s}/\widehat Y^{raw}_{i,s}=Y_{i,s}/\max(252\epsilon_i,\exp(\widehat z_{i,s})).\]
 They coincide when both floors are inactive. For fixed positive forecasts $f_s$ and nonnegative outcomes $Y_s$, write $A=n^{-1}\sum_s Y_s/f_s$. The average historical QLIKE of $cf_s$ has derivative $-A/c^2+1/c$ and, for $A>0$, is minimized over $c>0$ at $A$. If $A=0$ there is no positive interior minimizer. The historical ratio optimum places no restriction on prospective performance. Appendix~\ref{app:historical-calibration} reports the training, warm-up and evaluation floor census.

\subsection{Fixed-rule forecast on the same range target}

For a parsimonious same-target comparison, an exponentially weighted moving average (EWMA) uses the daily Garman--Klass measure already available at issuance. Let $w_{i,t}$ be its smoothed variance state and $\lambda=0.94$ the fixed decay parameter:
\begin{equation}\label{eq:ewma}
w_{i,t}=\lambda w_{i,t-1}+(1-\lambda)v_{i,t},\qquad \widehat Y^{EWMA}_{i,t}=252w_{i,t}.
\end{equation}
The state is initialized at the mean of the first 252 finite, nonnegative daily measures, before the evaluation period. If an input is unavailable, the preceding state is retained. The rule projects the current state across the next five sessions, matching the main target's annualization and within-session measurement. The decay parameter is fixed without loss-based selection. This additional fixed-rule comparison is descriptive and is outside the original principal test family.

\subsection{Conditional-variance comparison}\label{sec:garch-design}

We use Student-$t$ GARCH(1,1) and GJR-GARCH(1,1) forecasts for the same 21 instruments to examine whether history and an additional past-only calibration also affect models that forecast conditional variance directly in levels \citep{bollerslev1986,glosten1993,hansen2005}. Their target is the annualized one-session squared adjusted-close return, which includes overnight variation. This exercise compares histories within the conditional-variance family and uses a different target and refit schedule from the primary range-variance comparison. The rolling history contains 1,008 returns; expansion begins in 2004 or at the instrument's first available observation thereafter. Parameters are re-estimated every 22 sessions, with daily variance updates between refits. Appendix~\ref{app:garch-details} gives the likelihood and fitting details.

Let $R_{i,t}=100r_{i,t}$ be the adjusted-close log return in percentage points, where $r_{i,t}$ is the decimal return defined in Section~\ref{sec:data}. In these auxiliary models, $\mu_i$ is the conditional return mean, $\varepsilon_{i,t}$ is the return innovation, $q_{i,t}$ is its conditional variance, and $\eta_{i,t}$ is a standardized Student-$t$ innovation. With $\nu_i>2$ degrees of freedom,
\begin{equation}\label{eq:garch-mean}
R_{i,t}=\mu_i+\varepsilon_{i,t},\qquad \varepsilon_{i,t}=\sqrt{q_{i,t}}\eta_{i,t},\qquad \eta_{i,t}\sim \sqrt{\frac{\nu_i-2}{\nu_i}}\,t_{\nu_i}.
\end{equation}
Here $t_{\nu_i}$ denotes the Student-$t$ distribution; the scaling gives $\eta_{i,t}$ unit variance. For GARCH(1,1), $\omega_i$ is the variance intercept, $\alpha_i$ weights the most recent squared innovation, and $\beta_i$ weights the preceding conditional variance:
\begin{equation}\label{eq:garch}
q_{i,t}=\omega_i+\alpha_i\varepsilon_{i,t-1}^{2}+\beta_i q_{i,t-1},\qquad \omega_i>0,\quad \alpha_i,\beta_i\geq0,\quad \alpha_i+\beta_i<1.
\end{equation}
The GJR-GARCH(1,1) model adds coefficient $\gamma_i$ for a negative innovation. The indicator $\mathbf{1}\{\cdot\}$ equals one when its condition holds and zero otherwise:
\begin{align}\label{eq:gjr}
q_{i,t}&=\omega_i+\alpha_i\varepsilon_{i,t-1}^{2}+\gamma_i\mathbf{1}\{\varepsilon_{i,t-1}<0\}\varepsilon_{i,t-1}^{2}+\beta_i q_{i,t-1},\\
\gamma_i&\geq0,\qquad \alpha_i+\beta_i+\frac{\gamma_i}{2}<1.
\end{align}

The one-session close-to-close (CC) target $Y^{CC}_{i,t+1}$ is an annualized squared return in decimal units. Hats denote the estimated conditional mean and next-session variance available after the close at $t$. The direct conditional-second-moment forecast is
\begin{equation}\label{eq:garch-target}
Y^{CC}_{i,t+1}=252\left(\frac{R_{i,t+1}}{100}\right)^2,\qquad
\widehat Y^{CC}_{i,t+1}=\frac{252}{10{,}000}\left(\widehat q_{i,t+1}+\widehat\mu_i^{\,2}\right).
\end{equation}
The squared conditional mean is included in the forecast because the target is a second moment. An additional calibration sensitivity multiplies this direct forecast by the mean of up to 504 previously matured outcome-to-forecast ratios, computed separately for each model, instrument, and history. At least 126 ratios are required; ratio-adjusted forecasts are unavailable before that threshold. This is a sequential ratio adjustment of a level forecast, distinct from log-target smearing. All eight model-history-treatment combinations are evaluated on their common complete-case dates.

\section{Evaluation and inference}\label{sec:evaluation}

For nonnegative outcome $Y$ and positive forecast $f$, the primary loss $L_{QLIKE}(Y,f)$ is
\begin{equation}\label{eq:qlike}
L_{QLIKE}(Y,f)=\frac{Y}{f}+\log f,
\end{equation}
which is well defined at $Y=0$. This definition omits outcome-only normalization terms and can produce negative absolute loss levels, as in Table~\ref{tab:absolute-loss}. Omitting those terms does not change paired differences on a common target. We evaluate the observed GK measure; the use of QLIKE alone does not establish proxy-robust comparisons for latent total return variance. We also report squared error $(Y-f)^2$ and its sample mean, the mean squared error (MSE), in squared annualized decimal-variance units. QLIKE is the documented primary loss: its outcome term measures variance relative to the forecast, whereas squared error evaluates absolute differences in variance units. Loss rankings with imperfect volatility proxies require care \citep{patton2011}; we interpret both scores for the stated GK target and do not assume that all proxy-noise conditions hold automatically. Forecast comparisons are paired on common dates and targets. For models $A$ and $B$, $L$ denotes the chosen evaluation loss and $d_{i,t}=L(Y_{i,t},\widehat Y^A_{i,t})-L(Y_{i,t},\widehat Y^B_{i,t})$ is their paired loss difference. Let $N_t$ be the number of paired instrument forecasts on date $t$ and $T$ the number of evaluation dates. The date-level mean is $D_t=N_t^{-1}\sum_{i=1}^{N_t}d_{i,t}$; the reported period mean is $\overline D=T^{-1}\sum_{t=1}^{T}D_t$. Each date receives equal weight. This equals instrument-date weighting for the balanced primary panel of 21 forecasts on each of 1,715 dates. The two weightings can differ when holdings rosters change.

For learner $A$ relative to comparator $B$, the history interaction $\widehat\Delta_{A,B}$ is the expanding-history mean difference (superscript $E$) minus the rolling-history mean difference (superscript $R$):
\begin{equation}\label{eq:interaction}
\widehat\Delta_{A,B}=\overline D^{E}_{A,B}-\overline D^{R}_{A,B}=\frac{1}{T}\sum_{t=1}^{T}\frac{1}{N_t}\sum_{i=1}^{N_t}\bigl(d^{E}_{i,t}-d^{R}_{i,t}\bigr).
\end{equation}
A negative value means that the realized learner-minus-comparator loss difference is smaller under expansion. Its sign does not identify the lower-loss learner within either history. The contrast compares complete history-specific procedures, including their fitted coefficients or trees, selected boosting rounds and subsequent calibration ledgers.

The point estimate is a date-first average for this finite panel and period. A repeated-sampling interpretation treats the saved paired loss series, for the documented design, as a realization whose variation is approximated under temporal stability, sufficiently weak dependence (including summable autocovariances for HAC), and suitable finite moments, for example a finite $2+\delta$ moment for some $\delta>0$. These are substantive working assumptions given the 2020 concentration and growing expanding-window histories. The saved losses reflect the realized consequences of estimation. The resampling and long-run variance calculations use those losses; they do not regenerate training data, early-stopping choices, fitted parameters or sequential calibration, and do not establish unconditional validity for the recursive algorithm \citep{diebold1995,west1996}. 

We preserve the ten QLIKE and ten squared-error contrasts. The original Bartlett HAC(4) test and its within-loss ten-test Holm adjustment are retained, with each original family adjusted again separately at lags 10, 21 and 43 \citep{newey1987,holm1979}. Pointwise moving-block bootstrap (MBB) intervals use synchronized blocks of lengths 5, 22 and 44 trading dates \citep{kunsch1989}.

The auxiliary GARCH exercise applies date-first HAC and synchronized circular MBB-22 to common complete-case forecasts across all 21 instruments. Original Holm families for history, asymmetry, ratio adjustment, and history-by-asymmetry are preserved at each of the four bandwidths. Appendix~\ref{app:garch-details}, Table~\ref{app:garch-contrasts}, displays the results.

Four lags are retained as the documented initial short-bandwidth specification. Five-session target overlap motivates accounting for dependence through four adjacent origins; it does not imply that loss dependence ends there. The sample interaction autocorrelations at lags one and five are 0.9650 and 0.5561. Monthly refitting, overlapping training histories and sequential calibration can contribute to this persistence. Appendix~\ref{app:interaction-diagnostics} reports the interaction's autocorrelation, cumulative path and resampling diagnostics. HAC lags 10, 21 and 43 and block lengths 5, 22 and 44 remain sensitivity checks; we do not replace the original four-lag tests after inspecting their results.

Yahoo and historical-holdings intervals use 5,000 non-circular moving-block draws with NumPy's PCG64 generator and seed 1729. On the $T$-session calendar, each full block starts uniformly on $0,\ldots,T-b$, inclusive. The remainder $r=T-b\lfloor T/b\rfloor$, when positive, starts independently and uniformly on $0,\ldots,T-r$, inclusive. All instruments share sampled dates. In the holdings exercise the 1,967-date calendar contains 1,860 observed dates and 107 roster gaps. The mask travels with the losses. HAC uses observed-date centering, zero centered residuals at gaps and observed-date normalization of lag products and the mean variance. Bootstrap draws divide by their number of observed eligible dates and drop zero-denominator draws. This design does not presume that roster gaps are random. The percentile bounds use NumPy's linear quantiles at 0.025 and 0.975 of raw resampled means, without studentization, null centering or multiplicity adjustment. The separate GARCH exercise uses circular 22-session blocks with wraparound and truncation to $T$, with seeds 4122026 for the paired tables and 4122027 for the figure. Appendix~\ref{app:interaction-diagnostics} reports same-draw basic intervals.

HAC tests use a symmetric asymptotic normal approximation, whereas these percentile intervals reflect the possibly asymmetric resampled distribution and do not studentize it. They also use different smoothing scales; Holm correction applies to the stated tests, not to the percentile intervals. Agreement is therefore not automatic. We report the disagreement as uncertainty about precision, rather than allowing the favorable interval to override dependence-sensitive tests.

Inferential statements use unrounded $p$-values. Values below 0.0001 are shown as $<0.0001$ where indicated. If a tabulated value rounds to 0.0000, the display does not imply an exact zero. The primary designation refers to the internal retrospective reconstruction specification described in Appendix~\ref{app:analysis-chronology}, not preregistration or an untouched confirmation sample.

\section{Results}\label{sec:results}

\subsection{Distributional shape and persistence}

The instrument-level diagnostics in Appendix~\ref{app:data-diagnostics} show substantial scale differences across the panel and strong right skew in every range target. Return skewness varies in sign, with particularly large dispersion for SLV and USO. For SPY, Figures~\ref{fig:spy-gk}--\ref{fig:spy-acf} contrast the weak persistence of raw returns with the stronger persistence of range variance. The autocorrelation function (ACF) bands in the figure are descriptive white-noise reference bands; forecast-loss inference uses the date-level procedures in Section~\ref{sec:evaluation}.

\subsection{Absolute accuracy and loss-function dependence}

Table~\ref{tab:absolute-loss} reports absolute losses for all five principal procedures under both histories and for the same-target fixed EWMA rule. All rows use identical instrument-dates. Expanding OLS-18 has lower QLIKE than expanding XGBoost-18, whereas the MSE point ordering is reversed (10.1323 versus 5.3851 using the common $10^{-3}$ MSE scale in Table~\ref{tab:absolute-loss}). The XGBoost-18 minus OLS-18 MSE difference is $-0.0047$ and does not pass the declared MSE family's Holm adjustment ($p=1.0000$; Appendix~\ref{app:additional-results}, Table~\ref{app:mse-contrasts}). A ranking under the primary loss therefore does not establish a loss-independent forecasting advantage. The EWMA row supplies a simple five-session range-target reference; the separate GARCH results concern a different close-to-close target.

\reservetable{320pt}
\begingroup\papertablesize\setlength{\tabcolsep}{2pt}
\begin{longtable}{@{}L{.15\linewidth}L{.42\linewidth}R{.20\linewidth}R{.18\linewidth}@{}}
\caption{Absolute loss levels on the common five-session range-variance target}\label{tab:absolute-loss}\\
\toprule History & Forecast procedure & Mean QLIKE & \shortstack[r]{Mean MSE\\($\times10^{-3}$)} \\ \midrule
\endfirsthead
\multicolumn{4}{c}{Table \thetable{} -- continued}\\
\toprule History & Forecast procedure & Mean QLIKE & \shortstack[r]{Mean MSE\\($\times10^{-3}$)} \\ \midrule\endhead
\midrule\multicolumn{4}{r}{Continued on next page}\\\endfoot
\bottomrule\endlastfoot
\multicolumn{4}{@{}l}{\textit{2020--2026: 1715 common dates, 21 instruments}}\\\addlinespace
Rolling & HAR (smeared) & -3.1784 & 5.9116 \\
Rolling & OLS-18 (smeared) & -3.1967 & 10.4439 \\
Rolling & XGBoost-3 (smeared) & -3.0884 & 9.6397 \\
Rolling & XGBoost-18 (smeared) & -3.1089 & 7.0931 \\
Rolling & DirectQLIKE-18 & -3.1969 & 9.3263 \\
Expanding & HAR (smeared) & -3.1906 & 5.4726 \\
Expanding & OLS-18 (smeared) & -3.2065 & 10.1323 \\
Expanding & XGBoost-3 (smeared) & -3.1764 & 5.9857 \\
Expanding & XGBoost-18 (smeared) & -3.1937 & 5.3851 \\
Expanding & DirectQLIKE-18 & -3.2070 & 11.5492 \\
Fixed rule & EWMA-GK & -3.1540 & 6.1264 \\
\end{longtable}
\noindent\footnotesize QLIKE denotes quasi-likelihood loss and MSE denotes mean squared error; lower values are better. Losses are averaged across the same 21 instruments within each date, then equally across dates. Multiply the MSE entries by $10^{-3}$ to recover squared annualized-variance units. HAR denotes heterogeneous autoregression; OLS denotes ordinary least squares; XGBoost denotes extreme gradient boosting; suffixes give predictor counts. Log-target models use sequential smearing. DirectQLIKE-18 estimates a positive log-link forecast by minimizing variance-scale QLIKE. EWMA-GK denotes the fixed-rule exponentially weighted moving average of daily Garman--Klass variance, projected to the same five-session target. These loss levels are descriptive; paired inference is reported separately.\normalsize
\endgroup

The reversal is concentrated rather than uniform. USO contributes 95.35\% of the signed expanding XGBoost-18-minus-OLS-18 MSE difference; the forecast origin 21 April 2020 contributes 66.25\%. These shares can be offset by other observations. Large absolute OLS forecast errors on a few dates receive much more weight under squared error than under QLIKE. Appendix~\ref{app:loss-calibration-diagnostics}, Tables~\ref{app:loss-concentration}--\ref{app:smearing-summary}, reports instrument/date contributions and smearing-factor dispersion. All valid observations remain in the primary results; these diagnostics do not establish a significant MSE advantage, whose Holm-adjusted $p$-value remains 1.0000.

Appendix~\ref{app:uso-source-audit} traces the five future sessions and the relevant forecasts for each of the five influential April USO origins. Those within-provider checks leave all five observations in the primary panel.

\subsection{Matched predictors reveal what history changes}

\begin{figure}[tbp]\centering
\includegraphics[width=0.98\linewidth]{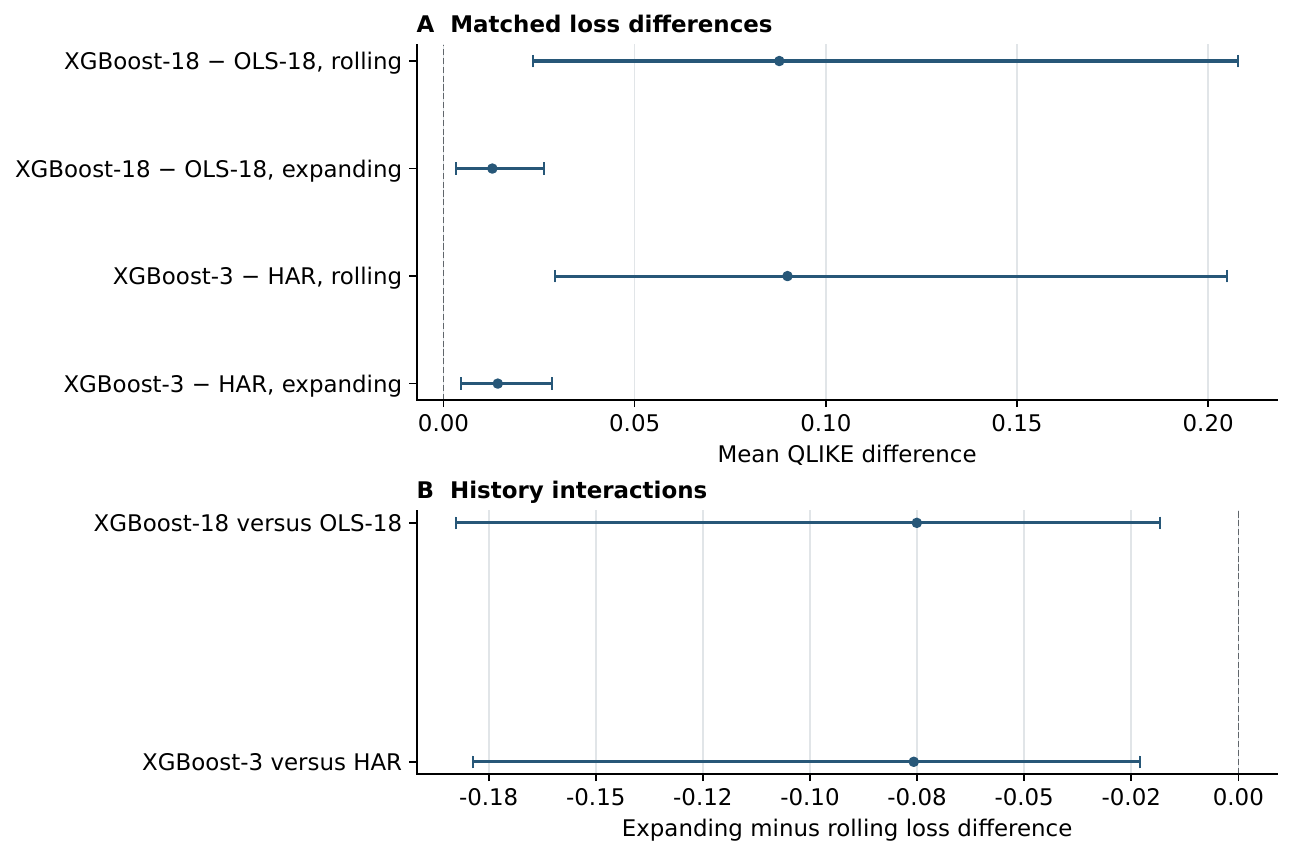}
\caption{Matched-predictor QLIKE differences and history interactions. Dots are date-first means; positive model-minus-comparator differences favor the comparator. Bars are pointwise 95\% percentile intervals using synchronized 22-date moving blocks of saved losses. They are unadjusted for multiplicity and do not refit models. Table~\ref{app:hac-sensitivity} reports the bandwidth and Holm sensitivities. For the XGBoost-18 interaction, the same-draw basic interval includes zero (Table~\ref{app:bootstrap-construction}).}\label{fig:matched}
\end{figure}

Table~\ref{tab:primary} and Figure~\ref{fig:matched} compare history-specific procedures. With 18 common supplied features, the date-first XGBoost-18-minus-OLS-18 QLIKE gap is 0.0878 under rolling estimation and 0.0128 under expansion. The estimated interaction is $-0.0751$, an 85.45\% reduction in the positive realized gap. OLS-18 retains lower realized QLIKE under both histories. The Holm-adjusted interaction $p$-values are 0.0497, 0.2481, 0.4475 and 0.5090 at HAC(4), (10), (21) and (43), respectively. The corresponding standard errors are 0.0300, 0.0395, 0.0460 and 0.0491. Table~\ref{app:hac-sensitivity} gives the raw $p$-values. The pointwise percentile MBB-22 interval $[-0.1827,-0.0183]$ excludes zero, and the same-draw basic interval $[-0.1318,0.0325]$ includes it. The MBB-44 basic interval also includes zero, $[-0.1341,0.0402]$.

In the matched three-log-feature comparison, XGBoost-3 has higher realized QLIKE than HAR under rolling and expanding estimation, with gaps 0.0900 and 0.0142. Its interaction is $-0.0758$ and its Holm-adjusted HAC(4) $p$-value is 0.0496; the wider tests do not reject at 5\%. Expanding OLS-18 has lower realized QLIKE than HAR by 0.0159. Its adjusted $p$-values include 0.0418 at HAC(21) and 0.0669 at HAC(43). Expanding XGBoost-18 versus HAR and expanding DirectQLIKE-18 versus OLS-18 do not reject equal mean loss under the documented ten-test HAC(4) calculations. Such non-rejection does not establish equivalence. The exposure and instrument evidence is exploratory (Tables~\ref{app:exposure-groups}--\ref{app:asset-level-interactions}).

No member of the primary QLIKE family rejects at 5\% after the HAC(43) adjustment. We emphasize the realized interactions and their temporal concentration because the working dependence and resampling approximations differ. See Tables~\ref{app:hac-sensitivity}--\ref{app:block-sensitivity} and \ref{app:bootstrap-construction}.

\begin{table}[tbp]\centering\papertablesize\setlength{\tabcolsep}{3pt}
\caption{Primary date-first QLIKE comparisons, 2020--2026}\label{tab:primary}
\begin{tabular}{@{}L{.43\linewidth}R{.10\linewidth}R{.10\linewidth}R{.115\linewidth}L{.195\linewidth}@{}}\toprule
Contrast (first minus second) & Mean & \multicolumn{2}{c}{$p$-value} & \shortstack[l]{MBB-22 95\%\\confidence interval} \\
\cmidrule(lr){3-4}
 & & HAC(4) & Holm-adjusted & \\\midrule
\multicolumn{5}{@{}l}{\textit{Panel A: 18 matched predictors}}\\\addlinespace
Rolling XGBoost-18 $-$ OLS-18 & 0.0878 & 0.0068 & 0.0476 & [0.0235, 0.2079] \\
Expanding XGBoost-18 $-$ OLS-18 & 0.0128 & 0.0083 & 0.0496 & [0.0032, 0.0264] \\
XGBoost-18 history interaction & -0.0751 & 0.0124 & 0.0497 & [-0.1827, -0.0183] \\
\multicolumn{5}{@{}l}{\textit{Panel B: three matched predictors}}\\\addlinespace
Rolling XGBoost-3 $-$ HAR & 0.0900 & 0.0045 & 0.0361 & [0.0292, 0.2050] \\
Expanding XGBoost-3 $-$ HAR & 0.0142 & 0.0011 & 0.0098 & [0.0047, 0.0285] \\
XGBoost-3 history interaction & -0.0758 & 0.0093 & 0.0496 & [-0.1787, -0.0230] \\
\multicolumn{5}{@{}l}{\textit{Panel C: sequential retransformation}}\\\addlinespace
Expanding XGBoost-18 smeared $-$ raw & -0.0088 & 0.0284 & 0.0851 & [-0.0187, 0.0005] \\
\multicolumn{5}{@{}l}{\textit{Panel D: information and direct-loss comparisons}}\\\addlinespace
Expanding XGBoost-18 $-$ HAR & -0.0031 & 0.3049 & 0.6099 & [-0.0084, 0.0028] \\
Expanding OLS-18 $-$ HAR & -0.0159 & 0.0002 & 0.0020 & [-0.0285, -0.0070] \\
Expanding DirectQLIKE-18 $-$ OLS-18 & -0.0005 & 0.8365 & 0.8365 & [-0.0071, 0.0047] \\
\bottomrule\end{tabular}
\notetext{QLIKE is quasi-likelihood loss. HAC(4) denotes heteroskedasticity and autocorrelation consistent inference with four lags; MBB-22 denotes a moving-block bootstrap with synchronized blocks of 22 trading dates. OLS-18 is ordinary least squares with 18 supplied features; XGBoost-18 and XGBoost-3 use 18 and three predictors; HAR denotes heterogeneous autoregression; DirectQLIKE-18 estimates a positive log-link forecast by minimizing variance-scale QLIKE. Every date is equally weighted after averaging paired losses over the 21 instruments. The panels organize the presentation; Holm adjustment remains within the single family of ten documented QLIKE contrasts.  HAC(4) $p$-values are unadjusted; the Holm column reports the ten-test adjustment. The MBB-22 percentile intervals use 5,000 synchronized 22-date moving-block draws and are pointwise and unadjusted. A positive model-minus-comparator difference favors the comparator.}
\end{table}

The targeted redundancy check removes the negative-return feature from both procedures without changing the available linear span. The 17-feature QLIKE gaps are 0.0859 under rolling and 0.0121 under expanding estimation; the history interaction is -0.0738. The matched linear model retains the lower point QLIKE under both histories. Appendix~\ref{app:redundancy-check} reports absolute losses, both loss functions and the separate exploratory tests. Similar point values are not an equivalence test.

\begin{table}[tbp]\centering\papertablesize\setlength{\tabcolsep}{4pt}
\caption{Factorial QLIKE decomposition by estimation history and retransformation}\label{tab:factorial-qlike}
\begin{tabular}{@{}llrrr@{}}\toprule
Estimation history & Forecast treatment & OLS-18 & XGBoost-18 & XGBoost-18 $-$ OLS-18 \\
\midrule
Rolling & Raw exponentiation & -3.1759 & -3.0712 & 0.1047 \\
Rolling & Sequentially smeared & -3.1967 & -3.1089 & 0.0878 \\
Expanding & Raw exponentiation & -3.1962 & -3.1849 & 0.0113 \\
Expanding & Sequentially smeared & -3.2065 & -3.1937 & 0.0128 \\
\bottomrule\end{tabular}
\notetext{QLIKE levels are date-first means over the 21-instrument matched panel and are negative under the stated convention. The matched gap is XGBoost-18 minus OLS-18, so positive values favor OLS-18. All cells use the same 1,715 evaluation dates and paired observations.}
\end{table}

The four history-by-treatment rows show lower realized QLIKE under expansion for both procedures. The matched XGBoost-18-minus-OLS-18 gap falls from 0.1047 to 0.0113 under raw exponentiation and from 0.0878 to 0.0128 after smearing. The corresponding interactions are $-0.0934$ and $-0.0751$. Smearing shifts the realized interaction by $+0.0183$, or 19.6\% of its raw magnitude. The common rule uses a distinct ledger of past-issued errors for each fitted procedure.

\subsection{Retransformation and the timing of forecast errors}

The exploratory direct test of the $+0.0183$ change in interaction has unadjusted HAC $p$-values of 0.2100, 0.3599, 0.4483 and 0.4874 at lags 4, 10, 21 and 43. Its pointwise MBB-22 percentile interval $[-0.0137,0.0769]$ and basic interval $[-0.0403,0.0503]$ both include zero. These approximations leave substantial uncertainty about the incremental effect of smearing. Separately, smearing reduces expanding XGBoost-18's realized QLIKE by 0.0088 relative to raw exponentiation; its Holm-adjusted HAC(4) $p$-value is 0.0851 and its pointwise MBB-22 interval includes zero.

\subsection{Temporal concentration}

\begin{figure}[tbp]\centering
\includegraphics[width=0.98\linewidth]{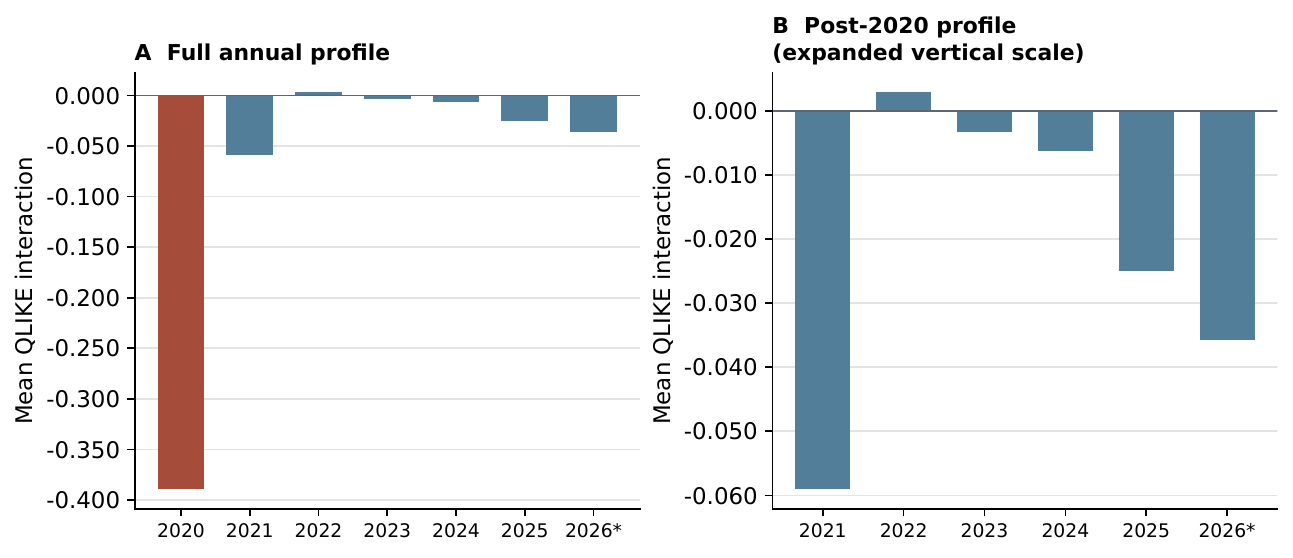}
\caption{Annual date-level means of the extreme gradient boosting (XGBoost) quasi-likelihood (QLIKE) history interaction after sequential smearing. Panel A shows all evaluation years; Panel B expands the vertical scale for 2021--2026. The panels use different vertical scales. OLS denotes ordinary least squares; the suffix 18 indicates 18 common supplied features. The interaction is (expanding XGBoost-18 minus OLS-18) minus (rolling XGBoost-18 minus OLS-18). Annual means are descriptive; 2026 ends on 28 October.}\label{fig:temporal}
\end{figure}

Figure~\ref{fig:temporal} and Table~\ref{tab:temporal} show that the full-period interaction is not a stable average across years. In the specified 2020 block it is $-0.3894$ over 253 dates and contributes 76.5\% of the full signed sum. The grouped means for 2021--2022 and 2023--2026 are $-0.0282$ and $-0.0167$. Those later values are much smaller in magnitude. This decomposition is descriptive, not a sequence of independently confirmatory tests; 2026 is observed only through 28 October. Appendix~\ref{app:crisis-diagnostics}, Tables~\ref{tab:crisis-forecasts}--\ref{tab:crisis-fit-support}, shows selected issued forecasts and their saved refit diagnostics for the largest March 2020 interaction dates. 

\begin{table}[tbp]\centering\papertablesize\setlength{\tabcolsep}{3pt}
\caption{Specified temporal profile of the XGBoost-18 QLIKE history interaction}\label{tab:temporal}
\begin{tabular}{@{}L{.20\linewidth}R{.13\linewidth}R{.32\linewidth}R{.29\linewidth}@{}}\toprule
Period & Dates & \shortstack[r]{Mean date-level\\interaction} & \shortstack[r]{Share of\\signed total} \\\midrule
2020 & 253 & $-0.3894$ & 76.5\% \\
2021--2022 & 503 & $-0.0282$ & 11.0\% \\
2023--2026 & 959 & $-0.0167$ & 12.5\% \\
Full period & 1,715 & $-0.0751$ & 100.0\% \\
\bottomrule\end{tabular}
\notetext{Shares are period sums of date-level interactions divided by the full-period sum. They describe contributions to this realized sample and are not period-specific significance tests.}
\end{table}

An exploratory omission analysis holds the forecast paths fixed and evaluates the periods after 2020. The XGBoost-18 history interactions are $-0.0207$ in 2021--2026 and $-0.0167$ in 2023--2026, with negative MBB-22 and MBB-44 intervals; the matched three-predictor interactions are also negative. These smaller effects show that the signed interaction is not confined to 2020. Appendix~\ref{app:additional-results}, Tables~\ref{app:post2020-levels}--\ref{app:post2020-interactions}, reports every procedure's loss levels and the separate exploratory inference. The four window-by-learner interactions use their own Holm adjustment at each HAC lag. These post-hoc comparisons do not change the original principal test family, and 2020 remains in all primary results.

As exploratory checks on concentration, the median of the 21 instrument mean XGBoost-18 history interactions is $-0.0705$, and dropping one instrument at a time yields mean interactions from $-0.0785$ to $-0.0686$. Each omission recomputes the date-first mean over the remaining instruments. These summaries are descriptive; the full-panel mean remains the primary estimand.

\subsection{Historical and holdings sensitivities}

A separate 2014--2017 Yahoo reconstruction yields a same-sign XGBoost-18 interaction of $-0.0084$ over 1,007 dates, without rejection under its ten-test Holm adjustment ($p=0.1080$). Appendix~\ref{app:additional-results}, Tables~\ref{app:temporal-2014}--\ref{app:historical-primary}, reports this historical sensitivity.

The holdings sensitivity uses the restricted sample described in Section~\ref{sec:holdings-data}.\footnote{Accessed November 6, 2026.} Native-CRSP rolling XGBoost-18 minus OLS-18 is $-0.1010$, compared with $+0.0107$ under the alternative-source Check; its history interaction changes from $+0.0093$ to $-0.0032$. We therefore treat this exercise as evidence about source sensitivity, rather than corroboration of the main ranking. Appendix~\ref{app:holdings-reproduction}, Tables~\ref{app:source-bars}--\ref{app:source-exclusions}, identifies the three bars, their propagation through the full forecasting pipeline, and exploratory evaluation exclusions. The original views and all strata remain in Tables~\ref{app:crsp-overview}--\ref{app:crsp-strata}. With fits held fixed, deleting the direct evaluation footprint changes the native rolling gap to $+0.01124$ and deleting the three securities changes it to $+0.01075$. These exploratory deletions change the evaluation sample. The Check rerun instead changes training, feature and calibration inputs wherever the substituted bars propagate. A training bar can affect forecasts after its direct evaluation footprint ends.

\subsection{GARCH/GJR results on the close-to-close target}\label{sec:garch-results}

This comparison evaluates history and calibration within conditional-variance models on the one-session close-to-close squared-return target. Its loss levels are not directly comparable with the five-session Garman--Klass loss levels in Table~\ref{tab:absolute-loss}. The same-target fixed-rule benchmark for the primary range-variance exercise is the EWMA-GK forecast in that table.

Table~\ref{tab:garch-performance} reports all eight forecast variants on 1,715 common dates and 21 instruments. Expanding estimation lowers mean QLIKE relative to rolling estimation by 0.0093 for direct GARCH and 0.0101 for direct GJR-GARCH; the corresponding reductions for ratio-adjusted forecasts are 0.0090 and 0.0101. All four comparisons have Holm-adjusted HAC(4) $p=0.0464$ in the four-test history family. None rejects at 5\% at HAC(10), (21) or (43) (Table~\ref{app:garch-contrasts}). Mean squared error (MSE), however, is higher under expansion in each pair. The history ranking therefore depends on the evaluation loss.

\begin{figure}[tbp]\centering
\includegraphics[width=0.97\linewidth]{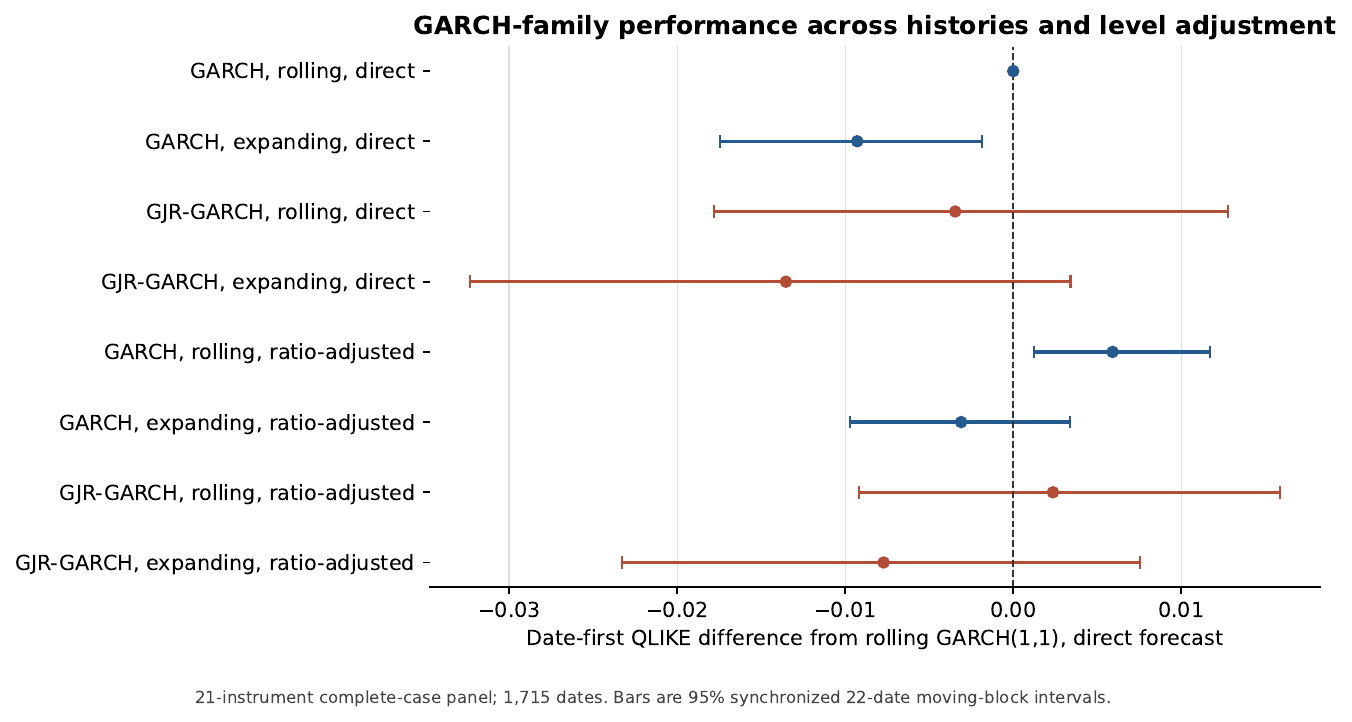}
\caption{One-session GARCH-family quasi-likelihood (QLIKE) loss differences across estimation histories and past-only ratio adjustment. GARCH denotes generalized autoregressive conditional heteroskedasticity; GJR is its Glosten--Jagannathan--Runkle asymmetric variant. Estimates are relative to direct rolling GARCH(1,1). Bars show pointwise 95\% percentile moving-block bootstrap intervals using synchronized circular blocks of 22 dates, unadjusted for multiplicity. The baseline point is zero by definition and has no estimated interval. Table~\ref{app:garch-contrasts} reports sensitivity to the HAC bandwidth and within-family adjustment. All procedures use the same complete-case sample of 21 instruments and one-session close-to-close squared returns.}\label{fig:garch-comparison}
\end{figure}

The negative GJR-minus-GARCH QLIKE point differences range from $-0.0046$ to $-0.0035$, but none passes the asymmetry family's Holm-adjusted HAC(4) test. The ratio adjustment increases QLIKE by 0.0058 to 0.0062 across model-history pairs. Two comparisons pass the ratio family's Holm-adjusted HAC(4) test, and three pointwise 22-date intervals exclude zero. Thus this additional past-only correction worsens point QLIKE in every pair, with uncertainty that differs by comparison. Appendix~\ref{app:garch-details}, Table~\ref{app:garch-contrasts}, retains the full paired estimates and inference, and Figure~\ref{fig:garch-comparison} summarizes differences from direct rolling GARCH.

\begin{table}[!htbp]
\centering
\papertablesize
\setlength{\tabcolsep}{3pt}
\caption{GARCH and GJR-GARCH one-session close-to-close forecast performance}
\label{tab:garch-performance}
\begin{tabular}{@{}L{.12\linewidth}L{.22\linewidth}L{.18\linewidth}R{.13\linewidth}R{.14\linewidth}R{.14\linewidth}@{}}
\toprule
Model & History & Forecast treatment & \mbox{$N$} & \shortstack[r]{Mean\\QLIKE} & \shortstack[r]{Mean\\MSE} \\ \midrule
\multicolumn{6}{@{}l}{\textit{Panel A: Direct conditional-second-moment forecast}} \\
GARCH & Rolling (1,008) & Direct & 36,015 & -2.3911 & 0.0888 \\
GARCH & Expanding & Direct & 36,015 & -2.4004 & 0.0897 \\
GJR & Rolling (1,008) & Direct & 36,015 & -2.3945 & 0.0893 \\
GJR & Expanding & Direct & 36,015 & -2.4046 & 0.0900 \\
\addlinespace
\multicolumn{6}{@{}l}{\textit{Panel B: Past-only ratio-adjusted forecast}} \\
GARCH & Rolling (1,008) & Ratio-adjusted & 36,015 & -2.3852 & 0.0899 \\
GARCH & Expanding & Ratio-adjusted & 36,015 & -2.3942 & 0.0914 \\
GJR & Rolling (1,008) & Ratio-adjusted & 36,015 & -2.3887 & 0.0912 \\
GJR & Expanding & Ratio-adjusted & 36,015 & -2.3988 & 0.0922 \\
\bottomrule
\end{tabular}
\par\vspace{2pt}\begin{minipage}{0.98\linewidth}\footnotesize\emph{Notes:} \mbox{$N$} counts instrument-date observations. QLIKE is quasi-likelihood loss; MSE is mean squared error. GARCH denotes generalized autoregressive conditional heteroskedasticity; GJR denotes its Glosten--Jagannathan--Runkle variant. All rows use the same complete-case sample of 1,715 dates and 21 instruments (36,015 instrument-dates). The target is the annualized one-session adjusted-close squared return. The direct forecast includes both conditional return variance and the squared conditional mean. QLIKE is reported as \mbox{$y/f+\log f$}; lower values are better. Ratio-adjustment uses the mean of up to 504 previously matured \mbox{$y/f$} ratios, with a minimum of 126, and is an added past-only calibration sensitivity. Standard GARCH forecasts conditional variance directly and does not require log-scale retransformation.\end{minipage}
\end{table}

\FloatBarrier

\FloatBarrier

\section{Discussion and conclusion}\label{sec:conclusion}

In the selected 21-instrument panel, the expanding history-specific procedure reduces the realized XGBoost-18-minus-OLS-18 QLIKE gap from 0.0878 to 0.0128, about 85\%. OLS-18 retains lower realized QLIKE under both histories. The interaction is $-0.0751$, and its full-period precision is sensitive to the dependence approximation. The documented Holm-adjusted HAC(4) $p$-value is 0.0497; the wider-lag adjusted values exceed 0.05, and the same-draw basic interval includes zero. The 2020 dates supply 76.5\% of the signed interaction. These results describe the chosen instruments, target, design and reconstructed period.

Retransformation provides a second reason to examine the complete forecasting procedure. Sequential smearing changes the point estimates, but its incremental effect on the history interaction is imprecisely estimated. The auxiliary GARCH exercise also shows that an additional past-only adjustment can increase QLIKE loss. Calibration therefore needs its own paired evaluation under the intended target and loss. Its availability at the forecast date alone does not establish an accuracy improvement.

Temporal concentration and persistence are distinct: 2020 supplies most of the signed interaction, while exploratory post-2020 estimates remain negative and smaller. Full-period precision depends on the HAC bandwidth. The fixed-design inference uses saved losses from the selected panel. The original comparison uses one rolling length, 1,008 eligible labels. The targeted feature-redundancy rerun is exploratory and is reported separately in Appendix~\ref{app:redundancy-check}; other rolling lengths require additional fits. Wider constituent coverage and a frozen later-period evaluation would test the scope of the findings.

\vspace*{1cm}

\par\noindent\begin{minipage}{\linewidth}
	
\section*{Declarations}

\noindent\textbf{Author contributions.}  N.A., and S.T.R. contributed to conceptualization. N.A., Methodology. N.A formal analysis. N.A., A.A.T., S.T.R., original draft. N.A. developed the software. N.A. data collection. A.A.T., N.A., and S.T.R. reviewed and edited the manuscript. S.T.R., and A.A.T., supervised the work.

\vspace*{0.5cm}
\noindent\textbf{Data and code availability.} The authors can arrange authorized access to the forecast replication code, the frozen Yahoo-input, forecast, fit and verification snapshot, subject to the applicable data-use terms. Yahoo Finance prices were retrieved on 4 November 2026; source hashes and run manifests identify the saved extracts. CRSP prices and FTSE Russell holdings accessed through WRDS are licensed and are not redistributed; provider authorization is required. Retrieval and adjustment procedures are described in Section~\ref{sec:data}. Later provider downloads may differ.

\vspace*{0.5cm}
\noindent\textbf{Funding.} This research received no funding.

\vspace*{0.5cm}
\noindent\textbf{Competing interests.} The authors declare no conflicts of interest.

\vspace*{0.5cm}
\noindent\textbf{Use of artificial intelligence.} The authors used Codex (OpenAI) to assist with code review and debugging and to review the manuscript for clarity and readability. The authors take responsibility for the final article.
\end{minipage}\par

\FloatBarrier
\reservetable{430pt}
\appendix
\numberwithin{table}{section}
\numberwithin{figure}{section}
\numberwithin{equation}{section}

\section{Instrument definitions and data diagnostics}\label{app:data-diagnostics}

\subsection{Instrument identifiers}
\begin{table}[!htbp]\centering\papertablesize
\caption{Instrument identifiers and selected exposures}\label{tab:instrument-key}
\begin{tabular}{@{}L{0.15\linewidth}L{0.77\linewidth}@{}}\toprule
Ticker & Instrument or exposure \\ \midrule
SPY & U.S. large-cap equity ETF tracking the Standard \& Poor's 500 index \\
QQQ & U.S.-listed equity ETF tracking the Nasdaq-100 \\
DIA & U.S. equity ETF tracking the Dow Jones Industrial Average \\
IWM & U.S. small-cap equity ETF tracking the Russell 2000 \\
EFA & Developed-market equity ETF covering Europe, Australasia, and the Far East \\
EEM & Emerging-market equity ETF \\
EWJ & Japan equity ETF \\
EWU & United Kingdom equity ETF \\
FXE & Euro-linked currency trust \\
FXY & Japanese-yen-linked currency trust \\
FXB & British-pound-linked currency trust \\
UUP & U.S.-dollar bullish currency fund \\
GLD & Gold-linked trust \\
SLV & Silver-linked trust \\
USO & Oil-futures-linked fund \\
DBC & Broad commodity-futures-linked fund \\
AAPL & Apple Inc. common stock \\
MSFT & Microsoft Corporation common stock \\
AMZN & Amazon.com, Inc. common stock \\
JPM & JPMorgan Chase \& Co. common stock \\
XOM & Exxon Mobil Corporation common stock \\
\bottomrule\end{tabular}
\notetext{Tickers identify the Yahoo Finance series used in the analysis. ETF means exchange-traded fund. Descriptions identify the selected exposure and do not imply identical fund structures or a probability sample of securities.}
\end{table}

\FloatBarrier

\reservetable{540pt}
\subsection{Distributional and serial-dependence diagnostics}

Table~\ref{tab:app-returns} reports adjusted-close return moments for each instrument; Table~\ref{tab:app-target} reports the same statistics for the annualized five-session Garman--Klass target. Each instrument contributes 1,715 observations in the primary evaluation. Pooled rows are retained alongside the instrument-level diagnostics; the ticker rows expose heterogeneity across the selected panel. Jarque--Bera reference $p$-values use the conventional $\chi^2(2)$ approximation and are not adjusted for the repeated-date panel dependence.

\reservetable{430pt}
\begingroup\papertablesize\setlength{\tabcolsep}{2pt}
\begin{longtable}{@{}L{.13\linewidth}R{.085\linewidth}R{.10\linewidth}R{.10\linewidth}R{.10\linewidth}R{.13\linewidth}R{.16\linewidth}R{.125\linewidth}@{}}
\caption{Instrument-level adjusted-close return distributions, 2020--2026}\label{tab:app-returns}\\
\toprule Instrument & $N$ & Mean & SD & Skewness & \shortstack[r]{Excess\\$K-3$} & \shortstack[r]{JB statistic\\($\times10^{4}$)} & JB $p$-value \\ \midrule
\endfirsthead\multicolumn{8}{c}{Table \thetable{} -- continued}\\
\toprule Instrument & $N$ & Mean & SD & Skewness & \shortstack[r]{Excess\\$K-3$} & \shortstack[r]{JB statistic\\($\times10^{4}$)} & JB $p$-value \\ \midrule\endhead
\midrule\multicolumn{8}{r}{Continued on next page}\\\endfoot\bottomrule\endlastfoot
SPY & 1,715 & 0.0566 & 1.2599 & -0.5547 & 13.9150 & 1.3924 & $<0.0001$ \\
QQQ & 1,715 & 0.0739 & 1.5629 & -0.3478 & 6.9383 & 0.3475 & $<0.0001$ \\
DIA & 1,715 & 0.0426 & 1.2218 & -0.8721 & 21.9109 & 3.4524 & $<0.0001$ \\
IWM & 1,715 & 0.0357 & 1.6298 & -0.6927 & 7.8142 & 0.4500 & $<0.0001$ \\
EFA & 1,715 & 0.0347 & 1.2076 & -0.9233 & 13.8751 & 1.4001 & $<0.0001$ \\
EEM & 1,715 & 0.0283 & 1.4217 & -0.7788 & 9.8420 & 0.7095 & $<0.0001$ \\
EWJ & 1,715 & 0.0356 & 1.2293 & -0.3890 & 6.0993 & 0.2702 & $<0.0001$ \\
EWU & 1,715 & 0.0330 & 1.2937 & -1.1676 & 16.1221 & 1.8963 & $<0.0001$ \\
FXE & 1,715 & 0.0013 & 0.4668 & 0.1627 & 1.5649 & 0.0183 & $<0.0001$ \\
FXY & 1,715 & -0.0238 & 0.6139 & 0.5818 & 4.5593 & 0.1582 & $<0.0001$ \\
FXB & 1,715 & 0.0058 & 0.5492 & -0.3137 & 4.7568 & 0.1645 & $<0.0001$ \\
UUP & 1,715 & 0.0152 & 0.4593 & -0.0714 & 5.2050 & 0.1937 & $<0.0001$ \\
GLD & 1,715 & 0.0560 & 1.1716 & -0.7626 & 6.7242 & 0.3397 & $<0.0001$ \\
SLV & 1,715 & 0.0629 & 2.4116 & -2.0693 & 25.8238 & 4.8877 & $<0.0001$ \\
USO & 1,715 & 0.0283 & 2.7899 & -1.6817 & 17.7610 & 2.3350 & $<0.0001$ \\
DBC & 1,715 & 0.0519 & 1.2422 & -0.8312 & 4.0990 & 0.1398 & $<0.0001$ \\
AAPL & 1,715 & 0.0923 & 1.9580 & -0.0226 & 6.3791 & 0.2908 & $<0.0001$ \\
MSFT & 1,715 & 0.0742 & 1.9016 & 0.0249 & 8.5111 & 0.5177 & $<0.0001$ \\
AMZN & 1,715 & 0.0591 & 2.2359 & 0.0801 & 4.7179 & 0.1592 & $<0.0001$ \\
JPM & 1,715 & 0.0626 & 1.9109 & -0.0823 & 12.5960 & 1.1339 & $<0.0001$ \\
XOM & 1,715 & 0.0733 & 2.0323 & -0.2318 & 4.7127 & 0.1602 & $<0.0001$ \\
\addlinespace
Pooled panel & 36,015 & 0.0428 & 1.5839 & -0.9855 & 23.3555 & 82.4386 & $<0.0001$ \\
\end{longtable}
\par\noindent\footnotesize $N$ is the observation count; SD is standard deviation; $K$ is kurtosis; JB is the Jarque--Bera statistic. The reference $p$-value uses a chi-squared distribution with two degrees of freedom.\par\normalsize

\noindent\footnotesize JB entries use the common column scale $10^{4}$. Sample skewness and excess kurtosis are unadjusted standardized central moments. The $p$-values use the conventional $\chi^2(2)$ reference distribution and do not account for dependence across dates. \normalsize
\endgroup

\par

\reservetable{430pt}
\begingroup\papertablesize\setlength{\tabcolsep}{2pt}
\begin{longtable}{@{}L{.13\linewidth}R{.085\linewidth}R{.10\linewidth}R{.10\linewidth}R{.10\linewidth}R{.13\linewidth}R{.16\linewidth}R{.125\linewidth}@{}}
\caption{Instrument-level annualized five-session Garman--Klass target distributions, 2020--2026}\label{tab:app-target}\\
\toprule Instrument & $N$ & \shortstack[r]{Mean\\($\times10^{-2}$)} & \shortstack[r]{SD\\($\times10^{-2}$)} & Skewness & \shortstack[r]{Excess\\$K-3$} & \shortstack[r]{JB statistic\\($\times10^{6}$)} & JB $p$-value \\ \midrule
\endfirsthead\multicolumn{8}{c}{Table \thetable{} -- continued}\\
\toprule Instrument & $N$ & \shortstack[r]{Mean\\($\times10^{-2}$)} & \shortstack[r]{SD\\($\times10^{-2}$)} & Skewness & \shortstack[r]{Excess\\$K-3$} & \shortstack[r]{JB statistic\\($\times10^{6}$)} & JB $p$-value \\ \midrule\endhead
\midrule\multicolumn{8}{r}{Continued on next page}\\\endfoot\bottomrule\endlastfoot
SPY & 1,715 & 2.3155 & 5.0824 & 8.6392 & 95.8224 & 0.6775 & $<0.0001$ \\
QQQ & 1,715 & 3.4756 & 5.3465 & 7.4861 & 74.7846 & 0.4157 & $<0.0001$ \\
DIA & 1,715 & 1.9486 & 4.6100 & 9.8326 & 116.9203 & 1.0045 & $<0.0001$ \\
IWM & 1,715 & 4.3663 & 7.4414 & 8.7023 & 94.1981 & 0.6557 & $<0.0001$ \\
EFA & 1,715 & 1.2495 & 2.5662 & 9.4478 & 111.3533 & 0.9116 & $<0.0001$ \\
EEM & 1,715 & 1.5477 & 2.5575 & 7.9858 & 85.3349 & 0.5386 & $<0.0001$ \\
EWJ & 1,715 & 1.1705 & 2.2302 & 9.2233 & 105.5129 & 0.8199 & $<0.0001$ \\
EWU & 1,715 & 1.3939 & 3.0535 & 9.1013 & 100.0261 & 0.7386 & $<0.0001$ \\
FXE & 1,715 & 0.1829 & 0.1868 & 4.4771 & 34.0393 & 0.0885 & $<0.0001$ \\
FXY & 1,715 & 0.2171 & 0.2742 & 4.0147 & 21.7310 & 0.0384 & $<0.0001$ \\
FXB & 1,715 & 0.1879 & 0.3988 & 9.5832 & 114.0682 & 0.9560 & $<0.0001$ \\
UUP & 1,715 & 0.2339 & 0.6181 & 10.6588 & 132.2065 & 1.2815 & $<0.0001$ \\
GLD & 1,715 & 1.3461 & 2.4192 & 10.0279 & 129.1484 & 1.2206 & $<0.0001$ \\
SLV & 1,715 & 5.5987 & 13.5031 & 13.8835 & 224.1061 & 3.6440 & $<0.0001$ \\
USO & 1,715 & 10.1157 & 29.0274 & 9.8645 & 116.9481 & 1.0051 & $<0.0001$ \\
DBC & 1,715 & 1.7710 & 2.1545 & 5.6314 & 43.2539 & 0.1428 & $<0.0001$ \\
AAPL & 1,715 & 5.9816 & 7.3319 & 5.8921 & 45.4511 & 0.1575 & $<0.0001$ \\
MSFT & 1,715 & 5.3146 & 7.2649 & 8.6600 & 102.9210 & 0.7784 & $<0.0001$ \\
AMZN & 1,715 & 7.6417 & 6.9725 & 3.7925 & 23.3794 & 0.0432 & $<0.0001$ \\
JPM & 1,715 & 5.6089 & 10.2264 & 9.5670 & 109.9154 & 0.8895 & $<0.0001$ \\
XOM & 1,715 & 6.9856 & 11.5558 & 8.9564 & 102.4916 & 0.7736 & $<0.0001$ \\
\addlinespace
Pooled panel & 36,015 & 3.2692 & 9.1327 & 21.0189 & 710.8885 & 761.0114 & $<0.0001$ \\
\end{longtable}
\par\noindent\footnotesize $N$ is the observation count; SD is standard deviation; $K$ is kurtosis; JB is the Jarque--Bera statistic. The reference $p$-value uses a chi-squared distribution with two degrees of freedom.\par\normalsize

\noindent\footnotesize JB entries use the common column scale $10^{6}$. Mean and SD entries use the common scale $10^{-2}$ in annualized decimal variance units. Sample skewness and excess kurtosis are unadjusted standardized central moments. The $p$-values use the conventional $\chi^2(2)$ reference distribution and do not account for dependence across dates. \normalsize
\endgroup

\par

\begin{figure}[h!]\centering
\includegraphics[width=0.96\linewidth]{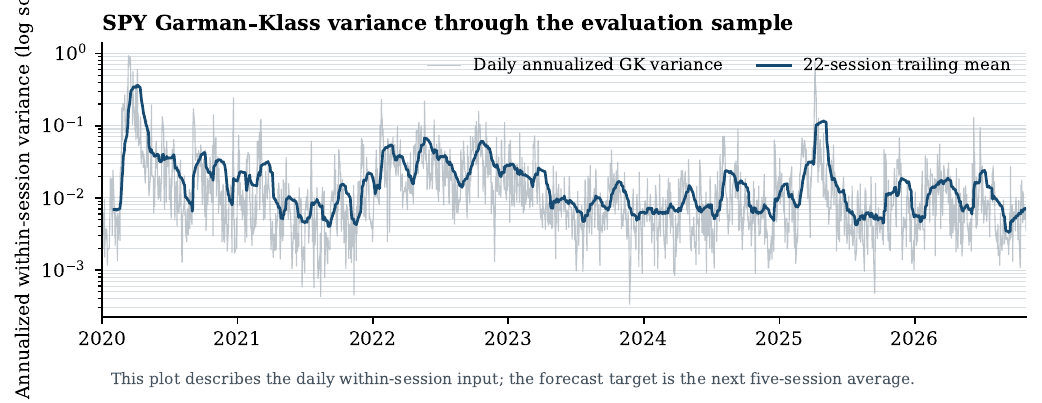}
\caption{SPY daily Garman--Klass variance and its 22-session trailing mean. Daily variance is annualized by 252 and plotted on a logarithmic scale. This is the within-session input measure, not the next-five-session forecast target.}\label{fig:spy-gk}
\end{figure}

\begin{figure}[h!]\centering
\includegraphics[width=0.96\linewidth]{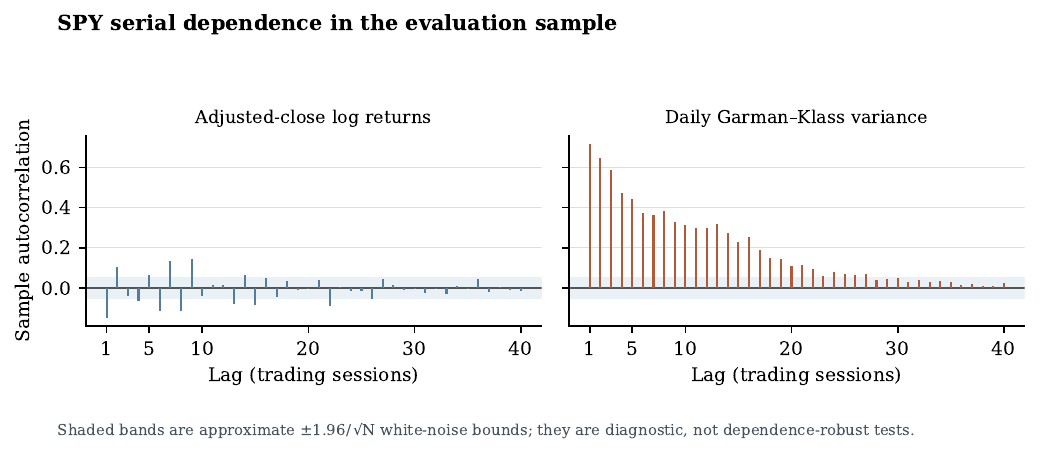}
\caption{Sample autocorrelation of SPY adjusted-close log returns and daily Garman--Klass variance through the primary evaluation period. The shaded approximate white-noise bands are descriptive diagnostics and do not replace the dependence-robust forecast-loss inference.}\label{fig:spy-acf}
\end{figure}

\FloatBarrier
\reservetable{300pt}
\subsection{Dependence of the date-level history interaction}\label{app:interaction-diagnostics}

\begin{figure}[h!]]\centering
\includegraphics[width=.98\linewidth]{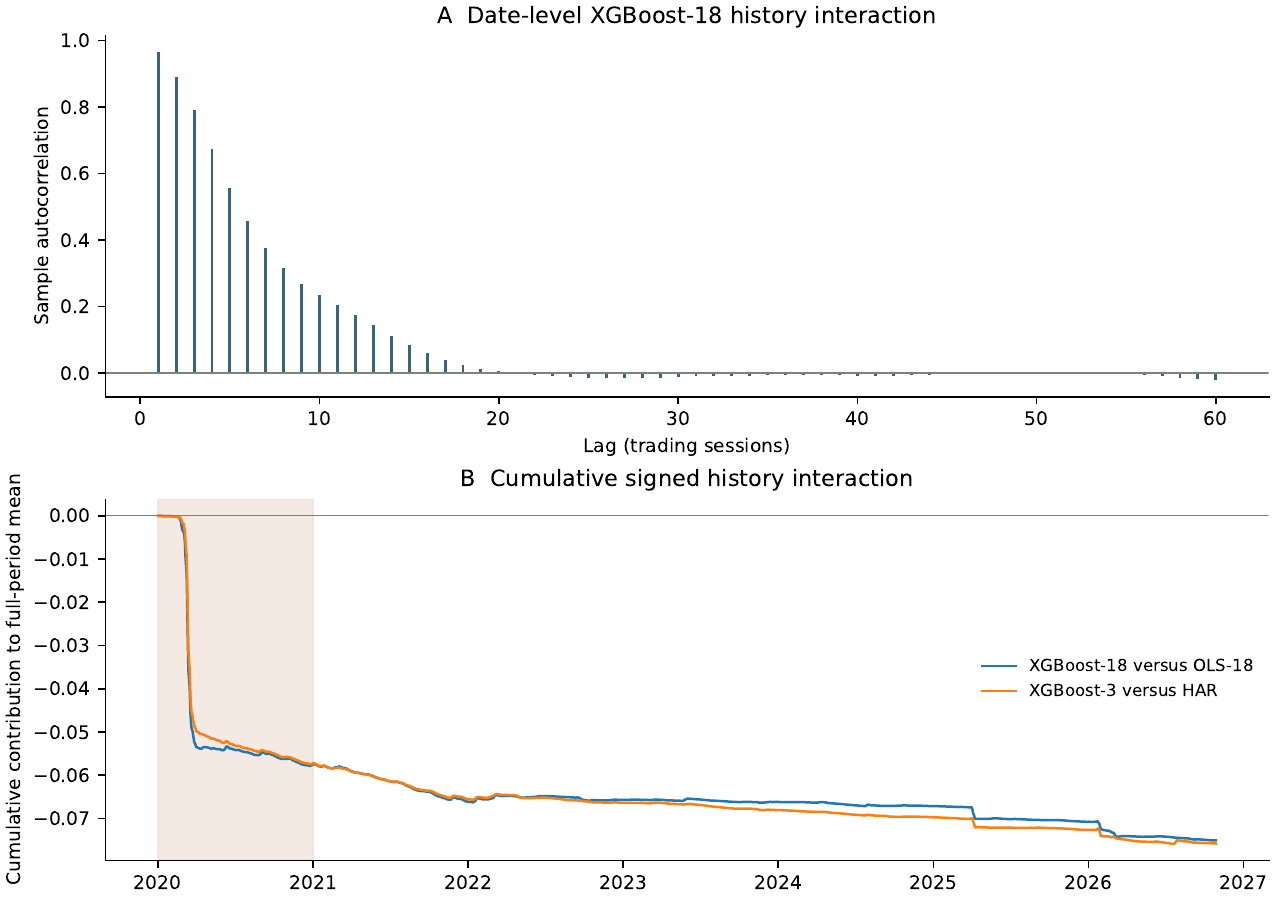}
\caption{Date-level interaction diagnostics. Panel A reports sample autocorrelations through 60 sessions for the expanding-minus-rolling XGBoost-18-minus-OLS-18 QLIKE interaction. Panel B cumulates each matched interaction and divides by the full-period count of 1,715 dates; its endpoint therefore equals the reported mean interaction. The shaded interval is 2020. Neither panel is a new forecast comparison or a selection rule for the primary bandwidth.}\label{fig:interaction-diagnostics}
\end{figure}

Figure~\ref{fig:interaction-diagnostics} examines the paired date-level loss process used for inference, rather than autocorrelation of raw returns or variance alone. The full-period XGBoost-18 interaction has autocorrelations 0.9650 at lag one and 0.5561 at lag five, so four lags are not a data-established dependence cutoff. The cumulative paths show how signed contributions accumulate and how much of the final mean is supplied during 2020. These are descriptive diagnostics of the saved forecast paths; structural instability and the concentrated early contribution limit a moving-block bootstrap interpretation that assumes a stable loss process.

\FloatBarrier

\begingroup\papertablesize\setlength{\tabcolsep}{2pt}
\begin{longtable}{@{}R{.10\linewidth}L{.29\linewidth}L{.29\linewidth}R{.22\linewidth}@{}}
\caption{Same-draw percentile and basic intervals for the primary history interaction}\label{app:bootstrap-construction}\\
\toprule Block length & Percentile 95\% interval & Basic 95\% interval & \shortstack[r]{Bootstrap mean bias\\($\times10^{-3}$)} \\ \midrule
\endfirsthead
\multicolumn{4}{c}{Table \thetable{} -- continued}\\
\toprule Block length & Percentile 95\% interval & Basic 95\% interval & \shortstack[r]{Bootstrap mean bias\\($\times10^{-3}$)} \\ \midrule
\endhead
\midrule\multicolumn{4}{r}{Continued on next page}\\\endfoot
\bottomrule\endlastfoot
5 & [-0.1409, -0.0286] & [-0.1215, -0.0092] & -0.1338 \\
22 & [-0.1827, -0.0183] & [-0.1318, 0.0325] & -1.3911 \\
44 & [-0.1903, -0.0160] & [-0.1341, 0.0402] & -1.2305 \\
\end{longtable}
\noindent\footnotesize All rows use the original non-circular synchronized resampling schedule and 5,000 draws. The observed mean is $-0.0751$. Bias is the resampled mean minus the observed mean. Basic intervals are diagnostic, pointwise transformations of the same draws, not replacement primary intervals or multiplicity-adjusted confidence sets.\normalsize
\endgroup

The original intervals use the raw resampled means. No null mean is imposed, and a fraction of resampled means above zero is not interpreted as a bootstrap $p$-value. Table~\ref{app:bootstrap-construction} also shows the basic-interval transformation, $[2\bar D-q_{.975},\,2\bar D-q_{.025}]$, using exactly the same draws, where $\bar D$ is the observed mean and $q_a$ is the empirical $a$ quantile. This diagnostic reflects the asymmetry of the sampling approximation: for a 22-session block, the percentile interval excludes zero while the basic interval includes it. Neither construction replaces the reported original percentile intervals or the ten-contrast principal test family. The alternative basic interval is exploratory and pointwise, and its disagreement reinforces the qualification on precision.

\FloatBarrier
\subsection{Early-crisis forecast and fit diagnostics}\label{app:crisis-diagnostics}
The full-period interaction is temporally concentrated: the 2020 block contributes 76.5\% of its signed sum. The dates below (12, 16 and 19 March 2020) were selected retrospectively from the largest absolute date-level interaction contributions; they are diagnostic examples, not a new evaluation sample or a selection rule. USO represents the commodity-linked instruments and SPY the broad U.S. equity market. All issued forecasts and targets shown are retained in the main analysis; no observation is deleted.
\begin{longtable}{@{}lllr rrrr@{}}
\caption{Selected early-crisis forecasts: realized target, raw forecast, smearing factor and final forecast}\label{tab:crisis-forecasts}\\
\toprule Instrument & Origin & History & Model & Target & Raw & Factor & Final \\\midrule\endfirsthead
\multicolumn{8}{c}{Table \thetable{} -- continued}\\\toprule Instrument & Origin & History & Model & Target & Raw & Factor & Final \\\midrule\endhead
\midrule\multicolumn{8}{r}{Continued on next page}\\\endfoot\bottomrule\endlastfoot
SPY & 2020-03-12 & expanding & OLS-18 & 0.7552 & 1.1375 & 1.3126 & 1.4930 \\
SPY & 2020-03-12 & expanding & XGBoost-18 & 0.7552 & 0.2372 & 1.2926 & 0.3067 \\
SPY & 2020-03-12 & rolling & OLS-18 & 0.7552 & 3.4917 & 1.4431 & 5.0388 \\
SPY & 2020-03-12 & rolling & XGBoost-18 & 0.7552 & 0.0378 & 1.5304 & 0.0579 \\
SPY & 2020-03-16 & expanding & OLS-18 & 0.5159 & 1.1322 & 1.3239 & 1.4989 \\
SPY & 2020-03-16 & expanding & XGBoost-18 & 0.5159 & 0.2447 & 1.3172 & 0.3223 \\
SPY & 2020-03-16 & rolling & OLS-18 & 0.5159 & 3.4108 & 1.4558 & 4.9656 \\
SPY & 2020-03-16 & rolling & XGBoost-18 & 0.5159 & 0.0336 & 1.5919 & 0.0534 \\
SPY & 2020-03-19 & expanding & OLS-18 & 0.2685 & 0.4160 & 1.3373 & 0.5563 \\
SPY & 2020-03-19 & expanding & XGBoost-18 & 0.2685 & 0.2118 & 1.3358 & 0.2829 \\
SPY & 2020-03-19 & rolling & OLS-18 & 0.2685 & 0.2913 & 1.4663 & 0.4271 \\
SPY & 2020-03-19 & rolling & XGBoost-18 & 0.2685 & 0.0282 & 1.7112 & 0.0483 \\
USO & 2020-03-12 & expanding & OLS-18 & 2.1764 & 0.4995 & 1.1696 & 0.5843 \\
USO & 2020-03-12 & expanding & XGBoost-18 & 2.1764 & 0.1361 & 1.1645 & 0.1585 \\
USO & 2020-03-12 & rolling & OLS-18 & 2.1764 & 0.3626 & 1.1527 & 0.4179 \\
USO & 2020-03-12 & rolling & XGBoost-18 & 2.1764 & 0.0723 & 1.1711 & 0.0847 \\
USO & 2020-03-16 & expanding & OLS-18 & 2.3142 & 0.9242 & 1.1734 & 1.0845 \\
USO & 2020-03-16 & expanding & XGBoost-18 & 2.3142 & 0.1614 & 1.1836 & 0.1910 \\
USO & 2020-03-16 & rolling & OLS-18 & 2.3142 & 0.4990 & 1.1562 & 0.5769 \\
USO & 2020-03-16 & rolling & XGBoost-18 & 2.3142 & 0.0724 & 1.2027 & 0.0870 \\
USO & 2020-03-19 & expanding & OLS-18 & 0.8645 & 1.8522 & 1.1867 & 2.1980 \\
USO & 2020-03-19 & expanding & XGBoost-18 & 0.8645 & 0.1684 & 1.2396 & 0.2088 \\
USO & 2020-03-19 & rolling & OLS-18 & 0.8645 & 0.9531 & 1.1785 & 1.1232 \\
USO & 2020-03-19 & rolling & XGBoost-18 & 0.8645 & 0.0722 & 1.3127 & 0.0947 \\
\end{longtable}
\begin{center}\begingroup\papertablesize\setlength{\tabcolsep}{2pt}
\captionof{table}{2 March 2020 refit support and XGBoost validation diagnostics}\label{tab:crisis-fit-support}
\begin{tabular}{@{}llrrrrrr@{}}\toprule
Instrument & History & $n$ & Log-target min & Log-target max & Out-of-range / checked & Boosting rounds & Validation $n$ \\\midrule
USO & rolling & 1008 & -4.7676 & -1.4188 & 68 / 162 & 102 & 200 \\
USO & expanding & 3470 & -5.6132 & -0.2268 & 31 / 162 & 105 & 200 \\
SPY & rolling & 1008 & -7.3020 & -1.9840 & 100 / 162 & 193 & 200 \\
SPY & expanding & 4063 & -7.3020 & -0.0301 & 4 / 162 & 267 & 200 \\
\bottomrule\end{tabular}
\notetext{The refit date is 2 March 2020. Each denominator of 162 is 18 supplied predictor values on each of nine trading sessions from 9 through 19 March 2020. A value is flagged when it falls outside its own predictor's minimum--maximum training range. These marginal-coordinate checks are exploratory diagnostics; they do not test joint support or isolate a cause. Each fit uses 200 validation observations and selected rounds are best iteration plus one.}
\endgroup\end{center}

\FloatBarrier
\subsection{Within-provider audit of influential USO observations}\label{app:uso-source-audit}

We independently screened the 122,322 archived Yahoo bars for 21 instruments. Observed prices must be finite and positive, $H\geq\max(O,C,L)$ and $L\leq\min(O,C,H)$. The native-price screen uses exact ordering and finds zero invalid bars, duplicate instrument-date keys or flagged duplicate dates. For rounded adjusted fields the ordering tolerance is $2\times10^{-12}$ times the largest absolute field in the bar; nine apparent ordering differences at zero tolerance disappear within that tolerance. The adjustment identity uses one positive factor, Yahoo adjusted close divided by the corresponding native close, for all four OHLC fields. The maximum relative discrepancy from the stored fields was $1.63\times10^{-14}$, and common-factor GK invariance held to $1.42\times10^{-15}$ in absolute terms. Missing price rows remain unfilled. The seven observed nonpositive volume values do not supply log-volume changes, including at the relevant lagged origins. No scored origin is missing any of the three volume changes. Close-to-close features use successive vendor adjusted closes; the analysis does not apply an additional split or distribution adjustment to Yahoo's recorded prices or volume.

The archived USO corporate-action file records a reverse split on 29 April 2020 (reported \texttt{quantmod} split value 8). Some influential five-session outcomes straddle the event. In the archived close series the 28 and 29 April bars close at 17.04 and 18.00, with a common within-bar adjustment factor of one on both dates. The boundary shows no eightfold discontinuity. We regenerated GK on each native and common-factor adjusted bar and reconciled it with the prepared panel and the exact forecast targets. This verifies the archived price basis internally and does not authenticate the exchange prices or independently validate Yahoo's historical split conventions. Table~\ref{app:uso-april-audit} lists the expanding forecasts behind the MSE concentration. For each origin, the replication records give the five target-session dates and GK values, as well as raw and final forecasts, smearing factors and refit records under both histories. These checks retain the original observations.

\begin{center}\begingroup\papertablesize
\captionof{table}{Five high-influence USO forecast origins in April 2020}\label{app:uso-april-audit}
\begin{tabular}{@{}llrrr@{}}\toprule
Origin & Target ends & Observed GK target & OLS-18 final & XGBoost-18 final\\ \midrule
2020-04-21 & 2020-04-28 & 0.8372 & 11.4906 & 1.2391 \\
2020-04-22 & 2020-04-29 & 0.6267 & 5.4442 & 1.2358 \\
2020-04-23 & 2020-04-30 & 0.5296 & 3.2123 & 1.0402 \\
2020-04-27 & 2020-05-04 & 0.3862 & 5.9045 & 1.1595 \\
2020-04-28 & 2020-05-05 & 0.3229 & 2.7840 & 0.9364 \\
\bottomrule\end{tabular}
\notetext{Forecasts use the expanding procedure with sequential smearing. Each target averages the next five calendar trading sessions and annualizes by 252. All shown origins use the 1 April 2020 refit. Table~\ref{app:loss-concentration} reports contributions to the pooled mean, so each date contribution is divided by the full 1,715-date denominator.}
\endgroup\end{center}

\FloatBarrier
\clearpage
\section{Additional inferential and benchmark evidence}\label{app:additional-results}

This appendix reports the exposure and instrument summaries, historical-window and dependence sensitivities, restricted holdings strata, and the supplementary same-close volatility-targeting accounting illustration. Exposure and security-level contrasts are exploratory. The holdings exercise covers only the snapshots available under the institutional entitlement. The portfolio rule uses separately fitted one-session forecasts and assumes an idealized same-close trade. Its retrospective accounting does not test implementable economic value for the five-session range forecast. Table labels XGBoost-3 and XGBoost-18 denote extreme gradient boosting with three and 18 supplied features; OLS-18 denotes ordinary least squares with 18 supplied features; HAR denotes the three-predictor heterogeneous autoregressive model. The exponentially weighted moving average (EWMA) forecast is an additional portfolio benchmark. All test columns report numerical $p$-values. Unadjusted HAC and Holm-adjusted values are identified in each table. 

\subsection{Exposure-class and instrument-level summaries}

Subgroup and instrument summaries in Tables~\ref{app:exposure-groups}--\ref{app:asset-level-interactions} are exploratory. Instrument-date counts enumerate paired outcomes; inference uses the ordered date means, and the counts do not represent independent observations.

\begingroup
\setlength{\tabcolsep}{3pt}
\papertablesize
\reservetable{290pt}
\begin{longtable}{@{}L{.22\linewidth}L{.10\linewidth}R{.14\linewidth}R{.13\linewidth}R{.10\linewidth}L{.23\linewidth}@{}}
\caption{Exploratory exposure-group history interactions in the 2020--2026 Yahoo panel}\label{app:exposure-groups}\\
\toprule Exposure group & Learner & \shortstack[r]{Mean\\interaction} & Instrument-dates & \shortstack[r]{HAC(4)\\$p$-value} & \shortstack[l]{MBB-22 95\%\\confidence interval} \\ \midrule
\endfirsthead
\multicolumn{6}{c}{\tablename\ \thetable{} -- continued}\\
\toprule Exposure group & Learner & \shortstack[r]{Mean\\interaction} & Instrument-dates & \shortstack[r]{HAC(4)\\$p$-value} & \shortstack[l]{MBB-22 95\%\\confidence interval} \\ \midrule
\endhead
\midrule\multicolumn{6}{r}{Continued on next page}\\\endfoot
\bottomrule\endlastfoot
US equity ETFs & XGBoost-18 & -0.0975 & 6860 & 0.0151 & [-0.2467, -0.0167] \\
US equity ETFs & XGBoost-3 & -0.0937 & 6860 & 0.0103 & [-0.2251, -0.0236] \\
international equity ETFs & XGBoost-18 & -0.1268 & 6860 & 0.0245 & [-0.3267, -0.0231] \\
international equity ETFs & XGBoost-3 & -0.1153 & 6860 & 0.0198 & [-0.2903, -0.0254] \\
currency ETFs & XGBoost-18 & -0.0352 & 6860 & 0.0222 & [-0.0804, -0.0082] \\
currency ETFs & XGBoost-3 & -0.0333 & 6860 & 0.0162 & [-0.0686, -0.0101] \\
commodity ETFs & XGBoost-18 & -0.0607 & 6860 & 0.0036 & [-0.1309, -0.0147] \\
commodity ETFs & XGBoost-3 & -0.0575 & 6860 & 0.0040 & [-0.1263, -0.0163] \\
individual stocks & XGBoost-18 & -0.0590 & 8575 & 0.0175 & [-0.1507, -0.0098] \\
individual stocks & XGBoost-3 & -0.0784 & 8575 & 0.0107 & [-0.1889, -0.0208] \\
\end{longtable}
\endgroup
\noindent\footnotesize Groups are descriptive summaries over instruments with the same traded exposure class. HAC(4) $p$-values are unadjusted; confidence intervals and tests are not adjusted across these subgroup contrasts. \normalsize

\begingroup
\setlength{\tabcolsep}{3pt}
\papertablesize
\reservetable{570pt}
\begin{longtable}{@{}L{.28\linewidth}L{.22\linewidth}R{.28\linewidth}R{.14\linewidth}@{}}
\caption{Instrument-specific mean QLIKE history interactions}\label{app:asset-level-interactions}\\
\toprule Instrument & Learner & Mean interaction & Dates \\ \midrule
\endfirsthead
\multicolumn{4}{c}{\tablename\ \thetable{} -- continued}\\
\toprule Instrument & Learner & Mean interaction & Dates \\ \midrule
\endhead
\midrule\multicolumn{4}{r}{Continued on next page}\\\endfoot
\bottomrule\endlastfoot
AAPL & XGBoost-18 & -0.0322 & 1715 \\
AMZN & XGBoost-18 & -0.0101 & 1715 \\
DBC & XGBoost-18 & -0.0402 & 1715 \\
DIA & XGBoost-18 & -0.1249 & 1715 \\
EEM & XGBoost-18 & -0.0705 & 1715 \\
EFA & XGBoost-18 & -0.2040 & 1715 \\
EWJ & XGBoost-18 & -0.1133 & 1715 \\
EWU & XGBoost-18 & -0.1195 & 1715 \\
FXB & XGBoost-18 & -0.0380 & 1715 \\
FXE & XGBoost-18 & -0.0066 & 1715 \\
FXY & XGBoost-18 & -0.0181 & 1715 \\
GLD & XGBoost-18 & -0.0384 & 1715 \\
IWM & XGBoost-18 & -0.1249 & 1715 \\
JPM & XGBoost-18 & -0.0713 & 1715 \\
MSFT & XGBoost-18 & -0.0264 & 1715 \\
QQQ & XGBoost-18 & -0.0643 & 1715 \\
SLV & XGBoost-18 & -0.0668 & 1715 \\
SPY & XGBoost-18 & -0.0760 & 1715 \\
USO & XGBoost-18 & -0.0973 & 1715 \\
UUP & XGBoost-18 & -0.0779 & 1715 \\
XOM & XGBoost-18 & -0.1552 & 1715 \\
AAPL & XGBoost-3 & -0.0453 & 1715 \\
AMZN & XGBoost-3 & -0.0086 & 1715 \\
DBC & XGBoost-3 & -0.0352 & 1715 \\
DIA & XGBoost-3 & -0.1218 & 1715 \\
EEM & XGBoost-3 & -0.0606 & 1715 \\
EFA & XGBoost-3 & -0.1923 & 1715 \\
EWJ & XGBoost-3 & -0.1021 & 1715 \\
EWU & XGBoost-3 & -0.1060 & 1715 \\
FXB & XGBoost-3 & -0.0348 & 1715 \\
FXE & XGBoost-3 & -0.0122 & 1715 \\
FXY & XGBoost-3 & -0.0251 & 1715 \\
GLD & XGBoost-3 & -0.0330 & 1715 \\
IWM & XGBoost-3 & -0.1190 & 1715 \\
JPM & XGBoost-3 & -0.1041 & 1715 \\
MSFT & XGBoost-3 & -0.0280 & 1715 \\
QQQ & XGBoost-3 & -0.0583 & 1715 \\
SLV & XGBoost-3 & -0.0772 & 1715 \\
SPY & XGBoost-3 & -0.0758 & 1715 \\
USO & XGBoost-3 & -0.0847 & 1715 \\
UUP & XGBoost-3 & -0.0611 & 1715 \\
XOM & XGBoost-3 & -0.2059 & 1715 \\
\end{longtable}
\endgroup
\noindent\footnotesize These instrument-level averages are descriptive. Their date-level dependence is not used to make separate significance claims.\normalsize

\reservetable{230pt}
\subsection{Historical Yahoo window and inference sensitivities}

\begingroup
\setlength{\tabcolsep}{3pt}
\papertablesize
\reservetable{180pt}
\begin{longtable}{@{}L{.20\linewidth}R{.13\linewidth}R{.30\linewidth}R{.29\linewidth}@{}}
\caption{Specified temporal summaries for the 2014--2017 XGBoost-18 interaction}\label{app:temporal-2014}\\
\toprule Period & Dates & \shortstack[r]{Mean date-level\\interaction} & \shortstack[r]{Share of\\signed total} \\ \midrule
\endfirsthead
\multicolumn{4}{c}{\tablename\ \thetable{} -- continued}\\
\toprule Period & Dates & \shortstack[r]{Mean date-level\\interaction} & \shortstack[r]{Share of\\signed total} \\ \midrule
\endhead
\midrule\multicolumn{4}{r}{Continued on next page}\\\endfoot
\bottomrule\endlastfoot
2014 & 252 & -0.0042 & 12.6\% \\
2015 & 252 & -0.0131 & 39.1\% \\
2016 & 252 & -0.0146 & 43.5\% \\
2017 & 251 & -0.0016 & 4.9\% \\
2014-2017 & 1007 & -0.0084 & 100.0\% \\
\end{longtable}
\endgroup
\noindent\footnotesize The 2014--2017 evaluation is historical and retrospective. Period means and signed shares are descriptive.\normalsize

\begingroup
\setlength{\tabcolsep}{3pt}
\papertablesize
\reservetable{300pt}
\begin{longtable}{@{}L{.39\linewidth}R{.11\linewidth}R{.10\linewidth}R{.10\linewidth}L{.23\linewidth}@{}}
\caption{Primary 2014--2017 historical QLIKE comparisons}\label{app:historical-primary}\\
\toprule Contrast & Mean & \shortstack[r]{HAC(4)\\$p$-value} & \shortstack[r]{Holm\\$p$-value} & \shortstack[l]{MBB-22 95\%\\confidence interval} \\ \midrule
\endfirsthead
\multicolumn{5}{c}{\tablename\ \thetable{} -- continued}\\
\toprule Contrast & Mean & \shortstack[r]{HAC(4)\\$p$-value} & \shortstack[r]{Holm\\$p$-value} & \shortstack[l]{MBB-22 95\%\\confidence interval} \\ \midrule
\endhead
\midrule\multicolumn{5}{r}{Continued on next page}\\\endfoot
\bottomrule\endlastfoot
XGBoost-18 smeared history interaction & -0.0084 & 0.0288 & 0.1080 & [-0.0157, -0.0022] \\
XGBoost-3 smeared history interaction & -0.0084 & 0.0012 & 0.0071 & [-0.0152, -0.0026] \\
expanding DirectQLIKE-18 minus OLS-18 smeared & 0.0017 & 0.7403 & 1.0000 & [-0.0132, 0.0111] \\
expanding OLS-18 smeared minus HAR smeared & -0.0085 & 0.0270 & 0.1080 & [-0.0140, -0.0039] \\
expanding XGBoost-18 minus OLS-18 smeared & 0.0076 & 0.0018 & 0.0091 & [0.0020, 0.0139] \\
expanding XGBoost-18 smear minus raw & 0.0092 & $<0.0001$ & 0.0003 & [0.0038, 0.0145] \\
expanding XGBoost-18 smeared minus HAR smeared & -0.0010 & 0.7938 & 1.0000 & [-0.0080, 0.0060] \\
expanding XGBoost-3 minus HAR smeared & 0.0166 & 0.0006 & 0.0040 & [0.0081, 0.0289] \\
rolling XGBoost-18 minus OLS-18 smeared & 0.0160 & 0.0003 & 0.0027 & [0.0094, 0.0243] \\
rolling XGBoost-3 minus HAR smeared & 0.0250 & $<0.0001$ & $<0.0001$ & [0.0156, 0.0380] \\
\end{longtable}
\endgroup
\noindent\footnotesize The ten primary QLIKE comparisons form the declared historical-window Holm family. The HAC(4) column reports unadjusted $p$-values and the Holm column reports adjusted $p$-values.  This window is a retrospective sensitivity analysis.\normalsize

\begingroup\papertablesize\setlength{\tabcolsep}{3pt}
\begin{longtable}{@{}L{.39\linewidth}rrrrrrrr@{}}
\caption{Raw and Holm-adjusted HAC bandwidth sensitivity for primary QLIKE contrasts}\label{app:hac-sensitivity}\\
\toprule Contrast & \multicolumn{2}{c}{HAC(4)} & \multicolumn{2}{c}{HAC(10)} & \multicolumn{2}{c}{HAC(21)} & \multicolumn{2}{c}{HAC(43)}\\
\cmidrule(lr){2-3}\cmidrule(lr){4-5}\cmidrule(lr){6-7}\cmidrule(l){8-9}
 & Raw & Holm & Raw & Holm & Raw & Holm & Raw & Holm\\ \midrule\endfirsthead
\toprule Contrast & \multicolumn{2}{c}{HAC(4)} & \multicolumn{2}{c}{HAC(10)} & \multicolumn{2}{c}{HAC(21)} & \multicolumn{2}{c}{HAC(43)}\\
 & Raw & Holm & Raw & Holm & Raw & Holm & Raw & Holm\\ \midrule\endhead
\bottomrule\endlastfoot
Rolling XGBoost-18 vs. OLS-18 & 0.0068 & 0.0476 & 0.0414 & 0.2481 & 0.0814 & 0.4475 & 0.1018 & 0.5090 \\
Expanding XGBoost-18 vs. OLS-18 & 0.0083 & 0.0496 & 0.0231 & 0.1851 & 0.0319 & 0.2555 & 0.0339 & 0.2710 \\
XGBoost-18 history interaction & 0.0124 & 0.0497 & 0.0576 & 0.2481 & 0.1031 & 0.4475 & 0.1263 & 0.5090 \\
Rolling XGBoost-3 vs. HAR & 0.0045 & 0.0361 & 0.0313 & 0.2194 & 0.0639 & 0.4475 & 0.0839 & 0.5034 \\
Expanding XGBoost-3 vs. HAR & 0.0011 & 0.0098 & 0.0091 & 0.0823 & 0.0206 & 0.1851 & 0.0248 & 0.2228 \\
XGBoost-3 history interaction & 0.0093 & 0.0496 & 0.0457 & 0.2481 & 0.0817 & 0.4475 & 0.1031 & 0.5090 \\
Expanding XGBoost-18, smeared vs. raw & 0.0284 & 0.0851 & 0.0546 & 0.2481 & 0.0672 & 0.4475 & 0.0708 & 0.4954 \\
Expanding XGBoost-18 vs. HAR & 0.3049 & 0.6099 & 0.2939 & 0.5878 & 0.2784 & 0.5568 & 0.2413 & 0.5090 \\
Expanding OLS-18 vs. HAR & 0.0002 & 0.0020 & 0.0016 & 0.0160 & 0.0042 & 0.0418 & 0.0067 & 0.0669 \\
Expanding DirectQLIKE-18 vs. OLS-18 & 0.8365 & 0.8365 & 0.8554 & 0.8554 & 0.8633 & 0.8633 & 0.8647 & 0.8647 \\
\end{longtable}
\noindent\footnotesize Holm adjustments use all ten unrounded QLIKE tests separately at each bandwidth. HAC(4) is the documented initial specification. The replication materials report the same calculation for the ten squared-error contrasts. \normalsize
\endgroup

\begingroup
\setlength{\tabcolsep}{3pt}
\papertablesize
\reservetable{320pt}
\begin{longtable}{@{}L{.34\linewidth}L{.195\linewidth}L{.195\linewidth}L{.195\linewidth}@{}}
\caption{Moving-block interval sensitivity for primary 2020--2026 QLIKE contrasts}\label{app:block-sensitivity}\\
\toprule Contrast & \shortstack[l]{MBB-5 95\%\\confidence interval} & \shortstack[l]{MBB-22 95\%\\confidence interval} & \shortstack[l]{MBB-44 95\%\\confidence interval} \\ \midrule
\endfirsthead
\multicolumn{4}{c}{\tablename\ \thetable{} -- continued}\\
\toprule Contrast & \shortstack[l]{MBB-5 95\%\\confidence interval} & \shortstack[l]{MBB-22 95\%\\confidence interval} & \shortstack[l]{MBB-44 95\%\\confidence interval} \\ \midrule
\endhead
\midrule\multicolumn{4}{r}{Continued on next page}\\\endfoot
\bottomrule\endlastfoot
XGBoost-18 smeared history interaction & [-0.1409, -0.0286] & [-0.1827, -0.0183] & [-0.1903, -0.0160] \\
XGBoost-3 smeared history interaction & [-0.1400, -0.0312] & [-0.1787, -0.0230] & [-0.1855, -0.0206] \\
expanding DirectQLIKE-18 minus OLS-18 smeared & [-0.0060, 0.0041] & [-0.0071, 0.0047] & [-0.0071, 0.0049] \\
expanding OLS-18 smeared minus HAR smeared & [-0.0253, -0.0087] & [-0.0285, -0.0070] & [-0.0293, -0.0067] \\
expanding XGBoost-18 minus OLS-18 smeared & [0.0044, 0.0234] & [0.0032, 0.0264] & [0.0029, 0.0270] \\
expanding XGBoost-18 smear minus raw & [-0.0171, -0.0011] & [-0.0187, 0.0005] & [-0.0190, 0.0003] \\
expanding XGBoost-18 smeared minus HAR smeared & [-0.0092, 0.0030] & [-0.0084, 0.0028] & [-0.0080, 0.0027] \\
expanding XGBoost-3 minus HAR smeared & [0.0066, 0.0235] & [0.0047, 0.0285] & [0.0043, 0.0294] \\
rolling XGBoost-18 minus OLS-18 smeared & [0.0361, 0.1576] & [0.0235, 0.2079] & [0.0213, 0.2165] \\
rolling XGBoost-3 minus HAR smeared & [0.0403, 0.1596] & [0.0292, 0.2050] & [0.0264, 0.2141] \\
\end{longtable}
\endgroup
\noindent\footnotesize Intervals use synchronized dates and 5,000 draws. They condition on the issued forecast paths and are not multiplicity-adjusted.\normalsize

\begingroup
\setlength{\tabcolsep}{3pt}
\papertablesize
\reservetable{330pt}
\begin{longtable}{@{}L{.39\linewidth}R{.11\linewidth}R{.10\linewidth}R{.10\linewidth}L{.23\linewidth}@{}}
\caption{Primary MSE contrasts for the 2020--2026 Yahoo panel}\label{app:mse-contrasts}\\
\toprule Contrast & \shortstack[r]{Mean\\($\times10^{-3}$)} & \shortstack[r]{HAC(4)\\$p$-value} & \shortstack[r]{Holm\\$p$-value} & \shortstack[l]{MBB-22 95\%\\confidence interval\\($\times10^{-3}$)} \\ \midrule
\endfirsthead
\multicolumn{5}{c}{\tablename\ \thetable{} -- continued}\\
\toprule Contrast & \shortstack[r]{Mean\\($\times10^{-3}$)} & \shortstack[r]{HAC(4)\\$p$-value} & \shortstack[r]{Holm\\$p$-value} & \shortstack[l]{MBB-22 95\%\\confidence interval\\($\times10^{-3}$)} \\ \midrule
\endhead
\midrule\multicolumn{5}{r}{Continued on next page}\\\endfoot
\bottomrule\endlastfoot
XGBoost-18 smeared history interaction & -1.3965 & 0.7386 & 1.0000 & [-10.6709, 7.5469] \\
XGBoost-3 smeared history interaction & -3.2150 & 0.0012 & 0.0111 & [-7.3098, -0.6119] \\
expanding DirectQLIKE-18 minus OLS-18 smeared & 1.4168 & 0.1229 & 0.9833 & [-0.0143, 3.7334] \\
expanding OLS-18 smeared minus HAR smeared & 4.6597 & 0.2306 & 1.0000 & [-0.1535, 14.6437] \\
expanding XGBoost-18 minus OLS-18 smeared & -4.7472 & 0.2447 & 1.0000 & [-15.0314, 0.1916] \\
expanding XGBoost-18 smear minus raw & -0.1591 & 0.4560 & 1.0000 & [-0.6513, 0.2555] \\
expanding XGBoost-18 smeared minus HAR smeared & -0.0875 & 0.7646 & 1.0000 & [-0.6097, 0.4565] \\
expanding XGBoost-3 minus HAR smeared & 0.5131 & 0.1853 & 1.0000 & [-0.0531, 1.4698] \\
rolling XGBoost-18 minus OLS-18 smeared & -3.3508 & 0.2871 & 1.0000 & [-10.8738, 1.0836] \\
rolling XGBoost-3 minus HAR smeared & 3.7280 & 0.0010 & 0.0103 & [0.6833, 8.1889] \\
\end{longtable}
\endgroup
\noindent\footnotesize Mean differences and both interval bounds use the common column scale $10^{-3}$. Ten declared MSE contrasts form a separate Holm family. The HAC(4) column reports unadjusted $p$-values and the Holm column reports adjusted $p$-values.  MSE non-rejection is not evidence of equivalence.\normalsize

\reservetable{250pt}
\subsection{Loss-ranking and calibration diagnostics}\label{app:loss-calibration-diagnostics}

Table~\ref{app:loss-concentration} decomposes the expanding XGBoost-18-minus-OLS-18 loss differences by instrument and by forecast origin. Each instrument contribution is its mean paired loss difference divided by 21; each date contribution is its within-date mean divided by 1,715. The complete signed contributions sum to the full-period contrast. The displayed rows have the largest absolute MSE contributions; unshown observations remain in the estimates, and shares of a signed total need not be bounded by zero and one because other contributions can offset them. The largest MSE contributors need not dominate QLIKE, which weights relative errors and log forecast levels differently from absolute squared error.

\begingroup\papertablesize\setlength{\tabcolsep}{2pt}
\begin{longtable}{@{}L{.39\linewidth}R{.25\linewidth}R{.25\linewidth}@{}}
\caption{Largest absolute MSE contributors to expanding XGBoost-18 minus OLS-18}\label{app:loss-concentration}\\
\toprule Instrument or origin & \shortstack[r]{MSE contribution\\($\times10^{-3}$)} & \shortstack[r]{QLIKE contribution\\($\times10^{-3}$)} \\ \midrule
\endfirsthead
\multicolumn{3}{c}{Table \thetable{} -- continued}\\
\toprule Instrument or origin & \shortstack[r]{MSE contribution\\($\times10^{-3}$)} & \shortstack[r]{QLIKE contribution\\($\times10^{-3}$)} \\ \midrule
\endhead
\midrule\multicolumn{3}{r}{Continued on next page}\\\endfoot
\bottomrule\endlastfoot
\multicolumn{3}{@{}l}{\textit{Panel A: instruments}}\\\addlinespace
USO & -4.5263 & 2.0325 \\
DIA & -0.2388 & 0.8268 \\
XOM & 0.1276 & 1.1133 \\
SLV & -0.1043 & -0.0441 \\
SPY & -0.0526 & 0.6270 \\
All instruments & -4.7472 & 12.7769 \\
\multicolumn{3}{@{}l}{\textit{Panel B: forecast origins}}\\\addlinespace
2020-04-21 & -3.1450 & -0.0645 \\
2020-04-27 & -0.8279 & -0.0305 \\
2020-04-22 & -0.6342 & -0.0251 \\
2020-04-23 & -0.1921 & -0.0109 \\
2020-04-28 & -0.1579 & -0.0293 \\
All dates & -4.7472 & 12.7769 \\
\end{longtable}
\noindent\footnotesize Each panel separately decomposes the full paired-loss mean; panels are not added together. The five largest absolute MSE contributions are displayed, and the final row includes all observations. The same selected rows report their QLIKE contributions. All source observations remain in the primary tests.\normalsize
\endgroup

\begingroup\papertablesize\setlength{\tabcolsep}{2pt}
\begin{longtable}{@{}L{.16\linewidth}L{.18\linewidth}R{.14\linewidth}R{.14\linewidth}R{.14\linewidth}R{.14\linewidth}@{}}
\caption{Sequential smearing factors on the primary evaluation sample}\label{app:smearing-summary}\\
\toprule Model & History & Minimum & Median & 95th percentile & Maximum \\ \midrule
\endfirsthead
\multicolumn{6}{c}{Table \thetable{} -- continued}\\
\toprule Model & History & Minimum & Median & 95th percentile & Maximum \\ \midrule
\endhead
\midrule\multicolumn{6}{r}{Continued on next page}\\\endfoot
\bottomrule\endlastfoot
HAR & Rolling & 0.9207 & 1.2230 & 1.5416 & 1.7605 \\
OLS-18 & Rolling & 0.9236 & 1.2232 & 1.5249 & 1.7289 \\
XGBoost-3 & Rolling & 0.8465 & 1.2782 & 2.0738 & 2.5140 \\
XGBoost-18 & Rolling & 0.8753 & 1.2640 & 2.0796 & 2.4952 \\
HAR & Expanding & 0.8422 & 1.1868 & 1.3941 & 1.5507 \\
OLS-18 & Expanding & 0.8665 & 1.1663 & 1.3695 & 1.5376 \\
XGBoost-3 & Expanding & 0.8698 & 1.1937 & 1.4450 & 1.5934 \\
XGBoost-18 & Expanding & 0.8976 & 1.1620 & 1.4066 & 1.5121 \\
\end{longtable}
\noindent\footnotesize Each model/history summary pools 36,015 instrument-date factors. Factors use only the most recent 504 matured issued errors, with minimum 126. Instrument/date detail and fifth percentiles are retained in the analysis files. The saved primary smeared forecasts have no positive-floor violations, although the code does not reapply that floor after smearing.\normalsize
\endgroup

Table~\ref{app:smearing-summary} pools the 36,015 primary instrument-date factors for each model/history combination. The factor dispersion describes an operational mixture of successive fitted-model errors, not a distribution of residuals from a single fitted regression. The numerical analysis files additionally identify maxima by instrument and date and retain every contribution, including the largest individual instrument-date MSE differences. None of these diagnostics excludes observations from the primary comparison.

\reservetable{250pt}
\subsection{Targeted supplied-feature redundancy sensitivity}\label{app:redundancy-check}

The exploratory sensitivity removes the negative-return feature from both OLS and XGBoost, retaining the current return and its absolute value. Equation~\ref{eq:return-redundancy} implies that this removal does not reduce the linear span. It does change the supplied feature pool seen by XGBoost's per-tree column sampling; the sampling fraction remains 0.7, rather than being selected to preserve an accuracy result. Every other setting, source mask, monthly refit date, training history and calibration maturity rule is retained. New forecasts are generated from the 2018 warm-up onward, and smearing is recomputed from their own originally issued errors. This is a separate exploratory rerun, not a replacement of the 18-feature primary estimates.

\begingroup\papertablesize\setlength{\tabcolsep}{2pt}
\begin{longtable}{@{}L{.20\linewidth}L{.28\linewidth}R{.22\linewidth}R{.22\linewidth}@{}}
\caption{Exploratory Absolute loss levels in the targeted 17-feature rerun}\label{app:redundancy-levels}\\
\toprule History & Procedure & Mean QLIKE & \shortstack[r]{Mean MSE\\($\times10^{-3}$)} \\ \midrule
\endfirsthead
\multicolumn{4}{c}{Table \thetable{} -- continued}\\
\toprule History & Procedure & Mean QLIKE & \shortstack[r]{Mean MSE\\($\times10^{-3}$)} \\ \midrule
\endhead
\midrule\multicolumn{4}{r}{Continued on next page}\\\endfoot
\bottomrule\endlastfoot
Expanding & OLS-17 & -3.2065 & 10.1323 \\
Expanding & XGBoost-17 & -3.1944 & 5.3212 \\
Rolling & OLS-17 & -3.1967 & 10.4439 \\
Rolling & XGBoost-17 & -3.1108 & 7.0680 \\
\end{longtable}
\noindent\footnotesize Every row has 36,015 instrument-dates. Loss levels are descriptive and the four exploratory paired comparisons are reported separately.\normalsize
\endgroup

\begingroup\papertablesize\setlength{\tabcolsep}{2pt}
\begin{longtable}{@{}L{.34\linewidth}R{.10\linewidth}R{.08\linewidth}R{.08\linewidth}L{.17\linewidth}L{.17\linewidth}@{}}
\caption{Exploratory Targeted 17-feature QLIKE comparisons}\label{app:redundancy-qlike}\\
\toprule Comparison & \shortstack[r]{Mean} & \shortstack[r]{HAC(4)\\$p$} & \shortstack[r]{Holm\\$p$} & \shortstack[l]{MBB-22 interval} & \shortstack[l]{MBB-44 interval} \\ \midrule
\endfirsthead
\multicolumn{6}{c}{Table \thetable{} -- continued}\\
\toprule Comparison & \shortstack[r]{Mean} & \shortstack[r]{HAC(4)\\$p$} & \shortstack[r]{Holm\\$p$} & \shortstack[l]{MBB-22 interval} & \shortstack[l]{MBB-44 interval} \\ \midrule
\endhead
\midrule\multicolumn{6}{r}{Continued on next page}\\\endfoot
\bottomrule\endlastfoot
Rolling XGBoost-17 minus OLS-17 & 0.0859 & 0.0070 & 0.0281 & [0.0227, 0.2039] & [0.0205, 0.2123] \\
Expanding XGBoost-17 minus OLS-17 & 0.0121 & 0.0113 & 0.0340 & [0.0029, 0.0251] & [0.0026, 0.0256] \\
17-feature history interaction & -0.0738 & 0.0131 & 0.0340 & [-0.1806, -0.0177] & [-0.1875, -0.0153] \\
17- minus 18-feature interaction & 0.0013 & 0.0139 & 0.0340 & [0.0003, 0.0026] & [0.0003, 0.0025] \\
\end{longtable}
\noindent\footnotesize All estimates use the same 21 instruments and 1,715 evaluation dates as the original supplied-18-feature comparison. The negative-return feature is removed without tuning another setting. These four exploratory tests form their own Holm family separately under each loss; they do not alter the original primary tests. Intervals are non-circular, uncentered, pointwise percentile intervals on synchronized date blocks. The change row subtracts the original 18-feature interaction from the new 17-feature interaction.\normalsize
\endgroup

\begingroup\papertablesize\setlength{\tabcolsep}{2pt}
\begin{longtable}{@{}L{.34\linewidth}R{.10\linewidth}R{.08\linewidth}R{.08\linewidth}L{.17\linewidth}L{.17\linewidth}@{}}
\caption{Exploratory Targeted 17-feature MSE comparisons}\label{app:redundancy-mse}\\
\toprule Comparison & \shortstack[r]{Mean\\($\times10^{-3}$)} & \shortstack[r]{HAC(4)\\$p$} & \shortstack[r]{Holm\\$p$} & \shortstack[l]{MBB-22 interval\\($\times10^{-3}$)} & \shortstack[l]{MBB-44 interval\\($\times10^{-3}$)} \\ \midrule
\endfirsthead
\multicolumn{6}{c}{Table \thetable{} -- continued}\\
\toprule Comparison & \shortstack[r]{Mean\\($\times10^{-3}$)} & \shortstack[r]{HAC(4)\\$p$} & \shortstack[r]{Holm\\$p$} & \shortstack[l]{MBB-22 interval\\($\times10^{-3}$)} & \shortstack[l]{MBB-44 interval\\($\times10^{-3}$)} \\ \midrule
\endhead
\midrule\multicolumn{6}{r}{Continued on next page}\\\endfoot
\bottomrule\endlastfoot
Rolling XGBoost-17 minus OLS-17 & -3.3759 & 0.2829 & 0.9557 & [-10.8865, 1.0271] & [-12.0058, 0.9780] \\
Expanding XGBoost-17 minus OLS-17 & -4.8111 & 0.2389 & 0.9557 & [-15.1504, 0.1369] & [-15.2069, 0.1156] \\
17-feature history interaction & -1.4352 & 0.7301 & 0.9557 & [-10.6977, 7.4151] & [-9.5003, 5.3974] \\
17- minus 18-feature interaction & -0.0387 & 0.4594 & 0.9557 & [-0.1742, 0.0636] & [-0.1873, 0.0604] \\
\end{longtable}
\noindent\footnotesize All estimates use the same 21 instruments and 1,715 evaluation dates as the original supplied-18-feature comparison. The negative-return feature is removed without tuning another setting. These four exploratory tests form their own Holm family separately under each loss; they do not alter the original primary tests. Intervals are non-circular, uncentered, pointwise percentile intervals on synchronized date blocks. The change row subtracts the original 18-feature interaction from the new 17-feature interaction.\normalsize
\endgroup

The authors reconstructed the unclipped OLS-17 and OLS-18 log predictions over all 42 histories and reconciled the reconstructed raw forecasts with the saved forecasts. The maximum absolute differences were $4.84\times10^{-13}$ for unclipped log predictions, $2.79\times10^{-13}$ for raw forecast levels, $6.75\times10^{-14}$ for smearing factors, and $3.65\times10^{-13}$ for final variance forecasts. The replication records give the instrument, history and date attaining each maximum. These checks concern predictions in the matched saved sample and do not imply equality of minimum-norm coefficients across parameterizations.

\subsection{Exploratory post-2020 evaluation}

These analyses were added after inspection of the primary temporal profile. They hold every model's saved forecasts fixed; they do not retrain models or delete 2020 from the primary evaluation. Table~\ref{app:post2020-levels} retains all principal procedures and the same-target fixed EWMA reference. Table~\ref{app:post2020-interactions} reports the four matched history interactions, with Holm adjustment across those four comparisons separately at each HAC lag. The windows overlap, so the estimates are dependent exploratory summaries rather than independent replications.

\reservetable{470pt}
\begingroup\papertablesize\setlength{\tabcolsep}{2pt}
\begin{longtable}{@{}L{.15\linewidth}L{.42\linewidth}R{.20\linewidth}R{.18\linewidth}@{}}
\caption{Exploratory post-2020 absolute loss levels on the five-session target}\label{app:post2020-levels}\\
\toprule History & Forecast procedure & Mean QLIKE & \shortstack[r]{Mean MSE\\($\times10^{-3}$)} \\ \midrule
\endfirsthead
\multicolumn{4}{c}{Table \thetable{} -- continued}\\
\toprule History & Forecast procedure & Mean QLIKE & \shortstack[r]{Mean MSE\\($\times10^{-3}$)} \\ \midrule\endhead
\midrule\multicolumn{4}{r}{Continued on next page}\\\endfoot
\bottomrule\endlastfoot
\multicolumn{4}{@{}l}{\textit{2021--2026: 1462 common dates, 21 instruments}}\\\addlinespace
Rolling & HAR (smeared) & -3.2758 & 1.9110 \\
Rolling & OLS-18 (smeared) & -3.2874 & 1.7989 \\
Rolling & XGBoost-3 (smeared) & -3.2456 & 2.2331 \\
Rolling & XGBoost-18 (smeared) & -3.2575 & 2.1076 \\
Rolling & DirectQLIKE-18 & -3.2876 & 1.8258 \\
Expanding & HAR (smeared) & -3.2806 & 1.8799 \\
Expanding & OLS-18 (smeared) & -3.2921 & 1.8972 \\
Expanding & XGBoost-3 (smeared) & -3.2722 & 1.8883 \\
Expanding & XGBoost-18 (smeared) & -3.2829 & 1.7836 \\
Expanding & DirectQLIKE-18 & -3.2913 & 1.9184 \\
Fixed rule & EWMA-GK & -3.2615 & 2.0338 \\
\multicolumn{4}{@{}l}{\textit{2023--2026: 959 common dates, 21 instruments}}\\\addlinespace
Rolling & HAR (smeared) & -3.3565 & 2.5799 \\
Rolling & OLS-18 (smeared) & -3.3710 & 2.3719 \\
Rolling & XGBoost-3 (smeared) & -3.3324 & 2.8874 \\
Rolling & XGBoost-18 (smeared) & -3.3456 & 2.7271 \\
Rolling & DirectQLIKE-18 & -3.3721 & 2.4241 \\
Expanding & HAR (smeared) & -3.3600 & 2.5477 \\
Expanding & OLS-18 (smeared) & -3.3728 & 2.5643 \\
Expanding & XGBoost-3 (smeared) & -3.3528 & 2.5549 \\
Expanding & XGBoost-18 (smeared) & -3.3642 & 2.3864 \\
Expanding & DirectQLIKE-18 & -3.3730 & 2.5981 \\
Fixed rule & EWMA-GK & -3.3362 & 2.7663 \\
\end{longtable}
\noindent\footnotesize QLIKE denotes quasi-likelihood loss and MSE denotes mean squared error; lower values are better. Losses are averaged across the same 21 instruments within each date, then equally across dates. Multiply the MSE entries by $10^{-3}$ to recover squared annualized-variance units. HAR denotes heterogeneous autoregression; OLS denotes ordinary least squares; XGBoost denotes extreme gradient boosting; suffixes give predictor counts. Log-target models use sequential smearing. DirectQLIKE-18 estimates a positive log-link forecast by minimizing variance-scale QLIKE. EWMA-GK denotes the fixed-rule exponentially weighted moving average of daily Garman--Klass variance, projected to the same five-session target. These loss levels are descriptive; paired inference is reported separately.\normalsize
\endgroup

\reservetable{250pt}
\begingroup\papertablesize\setlength{\tabcolsep}{2pt}
\begin{longtable}{@{}L{.29\linewidth}R{.10\linewidth}R{.07\linewidth}R{.075\linewidth}R{.075\linewidth}L{.16\linewidth}L{.16\linewidth}@{}}
\caption{Exploratory history interactions after omitting 2020}\label{app:post2020-interactions}\\
\toprule Matched comparison & Mean & \shortstack[r]{Holm\\HAC(4)\\$p$} & \shortstack[r]{Holm\\HAC(21)\\$p$} & \shortstack[r]{Holm\\HAC(43)\\$p$} & \shortstack[l]{MBB-22 95\%\\confidence interval} & \shortstack[l]{MBB-44 95\%\\confidence interval} \\ \midrule
\endfirsthead
\multicolumn{7}{c}{Table \thetable{} -- continued}\\
\toprule Matched comparison & Mean & \shortstack[r]{Holm\\HAC(4)\\$p$} & \shortstack[r]{Holm\\HAC(21)\\$p$} & \shortstack[r]{Holm\\HAC(43)\\$p$} & \shortstack[l]{MBB-22 95\%\\confidence interval} & \shortstack[l]{MBB-44 95\%\\confidence interval} \\ \midrule\endhead
\midrule\multicolumn{7}{r}{Continued on next page}\\\endfoot
\bottomrule\endlastfoot
\multicolumn{7}{@{}l}{\textit{2021--2026: 1462 common dates}}\\\addlinespace
XGBoost-18 versus OLS-18 & -0.0207 & $<0.0001$ & 0.0001 & 0.0006 & [-0.0310, -0.0111] & [-0.0319, -0.0103] \\
XGBoost-3 versus HAR & -0.0218 & $<0.0001$ & $<0.0001$ & $<0.0001$ & [-0.0296, -0.0139] & [-0.0301, -0.0132] \\
\multicolumn{7}{@{}l}{\textit{2023--2026: 959 common dates}}\\\addlinespace
XGBoost-18 versus OLS-18 & -0.0167 & 0.0011 & 0.0060 & 0.0088 & [-0.0308, -0.0067] & [-0.0313, -0.0063] \\
XGBoost-3 versus HAR & -0.0168 & 0.0005 & 0.0004 & 0.0006 & [-0.0271, -0.0092] & [-0.0273, -0.0094] \\
\end{longtable}
\noindent\footnotesize Each interaction is (expanding model minus matched linear comparator) minus (rolling model minus comparator), using quasi-likelihood loss. These are post-hoc analyses of the frozen forecasts, not new model fits or members of the original principal test family. HAC denotes heteroskedasticity and autocorrelation consistent inference, with the lag in parentheses. The three test columns report Holm-adjusted $p$-values across the four window-by-learner interactions, separately at each lag.  MBB denotes moving-block bootstrap; intervals use 5,000 synchronized non-circular draws with the indicated block length and seed 1729. Intervals are pointwise and condition on saved forecast paths.\normalsize
\endgroup

\reservetable{300pt}
\subsection{Restricted WRDS holdings sample}\label{app:holdings-reproduction}

The input holdings sample was obtained through the institution's FTSE Russell sample entitlement, not by downloading present-day constituents and carrying them backward. The nonlicensed source manifest records the table name, archive hashes and linkage configuration. Each observed eight-character CUSIP is matched to CRSP name history with \texttt{NAMEDT}$\leq$roster date$\leq$\texttt{NAMEENDDT}; all 20,320 roster observations have a unique date-valid link. Multiple historical CUSIPs can identify one PERMNO. The available sample contains 43 securities with supplied delisting-event records and is not restricted to firms surviving at the final date.

For a snapshot dated in month $m$, only sessions in $m+1$ are candidate issuance dates. There is no forward carry into $m+2$ without a new snapshot. The absent April--August 2017 snapshots therefore produce no candidate rosters in May--September. This policy assumes that the previous month-end roster was available at the first next-month session; actual vendor publication timestamps and exact daily index entries/exits are not available. We make no claim of authenticated daily point-in-time index membership. A security exits the candidate set when its monthly roster expires; no new forecasts are issued if its subsequent roster is absent. Missing prices or insufficient training history can also make a candidate unavailable. A security with fewer than 1,008 eligible mature labels at that month's refit receives no forecast during that month, even if it reaches the minimum later in the month.

Prices are aligned to the complete observed SPY session calendar from 2 January 2002 through 31 October 2017. Missing source rows remain missing; sessions are neither compressed nor forward-filled. The GK inputs are the native open (\texttt{OPENPRC}), high/ask (\texttt{ASKHI}), low/bid (\texttt{BIDLO}) and close/quote (\texttt{PRC}) fields. These field definitions and numerical screens do not independently authenticate a transaction-price range. All four must be finite and positive, and the high must be at least the open, close and low and the low at most the open, close and high, with numerical tolerance $10^{-8}$. Negative CRSP closing-price quote indicators do not become negative trade prices or valid GK ranges. Valid zero ranges and large positive ranges are retained. One inconsistent BIIB bar on 21 September 2009 is unavailable in both versions. There is no general crisis-date deletion, winsorization or outcome-based range cap.

The export's missing cells are read as missing values. Native \texttt{RET} and \texttt{RETX}, the one-session holding returns including and excluding distributions respectively, are used for checks only when finite and at least $-1$; they are not substituted for prices or range targets. Positive trade closes divided by the positive cumulative price-adjustment factor \texttt{CFACPR} form the adjusted price-return predictors, with no return bridging across a missing session. Applying the same factor to all four OHLC prices leaves within-bar GK ratios unchanged. Reported volume is retained. Ordinary distributions and delisting cash flows are not added to these adjusted-price log returns or to the within-session GK target. In the supplied delisting records, payment dates follow final trading dates; eventual payoffs are not silently booked as ordinary last-day price returns.

Size is $|\mathrm{PRC}|\times\mathrm{SHROUT}$, where SHROUT is shares outstanding in the native source units; liquidity is the 22-session trailing mean of $|\mathrm{PRC}|\times\mathrm{VOL}$, where VOL is reported share volume. Volatility is the 22-session trailing mean GK, recomputed for each data version. Within-date terciles are formed on available proxies, with sectors from the lagged roster. The price/volume scale is common within each cross-section and does not affect these rank groups. Missing group proxies are not assigned arbitrary groups. The accompanying full-calendar census reports issued and scorable counts, including zero-roster sessions. The final scored period has 1,860 dates, while its full 1,967-session calendar retains 107 no-roster gaps in inference. The estimand is the equally weighted mean over the $n=1{,}860$ observed eligible dates, $\bar D=n^{-1}\sum_t m_tD_t$, where $m_t=1$ only when the contrast is observed. For HAC inference on the full 1,967-session calendar, the centered influence series is $e_t=m_t(D_t-\bar D)$; missing sessions contribute zero influence, not a zero forecast loss. The Bartlett long-run variance estimate is $\widehat\Omega_q=\sum_t e_t^2+2\sum_{\ell=1}^q(1-\ell/(q+1))\sum_{t=\ell+1}^T e_te_{t-\ell}$ and $\widehat{\mathrm{Var}}(\bar D)=\widehat\Omega_q/n^2$. Lags count trading sessions on this complete calendar, with no pair-count renormalization or finite-sample correction. The synchronized block bootstrap also retains the full calendar and divides each replicate by its observed eligible-date count; the observation mask and losses are resampled jointly. Draws with zero observed dates are omitted. Zero centered influence at missing dates does not imply that the roster gaps are random.

\textbf{Three-observation source audit.} Table~\ref{app:source-bars} identifies the exact native and archived alternative bars. The native lows (0.0001 for EXC, 0.1400 for ACN and 2.0000 for PM) imply unusually large within-session ranges. A joint report of the U.S. Commodity Futures Trading Commission (CFTC) and U.S. Securities and Exchange Commission (SEC) independently identifies these securities among stocks with broken trades on 6 May 2010 \citep{sec_cftc2010}. The targeted audit was triggered by the extreme native ranges and the prior CFTC/SEC report; it was not a fully mechanical screen of every daily OHLC bar. This motivates the source audit; it does not authenticate a replacement daily bar. The archived alternative response supplies open, high, low and close from one quote object for each substituted observation, so each OHLC row is internally sourced from one response. Its visibly different EXC scale remains unexplained because corporate-action adjustment metadata were not archived. Vendor coverage, price bases and the treatment of broken trades may differ. We retain the neutral term \emph{alternative-source sensitivity} throughout.

For each affected series, the substitution changes five targets with origins 29 April--5 May 2010 and 22 feature vectors dated 6 May--7 June 2010. Effects can then persist through model estimation and the 504-error rolling calibration histories. For EXC, the first affected fitted-model refit is 1 June 2010, while its calibration factor first differs on 6 May. Its rolling OLS smearing differences persist to 6 July 2016; expanding differences persist to the source end. These are propagated fitted/calibration effects, not additional raw observations being substituted. The code bundle retains the full per-security, history, target, feature, refit and calibration audit.

For EXC, ACN and PM, the Check results were generated by rerunning the complete five-learner pipeline under both histories: targets, features, refits, validation and sequential calibration were recomputed. The other 310 securities have identical input channels and use verified identical-input aliases. The saved six affected Check job records and code/input hashes verify this distinction; the Check is not a post-hoc alteration of forecast loss cells.

Table~\ref{app:source-exclusions} adds exploratory checks specified before their calculation. The direct-footprint check removes evaluation cells for the three securities whose target or 22-session feature window touches the disputed bars (the interval 29 April--7 June 2010). There are 27 eligible paired cells in that footprint; not every affected security is forecast in that contemporaneous roster. The broader check removes all evaluation cells for EXC, ACN and PM. These are evaluation exclusions of saved forecasts, not new clean-input refits: the direct-footprint version can retain later fitted/calibration influence. The native and Check series coincide after removal of all three securities, as expected from own-security predictors. Original results remain unchanged, and no claim of broad cross-security corroboration is based on whichever source view gives a favorable ranking.

An author-run recalculation on authorized local records reconciled the original historical contrasts and both sets of exclusions over the 1,967-date calendar, including the 107 no-roster sessions. The original rows' means, HAC(4) $p$-values and percentile endpoints agreed within $4.13\times10^{-16}$. The replication package includes permitted source-lineage records and exclusion results. The underlying licensed security-level files remain in the authorized workspace.

The permitted alias verification independently compares both daily variance columns for each unaffected security. The audit reports the aggregate counts and a checksum of the local alias roster.

The observed FTSE Russell holdings sample contains 104 monthly snapshots from September 2008 through September 2017, with five snapshots missing in 2017, and links to 313 PERMNOs. The data versions and stratum definitions are described in Section~\ref{sec:holdings-data}. The table label \emph{CRSP} denotes native observations; \emph{Check} denotes the three-observation Yahoo-based sensitivity. Lower, middle, and upper groups are within-date terciles (thirds) of the size, liquidity, or volatility proxy. Size uses market capitalization; liquidity uses trailing 22-session mean dollar volume; volatility uses the trailing 22-session mean range variance. Sector assignments come from the lagged holdings roster. Security-dates counts the contributing security-date observations. The mean is the expanding-minus-rolling QLIKE loss difference relative to the matched linear comparator.

\begingroup\papertablesize\setlength{\tabcolsep}{2pt}
\begin{longtable}{@{}L{.08\linewidth}L{.10\linewidth}R{.13\linewidth}R{.13\linewidth}R{.13\linewidth}R{.13\linewidth}R{.14\linewidth}@{}}
\caption{Native and archived alternative OHLC bars on 6 May 2010}\label{app:source-bars}\\
\toprule Ticker & Version & Open & High & Low & Close & Daily GK \\ \midrule
\endfirsthead
\multicolumn{7}{c}{Table \thetable{} -- continued}\\
\toprule Ticker & Version & Open & High & Low & Close & Daily GK \\ \midrule
\endhead
\midrule\multicolumn{7}{r}{Continued on next page}\\\endfoot
\bottomrule\endlastfoot
EXC & CRSP & 43.3500 & 43.3500 & 0.0001 & 41.8600 & 84.2351 \\
EXC & Check & 30.9201 & 30.9201 & 20.1141 & 29.8573 & 0.0920 \\
ACN & CRSP & 41.9400 & 42.3000 & 0.1400 & 41.0900 & 16.3070 \\
ACN & Check & 41.9400 & 42.3000 & 38.7500 & 41.0900 & 0.0037 \\
PM & CRSP & 48.5800 & 48.9200 & 2.0000 & 47.0000 & 5.1101 \\
PM & Check & 48.5000 & 48.9100 & 44.0000 & 47.0000 & 0.0052 \\
\end{longtable}
\noindent\footnotesize EXC, ACN and PM identify Exelon, Accenture and Philip Morris International, respectively; their PERMNOs are 21776, 89071 and 92602. The raw vendor price bases are displayed. GK depends on within-bar ratios and is invariant to a common price scaling. The Check substitutes only these daily GK values, retaining native return and volume channels. The archived alternative responses and hashes are identified in the source audit; the bars are not authenticated exchange corrections.\normalsize
\endgroup

\begingroup\papertablesize\setlength{\tabcolsep}{2pt}
\begin{longtable}{@{}L{.07\linewidth}L{.10\linewidth}L{.12\linewidth}L{.14\linewidth}R{.09\linewidth}R{.09\linewidth}R{.10\linewidth}L{.20\linewidth}@{}}
\caption{Exploratory evaluation exclusions in the historical sample}\label{app:source-exclusions}\\
\toprule Version & Learner & Comparison & Exclusion & \shortstack[r]{Removed\\cells} & Mean & \shortstack[r]{HAC(4)\\$p$-value} & MBB-22 95\% interval \\ \midrule
\endfirsthead
\multicolumn{8}{c}{Table \thetable{} -- continued}\\
\toprule Version & Learner & Comparison & Exclusion & \shortstack[r]{Removed\\cells} & Mean & \shortstack[r]{HAC(4)\\$p$-value} & MBB-22 95\% interval \\ \midrule
\endhead
\midrule\multicolumn{8}{r}{Continued on next page}\\\endfoot
\bottomrule\endlastfoot
CRSP & XGBoost-18 & Rolling level & Original & 0 & -0.1010 & 0.3223 & [-0.3435, 0.0172] \\
CRSP & XGBoost-18 & Rolling level & Direct footprint & 27 & 0.0112 & 0.0001 & [0.0051, 0.0179] \\
CRSP & XGBoost-18 & Rolling level & Three securities & 4,848 & 0.0107 & 0.0003 & [0.0046, 0.0174] \\
CRSP & XGBoost-18 & Expanding level & Original & 0 & -0.0917 & 0.3516 & [-0.3112, 0.0167] \\
CRSP & XGBoost-18 & Expanding level & Direct footprint & 27 & 0.0081 & 0.0007 & [0.0029, 0.0141] \\
CRSP & XGBoost-18 & Expanding level & Three securities & 4,848 & 0.0075 & 0.0016 & [0.0025, 0.0135] \\
CRSP & XGBoost-18 & Interaction & Original & 0 & 0.0093 & 0.7237 & [-0.0219, 0.0517] \\
CRSP & XGBoost-18 & Interaction & Direct footprint & 27 & -0.0032 & 0.0217 & [-0.0061, -0.0001] \\
CRSP & XGBoost-18 & Interaction & Three securities & 4,848 & -0.0032 & 0.0221 & [-0.0062, -0.0002] \\
CRSP & XGBoost-3 & Rolling level & Original & 0 & -0.1309 & 0.2959 & [-0.4401, 0.0180] \\
CRSP & XGBoost-3 & Rolling level & Direct footprint & 27 & 0.0118 & $<0.0001$ & [0.0053, 0.0188] \\
CRSP & XGBoost-3 & Rolling level & Three securities & 4,848 & 0.0115 & 0.0001 & [0.0049, 0.0184] \\
CRSP & XGBoost-3 & Expanding level & Original & 0 & -0.2767 & 0.2381 & [-0.8876, 0.0125] \\
CRSP & XGBoost-3 & Expanding level & Direct footprint & 27 & 0.0064 & 0.0220 & [0.0004, 0.0134] \\
CRSP & XGBoost-3 & Expanding level & Three securities & 4,848 & 0.0061 & 0.0292 & [0.0001, 0.0130] \\
CRSP & XGBoost-3 & Interaction & Original & 0 & -0.1459 & 0.1968 & [-0.4447, -0.0039] \\
CRSP & XGBoost-3 & Interaction & Direct footprint & 27 & -0.0054 & $<0.0001$ & [-0.0076, -0.0029] \\
CRSP & XGBoost-3 & Interaction & Three securities & 4,848 & -0.0054 & $<0.0001$ & [-0.0076, -0.0029] \\
Check & XGBoost-18 & Rolling level & Original & 0 & 0.0107 & 0.0004 & [0.0044, 0.0174] \\
Check & XGBoost-18 & Rolling level & Direct footprint & 27 & 0.0108 & 0.0003 & [0.0047, 0.0175] \\
Check & XGBoost-18 & Rolling level & Three securities & 4,848 & 0.0107 & 0.0003 & [0.0046, 0.0174] \\
Check & XGBoost-18 & Expanding level & Original & 0 & 0.0075 & 0.0019 & [0.0023, 0.0136] \\
Check & XGBoost-18 & Expanding level & Direct footprint & 27 & 0.0077 & 0.0013 & [0.0026, 0.0136] \\
Check & XGBoost-18 & Expanding level & Three securities & 4,848 & 0.0075 & 0.0016 & [0.0025, 0.0135] \\
Check & XGBoost-18 & Interaction & Original & 0 & -0.0032 & 0.0219 & [-0.0061, -0.0001] \\
Check & XGBoost-18 & Interaction & Direct footprint & 27 & -0.0032 & 0.0219 & [-0.0061, -0.0002] \\
Check & XGBoost-18 & Interaction & Three securities & 4,848 & -0.0032 & 0.0221 & [-0.0062, -0.0002] \\
Check & XGBoost-3 & Rolling level & Original & 0 & 0.0114 & 0.0001 & [0.0046, 0.0184] \\
Check & XGBoost-3 & Rolling level & Direct footprint & 27 & 0.0116 & 0.0001 & [0.0050, 0.0185] \\
Check & XGBoost-3 & Rolling level & Three securities & 4,848 & 0.0115 & 0.0001 & [0.0049, 0.0184] \\
Check & XGBoost-3 & Expanding level & Original & 0 & 0.0058 & 0.0466 & [-0.0007, 0.0129] \\
Check & XGBoost-3 & Expanding level & Direct footprint & 27 & 0.0062 & 0.0275 & [0.0001, 0.0130] \\
Check & XGBoost-3 & Expanding level & Three securities & 4,848 & 0.0061 & 0.0292 & [0.0001, 0.0130] \\
Check & XGBoost-3 & Interaction & Original & 0 & -0.0056 & $<0.0001$ & [-0.0078, -0.0032] \\
Check & XGBoost-3 & Interaction & Direct footprint & 27 & -0.0054 & $<0.0001$ & [-0.0076, -0.0029] \\
Check & XGBoost-3 & Interaction & Three securities & 4,848 & -0.0054 & $<0.0001$ & [-0.0076, -0.0029] \\
\end{longtable}
\noindent\footnotesize Removed cells are instrument-date evaluation cells.  Level comparisons are XGBoost-18 minus OLS-18 or XGBoost-3 minus HAR within the named history. The interaction is expanding minus rolling paired loss. Direct footprint removes eligible cells whose target or 22-session GK feature footprint touches the three questioned bars; Three securities removes all evaluated cells for EXC, ACN and PM. These saved-forecast evaluation exclusions retain later fitting/calibration influence in the direct-footprint case. Every estimate uses 1,860 observed dates and preserves the 107 missing-roster sessions in the full calendar for inference. New $p$-values are unadjusted exploratory HAC(4) values; intervals are pointwise. Original families and samples remain unchanged.\normalsize
\endgroup

\reservetable{520pt}
\begingroup\papertablesize\setlength{\tabcolsep}{2pt}
\begin{longtable}{@{}L{.09\linewidth}L{.36\linewidth}R{.10\linewidth}R{.08\linewidth}R{.08\linewidth}L{.23\linewidth}@{}}
\caption{Native CRSP and Yahoo-check historical holdings-sample QLIKE comparisons}\label{app:crsp-overview}\\
\toprule \shortstack[l]{Data\\version} & Contrast & Mean & \shortstack[r]{HAC(4)\\$p$-value} & \shortstack[r]{Holm\\$p$-value} & \shortstack[l]{MBB-22 95\%\\confidence interval} \\ \midrule\endfirsthead
\multicolumn{6}{c}{Table \thetable{} -- continued}\\
\toprule \shortstack[l]{Data\\version} & Contrast & Mean & \shortstack[r]{HAC(4)\\$p$-value} & \shortstack[r]{Holm\\$p$-value} & \shortstack[l]{MBB-22 95\%\\confidence interval} \\ \midrule\endhead
\midrule\multicolumn{6}{r}{Continued on next page}\\\endfoot
\bottomrule\endlastfoot
CRSP & rolling XGBoost-18 minus OLS-18 smeared & -0.1010 & 0.3223 & 1.0000 & [-0.3435, 0.0172] \\
CRSP & expanding XGBoost-18 minus OLS-18 smeared & -0.0917 & 0.3516 & 1.0000 & [-0.3112, 0.0167] \\
CRSP & rolling XGBoost-3 minus HAR smeared & -0.1309 & 0.2959 & 1.0000 & [-0.4401, 0.0180] \\
CRSP & expanding XGBoost-3 minus HAR smeared & -0.2767 & 0.2381 & 1.0000 & [-0.8876, 0.0125] \\
CRSP & expanding XGBoost-18 smear minus raw & -0.6275 & 0.2395 & 1.0000 & [-2.0133, 0.0256] \\
CRSP & expanding XGBoost-18 smeared minus HAR smeared & -0.1964 & 0.2291 & 1.0000 & [-0.6147, 0.0028] \\
CRSP & expanding OLS-18 smeared minus HAR smeared & -0.1046 & 0.2610 & 1.0000 & [-0.3143, -0.0056] \\
CRSP & expanding DirectQLIKE-18 minus OLS-18 smeared & 0.0094 & 0.6267 & 1.0000 & [-0.0208, 0.0313] \\
CRSP & XGBoost-18 smeared history interaction & 0.0093 & 0.7237 & 1.0000 & [-0.0219, 0.0517] \\
CRSP & XGBoost-3 smeared history interaction & -0.1459 & 0.1968 & 1.0000 & [-0.4447, -0.0039] \\
Check & rolling XGBoost-18 minus OLS-18 smeared & 0.0107 & 0.0004 & 0.0065 & [0.0044, 0.0174] \\
Check & expanding XGBoost-18 minus OLS-18 smeared & 0.0075 & 0.0019 & 0.0308 & [0.0023, 0.0136] \\
Check & rolling XGBoost-3 minus HAR smeared & 0.0114 & 0.0001 & 0.0025 & [0.0046, 0.0184] \\
Check & expanding XGBoost-3 minus HAR smeared & 0.0058 & 0.0466 & 0.6521 & [-0.0007, 0.0129] \\
Check & expanding XGBoost-18 smear minus raw & 0.0045 & 0.4291 & 1.0000 & [-0.0107, 0.0160] \\
Check & expanding XGBoost-18 smeared minus HAR smeared & -0.0023 & 0.2884 & 1.0000 & [-0.0076, 0.0027] \\
Check & expanding OLS-18 smeared minus HAR smeared & -0.0098 & $<0.0001$ & 0.0002 & [-0.0147, -0.0061] \\
Check & expanding DirectQLIKE-18 minus OLS-18 smeared & -0.0012 & 0.7112 & 1.0000 & [-0.0101, 0.0050] \\
Check & XGBoost-18 smeared history interaction & -0.0032 & 0.0219 & 0.3290 & [-0.0061, -0.0001] \\
Check & XGBoost-3 smeared history interaction & -0.0056 & $<0.0001$ & $<0.0001$ & [-0.0078, -0.0032] \\
\end{longtable}
\noindent\footnotesize CRSP denotes native open, high, low, and close observations; Check denotes the three-observation Yahoo-based sensitivity. Both price-data versions use 1,860 evaluation dates and the observed lagged holdings rosters. The Check version changes three Garman--Klass observations on 6 May 2010 using archived Yahoo Finance bars; the Check version is not asserted to be an authenticated exchange correction. Negative differences favor the first procedure. HAC(4) uses unadjusted heteroskedasticity and autocorrelation consistent tests with four lags; Holm adjusts the 20 QLIKE contrasts across both versions as one family. MBB-22 denotes 5,000 synchronized 22-date moving-block bootstrap draws.  Intervals are pointwise. This restricted historical sample is not full Russell 3000 membership.\normalsize
\endgroup

\begingroup
\setlength{\tabcolsep}{2pt}
\papertablesize
\begin{longtable}{@{}L{.075\linewidth}L{.102\linewidth}L{.105\linewidth}L{.155\linewidth}R{.095\linewidth}R{.098\linewidth}R{.10\linewidth}L{.20\linewidth}@{}}
\caption{Exploratory QLIKE interactions by stratum in the restricted historical holdings sample}\label{app:crsp-strata}\\
\toprule \shortstack[l]{Data\\version} & Learner & Dimension & Group & \shortstack[r]{Mean\\($\times10^{-2}$)} & \shortstack[r]{Security-\\dates} & \shortstack[r]{HAC(4)\\$p$-value} & \shortstack[l]{MBB-22 95\%\\confidence interval\\($\times10^{-2}$)} \\ \midrule
\endfirsthead
\multicolumn{8}{c}{\tablename\ \thetable{} -- continued}\\
\toprule \shortstack[l]{Data\\version} & Learner & Dimension & Group & \shortstack[r]{Mean\\($\times10^{-2}$)} & \shortstack[r]{Security-\\dates} & \shortstack[r]{HAC(4)\\$p$-value} & \shortstack[l]{MBB-22 95\%\\confidence interval\\($\times10^{-2}$)} \\ \midrule
\endhead
\midrule\multicolumn{8}{r}{Continued on next page}\\\endfoot
\bottomrule\endlastfoot
CRSP & XGBoost-18 & Sector & Consumer Discretionary & -0.2111 & 52279 & 0.1731 & [-0.4980, 0.1013] \\
CRSP & XGBoost-18 & Sector & Consumer Staples & 0.1520 & 28101 & 0.7360 & [-0.6098, 1.2479] \\
CRSP & XGBoost-18 & Sector & Energy & -0.6045 & 30974 & 0.0064 & [-1.3642, -0.0871] \\
CRSP & XGBoost-18 & Sector & Financial Services & -0.3086 & 63123 & 0.2896 & [-0.9424, 0.3239] \\
CRSP & XGBoost-18 & Sector & Health Care & -0.2474 & 48935 & 0.0387 & [-0.5228, 0.0118] \\
CRSP & XGBoost-18 & Sector & Materials \& Processing & -0.4059 & 19759 & 0.0251 & [-0.8108, 0.0028] \\
CRSP & XGBoost-18 & Sector & Producer Durables & -0.4606 & 45715 & 0.0151 & [-0.8918, -0.0268] \\
CRSP & XGBoost-18 & Sector & Technology & -0.2790 & 38598 & 0.0096 & [-0.5421, -0.0480] \\
CRSP & XGBoost-18 & Sector & Utilities & 16.6333 & 21316 & 0.6483 & [-25.8915, 74.6499] \\
CRSP & XGBoost-18 & Size & lower & -0.4570 & 114293 & 0.0003 & [-0.7939, -0.1129] \\
CRSP & XGBoost-18 & Size & middle & 3.1566 & 116511 & 0.6761 & [-5.7252, 15.1254] \\
CRSP & XGBoost-18 & Size & upper & -0.0613 & 117996 & 0.7749 & [-0.4931, 0.3947] \\
CRSP & XGBoost-18 & Liquidity & lower & 3.3788 & 114556 & 0.6849 & [-6.3726, 16.5611] \\
CRSP & XGBoost-18 & Liquidity & middle & -0.2533 & 116738 & 0.0574 & [-0.5428, 0.0576] \\
CRSP & XGBoost-18 & Liquidity & upper & -0.1191 & 117475 & 0.5557 & [-0.5580, 0.3004] \\
CRSP & XGBoost-18 & Volatility & lower & 3.1904 & 119110 & 0.6730 & [-5.6859, 15.2407] \\
CRSP & XGBoost-18 & Volatility & middle & -0.3195 & 118141 & 0.0672 & [-0.7081, 0.0207] \\
CRSP & XGBoost-18 & Volatility & upper & -0.2163 & 111549 & 0.1547 & [-0.6254, 0.1843] \\
CRSP & XGBoost-3 & Sector & Consumer Discretionary & -0.5701 & 52279 & $<0.0001$ & [-0.8603, -0.2992] \\
CRSP & XGBoost-3 & Sector & Consumer Staples & 0.2463 & 28101 & 0.6655 & [-0.6798, 1.7927] \\
CRSP & XGBoost-3 & Sector & Energy & -0.6787 & 30974 & 0.0003 & [-1.2027, -0.2658] \\
CRSP & XGBoost-3 & Sector & Financial Services & -0.5151 & 63123 & 0.0038 & [-0.9509, -0.0855] \\
CRSP & XGBoost-3 & Sector & Health Care & -0.5949 & 48935 & $<0.0001$ & [-0.8904, -0.3189] \\
CRSP & XGBoost-3 & Sector & Materials \& Processing & -0.5313 & 19759 & $<0.0001$ & [-0.8312, -0.2074] \\
CRSP & XGBoost-3 & Sector & Producer Durables & -0.7122 & 45715 & $<0.0001$ & [-1.0776, -0.3540] \\
CRSP & XGBoost-3 & Sector & Technology & -0.4501 & 38598 & $<0.0001$ & [-0.6680, -0.2187] \\
CRSP & XGBoost-3 & Sector & Utilities & -196.7003 & 21316 & 0.2128 & [-614.5283, -0.4254] \\
CRSP & XGBoost-3 & Size & lower & -0.7269 & 114293 & $<0.0001$ & [-1.0264, -0.4217] \\
CRSP & XGBoost-3 & Size & middle & -41.1829 & 116511 & 0.2060 & [-127.2240, -0.6068] \\
CRSP & XGBoost-3 & Size & upper & -0.0648 & 117996 & 0.6910 & [-0.4480, 0.3868] \\
CRSP & XGBoost-3 & Liquidity & lower & -45.5287 & 114556 & 0.2063 & [-140.7378, -0.5829] \\
CRSP & XGBoost-3 & Liquidity & middle & -0.5267 & 116738 & $<0.0001$ & [-0.8097, -0.2174] \\
CRSP & XGBoost-3 & Liquidity & upper & -0.1831 & 117475 & 0.2017 & [-0.5117, 0.1982] \\
CRSP & XGBoost-3 & Volatility & lower & -40.6994 & 119110 & 0.2105 & [-126.6829, -0.1469] \\
CRSP & XGBoost-3 & Volatility & middle & -0.5523 & 118141 & $<0.0001$ & [-0.8133, -0.2999] \\
CRSP & XGBoost-3 & Volatility & upper & -0.7363 & 111549 & $<0.0001$ & [-1.1003, -0.3752] \\
Check & XGBoost-18 & Sector & Consumer Discretionary & -0.2111 & 52279 & 0.1731 & [-0.4980, 0.1013] \\
Check & XGBoost-18 & Sector & Consumer Staples & 0.1554 & 28101 & 0.7315 & [-0.6196, 1.2538] \\
Check & XGBoost-18 & Sector & Energy & -0.6045 & 30974 & 0.0064 & [-1.3642, -0.0871] \\
Check & XGBoost-18 & Sector & Financial Services & -0.3086 & 63123 & 0.2896 & [-0.9424, 0.3239] \\
Check & XGBoost-18 & Sector & Health Care & -0.2474 & 48935 & 0.0387 & [-0.5228, 0.0118] \\
Check & XGBoost-18 & Sector & Materials \& Processing & -0.4059 & 19759 & 0.0251 & [-0.8108, 0.0028] \\
Check & XGBoost-18 & Sector & Producer Durables & -0.5060 & 45715 & 0.0085 & [-0.9306, -0.0838] \\
Check & XGBoost-18 & Sector & Technology & -0.2790 & 38598 & 0.0096 & [-0.5421, -0.0480] \\
Check & XGBoost-18 & Sector & Utilities & -0.6909 & 21316 & 0.0134 & [-1.2423, -0.0021] \\
Check & XGBoost-18 & Size & lower & -0.4541 & 114293 & 0.0003 & [-0.7915, -0.1103] \\
Check & XGBoost-18 & Size & middle & -0.4372 & 116511 & 0.0008 & [-0.6977, -0.1831] \\
Check & XGBoost-18 & Size & upper & -0.0624 & 117996 & 0.7722 & [-0.4984, 0.3980] \\
Check & XGBoost-18 & Liquidity & lower & -0.5967 & 114556 & $<0.0001$ & [-0.9018, -0.2682] \\
Check & XGBoost-18 & Liquidity & middle & -0.2513 & 116738 & 0.0583 & [-0.5385, 0.0538] \\
Check & XGBoost-18 & Liquidity & upper & -0.1162 & 117475 & 0.5648 & [-0.5556, 0.3031] \\
Check & XGBoost-18 & Volatility & lower & -0.4000 & 119099 & 0.0249 & [-0.7544, -0.0503] \\
Check & XGBoost-18 & Volatility & middle & -0.3300 & 118131 & 0.0583 & [-0.7164, 0.0090] \\
Check & XGBoost-18 & Volatility & upper & -0.2058 & 111570 & 0.1749 & [-0.6136, 0.1941] \\
Check & XGBoost-3 & Sector & Consumer Discretionary & -0.5701 & 52279 & $<0.0001$ & [-0.8603, -0.2992] \\
Check & XGBoost-3 & Sector & Consumer Staples & 0.2415 & 28101 & 0.6718 & [-0.6964, 1.7829] \\
Check & XGBoost-3 & Sector & Energy & -0.6787 & 30974 & 0.0003 & [-1.2027, -0.2658] \\
Check & XGBoost-3 & Sector & Financial Services & -0.5151 & 63123 & 0.0038 & [-0.9509, -0.0855] \\
Check & XGBoost-3 & Sector & Health Care & -0.5949 & 48935 & $<0.0001$ & [-0.8904, -0.3189] \\
Check & XGBoost-3 & Sector & Materials \& Processing & -0.5313 & 19759 & $<0.0001$ & [-0.8312, -0.2074] \\
Check & XGBoost-3 & Sector & Producer Durables & -0.7617 & 45715 & $<0.0001$ & [-1.1225, -0.4163] \\
Check & XGBoost-3 & Sector & Technology & -0.4501 & 38598 & $<0.0001$ & [-0.6680, -0.2187] \\
Check & XGBoost-3 & Sector & Utilities & -1.2588 & 21316 & 0.0037 & [-2.2848, -0.3244] \\
Check & XGBoost-3 & Size & lower & -0.7229 & 114293 & $<0.0001$ & [-1.0231, -0.4186] \\
Check & XGBoost-3 & Size & middle & -0.8750 & 116511 & $<0.0001$ & [-1.1985, -0.5837] \\
Check & XGBoost-3 & Size & upper & -0.0695 & 117996 & 0.6685 & [-0.4508, 0.3809] \\
Check & XGBoost-3 & Liquidity & lower & -0.9930 & 114556 & $<0.0001$ & [-1.4332, -0.5707] \\
Check & XGBoost-3 & Liquidity & middle & -0.5170 & 116738 & $<0.0001$ & [-0.7965, -0.2202] \\
Check & XGBoost-3 & Liquidity & upper & -0.1789 & 117475 & 0.2104 & [-0.5068, 0.2035] \\
Check & XGBoost-3 & Volatility & lower & -0.4036 & 119099 & 0.0004 & [-0.6519, -0.0774] \\
Check & XGBoost-3 & Volatility & middle & -0.5490 & 118131 & $<0.0001$ & [-0.8099, -0.2958] \\
Check & XGBoost-3 & Volatility & upper & -0.7264 & 111570 & $<0.0001$ & [-1.0865, -0.3651] \\
\end{longtable}
\endgroup
\noindent\footnotesize The mean and both interval bounds use the common column scale $10^{-2}$. There are 72 exploratory subgroup estimates across two price-data versions (CRSP: native observations; Check: the three-observation Yahoo-based sensitivity), two learner contrasts, and size, liquidity, volatility, and sector groups. HAC(4) $p$-values are unadjusted across these exploratory strata.  This sample is not full Russell 3000 membership.\normalsize

\reservetable{220pt}
\subsection{Idealized volatility-targeting illustration}\label{app:portfolio-methods}

This is an \emph{idealized same-close accounting illustration}, not an executable trading strategy. Forecasts use information through close $t$, and the calculation assumes that the resulting target weight can be established at that same close for the return from $t$ to $t+1$. No closing-auction order cutoff, execution delay, opening trade, spread or market impact is modeled. An implementable next-open or delayed-close strategy would require different return alignment and a new analysis; these results are not presented as that analysis.

The separately fitted SPY target is $Y^{CC}_{t+1}=252r_{t+1}^{2}$, where $r$ is the adjusted-close log return. The squared log return is the specified one-session variance proxy, consistent with the log-return feature set. For portfolio wealth, it is converted to a simple return by $\exp(r)-1$ so wealth compounds on a simple-return basis; the adjusted-close series is the chosen return proxy, not an observed executable transaction price. The feature vector is the same past-only GK/return/volume vector in Section~\ref{sec:data}; it is not a shifted five-session forecast. HAR uses its three trailing GK means as predictors of this one-session close-to-close outcome. OLS-18 and both XGBoost specifications are refit to the floored log of that outcome, and DirectQLIKE-18 minimizes the positive log-link QLIKE objective for that outcome. The target floor is 252 times one millionth of the median of the first 252 positive observed squared close-to-close returns; its availability date is 2 January 2003. It differs from the GK floor. All histories require 1,008 matured one-session labels, $s+1\leq t$, and refit monthly from 2018. The archived XGBoost implementation retains its conservative five-origin inner-validation purge even for this one-session target. Sequential smearing uses up to 504 own one-session errors with a 126-error minimum and $s+1\leq t$ maturity. DirectQLIKE is not smeared.

The portfolio GARCH/GJR forecasts use Student-$t$ return models and the conditional second moment $252(\widehat q_{t+1}+\widehat\mu^2)/10{,}000$. They refit on the same monthly schedule as this SPY application and update daily. These are direct second-moment forecasts; the separate ratio-adjustment variants in the main all-instrument GARCH exercise are not additional portfolio rows. The return-based EWMA uses decay 0.94, starts from the mean of the first 100 observed squared returns available before 2004, retains its state across missing inputs, and forecasts 252 times the updated squared-return state. It is distinct from the range-based EWMA-GK in Table~\ref{tab:absolute-loss}.

For portfolio strategy $m$, let $\widehat f^{CC}_{m,t+1}>0$ be its annualized variance forecast, $w_{m,t}$ the target SPY weight and $\tau=0.10$ the fixed annualized volatility target. With zero-yield cash and no borrowing or shorting,
\begin{equation}\label{eq:portfolio-weight}
w_{m,t}=\min\left(1,\frac{\tau}{\sqrt{\widehat f^{CC}_{m,t+1}}}\right),\qquad
g_{m,t+1}=w_{m,t}\big(\exp(r_{t+1})-1\big).
\end{equation}
Here $g$ is the gross simple portfolio return. For proportional cost rate $\kappa\in\{0,0.0005,0.0010,0.0025\}$ (0, 5, 10 and 25 basis points), the exact implementation is
\begin{equation}\label{eq:portfolio-cost}
c_{m,t}=\kappa|w_{m,t}-w_{m,t-1}|,\qquad
R^{net}_{m,t+1}=(1+g_{m,t+1})(1-c_{m,t})-1.
\end{equation}
The initial position before the first evaluated origin is zero, including for buy-and-hold; its first purchase incurs the stated cost. Buy-and-hold then has weight one. Turnover uses differences in target weights, not weights drifted by realized asset returns. This is an approximation to rebalancing costs; there is no terminal liquidation charge. Wealth compounds the net returns above. Cash yield, financing, taxes, spread and market impact are omitted.

All 16 portfolio strategies are evaluated on the common 1,719 origins from 2 January 2020 through 3 November 2026, with next-session investment returns through 4 November 2026. The intersection requires an observed next-session return and finite positive forecasts for every forecast-based strategy; warm-up rows are excluded from performance. Annualized returns, volatility, Sharpe ratios and turnover use 252 sessions per year. The reported calculation holds the target, weight cap, starting position and four cost rates fixed across forecasts. These idealized close-to-close calculations do not establish economic value for the primary within-session GK forecasts.

Table~\ref{app:portfolio-all} retains all 16 portfolio strategies under all four transaction-cost cases. A basis point (bp) is 0.01 percentage point. Cum. net denotes cumulative net return; Ann. return is annualized geometric net return; realized vol. is annualized realized volatility; Max. DD is maximum drawdown; Ann. turnover sums absolute changes in portfolio weight per year. Sharpe is the annualized mean-to-standard-deviation ratio of daily net returns under zero cash yield. HAR denotes heterogeneous autoregression, OLS denotes ordinary least squares, DirectQLIKE denotes direct quasi-likelihood estimation, GARCH denotes generalized autoregressive conditional heteroskedasticity, GJR denotes its Glosten--Jagannathan--Runkle variant, and EWMA denotes exponentially weighted moving average. Model suffixes give predictor counts. The application uses the one-session close-to-close target and is separate from the five-session GK experiment. Its return-based EWMA forecast is distinct from the main range-based EWMA-GK benchmark.

\begingroup\papertablesize\setlength{\tabcolsep}{2pt}
\begin{longtable}{@{}R{.065\linewidth}L{.235\linewidth}R{.105\linewidth}R{.105\linewidth}R{.105\linewidth}R{.08\linewidth}R{.105\linewidth}R{.115\linewidth}@{}}
\caption{Idealized 10\% volatility targeting, 2 January 2020--3 November 2026, across cost cases}\label{app:portfolio-all}\label{app:portfolio-0}\label{app:portfolio-5}\label{app:portfolio-10}\label{app:portfolio-25}\\
\toprule \shortstack[r]{Cost\\(bp)} & Portfolio strategy & \shortstack[r]{Cum.\\net} & \shortstack[r]{Ann.\\return} & \shortstack[r]{Realized\\vol.} & Sharpe & \shortstack[r]{Max.\\DD} & \shortstack[r]{Ann.\\turnover} \\ \midrule\endfirsthead
\multicolumn{8}{c}{Table \thetable{} -- continued}\\
\toprule \shortstack[r]{Cost\\(bp)} & Portfolio strategy & \shortstack[r]{Cum.\\net} & \shortstack[r]{Ann.\\return} & \shortstack[r]{Realized\\vol.} & Sharpe & \shortstack[r]{Max.\\DD} & \shortstack[r]{Ann.\\turnover} \\ \midrule\endhead
\midrule\multicolumn{8}{r}{Continued on next page}\\\endfoot
\bottomrule\multicolumn{8}{@{}p{.98\linewidth}@{}}{\footnotesize The rule targets 10\% annualized volatility, uses zero-yield cash, and caps investment weight at one. Costs equal the stated basis-point rate times the absolute change in target weight, an approximation to rebalancing costs that does not adjust prior holdings for price drift. Returns, volatility, and drawdown are percentages; annualized turnover is a multiple of capital. The Sharpe ratio uses zero cash yield. This retrospective close-to-close illustration excludes market impact, taxes, financing, and alternative cash yields and is not evidence of economic value for the main five-session GK target.}\\\endlastfoot
0 & rolling HAR & 66.67\% & 7.78\% & 9.61\% & 0.827 & -15.14\% & 19.27 \\
0 & rolling OLS-18 & 57.09\% & 6.84\% & 9.28\% & 0.760 & -15.68\% & 24.59 \\
0 & rolling XGBoost-3 & 61.35\% & 7.26\% & 9.79\% & 0.766 & -14.53\% & 20.85 \\
0 & rolling XGBoost-18 & 64.73\% & 7.59\% & 9.80\% & 0.796 & -16.13\% & 24.17 \\
0 & rolling DirectQLIKE-18 & 72.23\% & 8.30\% & 10.10\% & 0.840 & -17.05\% & 19.56 \\
0 & rolling GARCH & 78.94\% & 8.90\% & 9.93\% & 0.909 & -11.87\% & 10.70 \\
0 & rolling GJR & 76.12\% & 8.65\% & 10.01\% & 0.879 & -11.84\% & 11.35 \\
0 & expanding HAR & 69.06\% & 8.00\% & 9.75\% & 0.838 & -15.59\% & 18.31 \\
0 & expanding OLS-18 & 57.53\% & 6.89\% & 9.74\% & 0.733 & -16.54\% & 20.92 \\
0 & expanding XGBoost-3 & 66.36\% & 7.75\% & 9.88\% & 0.805 & -15.87\% & 18.09 \\
0 & expanding XGBoost-18 & 57.86\% & 6.92\% & 9.72\% & 0.737 & -18.09\% & 22.05 \\
0 & expanding DirectQLIKE-18 & 71.89\% & 8.26\% & 10.48\% & 0.810 & -17.75\% & 15.48 \\
0 & expanding GARCH & 77.90\% & 8.81\% & 9.97\% & 0.897 & -12.37\% & 9.51 \\
0 & expanding GJR & 78.48\% & 8.86\% & 10.17\% & 0.886 & -11.98\% & 10.76 \\
0 & EWMA & 76.35\% & 8.67\% & 10.35\% & 0.856 & -13.05\% & 4.89 \\
0 & SPY buy and hold & 159.59\% & 15.01\% & 19.95\% & 0.801 & -33.72\% & 0.15 \\
5 & rolling HAR & 56.06\% & 6.74\% & 9.61\% & 0.727 & -15.75\% & 19.27 \\
5 & rolling OLS-18 & 44.45\% & 5.54\% & 9.29\% & 0.627 & -16.48\% & 24.59 \\
5 & rolling XGBoost-3 & 50.27\% & 6.15\% & 9.79\% & 0.659 & -15.07\% & 20.85 \\
5 & rolling XGBoost-18 & 51.69\% & 6.30\% & 9.80\% & 0.673 & -16.19\% & 24.17 \\
5 & rolling DirectQLIKE-18 & 61.12\% & 7.24\% & 10.10\% & 0.743 & -17.74\% & 19.56 \\
5 & rolling GARCH & 72.52\% & 8.32\% & 9.93\% & 0.855 & -12.35\% & 10.70 \\
5 & rolling GJR & 69.43\% & 8.04\% & 10.01\% & 0.823 & -12.20\% & 11.35 \\
5 & expanding HAR & 58.82\% & 7.02\% & 9.75\% & 0.745 & -16.12\% & 18.31 \\
5 & expanding OLS-18 & 46.69\% & 5.78\% & 9.74\% & 0.626 & -17.44\% & 20.92 \\
5 & expanding XGBoost-3 & 56.41\% & 6.78\% & 9.88\% & 0.713 & -16.34\% & 18.09 \\
5 & expanding XGBoost-18 & 46.43\% & 5.75\% & 9.72\% & 0.624 & -19.04\% & 22.05 \\
5 & expanding DirectQLIKE-18 & 63.05\% & 7.43\% & 10.48\% & 0.736 & -18.43\% & 15.48 \\
5 & expanding GARCH & 72.22\% & 8.30\% & 9.97\% & 0.850 & -12.68\% & 9.51 \\
5 & expanding GJR & 72.05\% & 8.28\% & 10.17\% & 0.833 & -12.27\% & 10.76 \\
5 & EWMA & 73.43\% & 8.41\% & 10.35\% & 0.832 & -13.17\% & 4.89 \\
5 & SPY buy and hold & 159.46\% & 15.00\% & 19.95\% & 0.801 & -33.72\% & 0.15 \\
10 & rolling HAR & 46.14\% & 5.72\% & 9.61\% & 0.627 & -16.38\% & 19.27 \\
10 & rolling OLS-18 & 32.82\% & 4.25\% & 9.29\% & 0.495 & -17.27\% & 24.59 \\
10 & rolling XGBoost-3 & 39.95\% & 5.05\% & 9.79\% & 0.553 & -15.61\% & 20.85 \\
10 & rolling XGBoost-18 & 39.69\% & 5.02\% & 9.80\% & 0.549 & -16.26\% & 24.17 \\
10 & rolling DirectQLIKE-18 & 50.72\% & 6.20\% & 10.10\% & 0.646 & -18.43\% & 19.56 \\
10 & rolling GARCH & 66.34\% & 7.75\% & 9.94\% & 0.801 & -12.82\% & 10.70 \\
10 & rolling GJR & 63.00\% & 7.42\% & 10.01\% & 0.766 & -12.63\% & 11.35 \\
10 & expanding HAR & 49.21\% & 6.04\% & 9.75\% & 0.651 & -16.68\% & 18.31 \\
10 & expanding OLS-18 & 36.59\% & 4.68\% & 9.74\% & 0.518 & -18.33\% & 20.92 \\
10 & expanding XGBoost-3 & 47.05\% & 5.82\% & 9.88\% & 0.622 & -16.90\% & 18.09 \\
10 & expanding XGBoost-18 & 35.82\% & 4.59\% & 9.72\% & 0.511 & -19.97\% & 22.05 \\
10 & expanding DirectQLIKE-18 & 54.66\% & 6.60\% & 10.48\% & 0.663 & -19.12\% & 15.48 \\
10 & expanding GARCH & 66.72\% & 7.78\% & 9.97\% & 0.802 & -13.00\% & 9.51 \\
10 & expanding GJR & 65.85\% & 7.70\% & 10.17\% & 0.780 & -12.55\% & 10.76 \\
10 & EWMA & 70.56\% & 8.14\% & 10.35\% & 0.809 & -13.28\% & 4.89 \\
10 & SPY buy and hold & 159.33\% & 14.99\% & 19.96\% & 0.800 & -33.72\% & 0.15 \\
25 & rolling HAR & 19.98\% & 2.71\% & 9.61\% & 0.326 & -18.41\% & 19.27 \\
25 & rolling OLS-18 & 3.26\% & 0.47\% & 9.30\% & 0.097 & -20.84\% & 24.59 \\
25 & rolling XGBoost-3 & 13.05\% & 1.81\% & 9.79\% & 0.233 & -17.32\% & 20.85 \\
25 & rolling XGBoost-18 & 9.07\% & 1.28\% & 9.80\% & 0.179 & -19.11\% & 24.17 \\
25 & rolling DirectQLIKE-18 & 23.37\% & 3.13\% & 10.11\% & 0.355 & -20.48\% & 19.56 \\
25 & rolling GARCH & 49.09\% & 6.03\% & 9.94\% & 0.639 & -14.22\% & 10.70 \\
25 & rolling GJR & 45.11\% & 5.61\% & 10.01\% & 0.595 & -14.20\% & 11.35 \\
25 & expanding HAR & 23.71\% & 3.17\% & 9.74\% & 0.369 & -18.34\% & 18.31 \\
25 & expanding OLS-18 & 10.26\% & 1.44\% & 9.75\% & 0.196 & -21.73\% & 20.92 \\
25 & expanding XGBoost-3 & 22.19\% & 2.98\% & 9.88\% & 0.347 & -18.56\% & 18.09 \\
25 & expanding XGBoost-18 & 8.37\% & 1.19\% & 9.72\% & 0.170 & -23.96\% & 22.05 \\
25 & expanding DirectQLIKE-18 & 32.00\% & 4.15\% & 10.49\% & 0.441 & -21.34\% & 15.48 \\
25 & expanding GARCH & 51.26\% & 6.25\% & 9.97\% & 0.659 & -13.93\% & 9.51 \\
25 & expanding GJR & 48.56\% & 5.97\% & 10.18\% & 0.621 & -13.73\% & 10.76 \\
25 & EWMA & 62.23\% & 7.35\% & 10.35\% & 0.737 & -13.63\% & 4.89 \\
25 & SPY buy and hold & 158.94\% & 14.97\% & 19.96\% & 0.799 & -33.72\% & 0.15 \\
\end{longtable}
\endgroup

\FloatBarrier
\section{Numerical implementation and conditional-variance inference}\label{app:garch-details}

The model specification and calibration rule are in Section~\ref{sec:garch-design}, and the headline performance table and comparison figure are in Section~\ref{sec:garch-results}. This appendix gives extended fitting details and all paired contrasts on the one-session close-to-close target. These comparisons are interpreted within that exercise.

\subsection{Likelihood and fitting details}

The conditional-return model and forecast target are specified in Section~\ref{sec:garch-design}. Its coefficients are distinct from those in the main log-target regressions. For a standardized innovation value $u$, the density $f_{\nu}(u)$, with $\Gamma(\cdot)$ denoting the gamma function, is
\begin{equation}\label{eq:student-t-density}
f_{\nu}(u)=\frac{\Gamma((\nu+1)/2)}{\Gamma(\nu/2)\sqrt{\pi(\nu-2)}}\left(1+\frac{u^2}{\nu-2}\right)^{-(\nu+1)/2}.
\end{equation}
The conditional log likelihood $\ell_i$ sums over the return observations in the current estimation history:
\begin{equation}\label{eq:garch-likelihood}
\ell_i=\sum_t\left[\log f_{\nu_i}\!\left(\frac{\varepsilon_{i,t}}{\sqrt{q_{i,t}}}\right)-\frac{1}{2}\log q_{i,t}\right].
\end{equation}
Constrained maximum likelihood uses two deterministic starting values and retains the successful fit with the larger likelihood. The initial filtered variance is the estimation-sample variance. The rolling history contains the latest 1,008 returns; the expanding history begins in 2004 or at the instrument's first available observation thereafter. Re-estimation occurs every 22 sessions, with daily conditional-variance updates between refits.

The two starts use the sample-mean return $\mu$ and $\omega=\max\{10^{-8},\,s^2(1-P)\}$, with $s^2=\max\{10^{-8},\operatorname{Var}(r)\}$ and persistence $P=\alpha+\beta+\gamma/2$. For GARCH the starts $(P,\alpha,\beta,\nu)$ are $(.95,.06,.89,8)$ and $(.90,.10,.80,10)$; for GJR the starts $(P,\alpha,\beta,\gamma,\nu)$ are $(.95,.05,.87,.06,8)$ and $(.90,.10,.72,.16,10)$. The transformed box bounds are $[-2,2]$ for $\mu$, $[-18,5]$ for $\log\omega$, $[-7,10]$ for the persistence logit, $[-9,9]$ for each share logit, and $[-4.5,4.5]$ for $\log(\nu-2.01)$. SciPy L-BFGS-B uses 1,200 iterations, \texttt{ftol} $10^{-10}$ and 40 maximum line-search steps per iteration; unspecified tolerances use that version's defaults. At refit the variance recursion starts from $s^2$. Between refits the last issued one-step variance is the current variance state, updated with the newly available centered return and the fixed fitted parameters. The filtered variance lower bound is $10^{-12}$. The replication code implements these parameter transformations, candidate checks and state updates.

\reservetable{520pt}
\subsection{Paired history, asymmetry, and calibration tests}

Table~\ref{app:garch-contrasts} gives the full paired estimates and bandwidth-sensitive inference behind Section~\ref{sec:garch-results}.
\begingroup\papertablesize\setlength{\tabcolsep}{2pt}
\begin{longtable}{@{}L{.115\linewidth}L{.230\linewidth}R{.076\linewidth}rrrrrL{.165\linewidth}@{}}
\caption{Paired GARCH contrasts by family and HAC bandwidth}\label{app:garch-contrasts}\\
\toprule Family & Contrast & \shortstack[r]{Mean\\($\times10^{-2}$)} & \shortstack[r]{Raw\\HAC(4)} & \multicolumn{4}{c}{Holm-adjusted $p$-values by HAC lag} & \shortstack[l]{MBB-22 95\%\\interval ($\times10^{-2}$)}\\
\cmidrule(lr){5-8}
 & & & & 4 & 10 & 21 & 43 & \\ \midrule
\endfirsthead
\toprule Family & Contrast & \shortstack[r]{Mean\\($\times10^{-2}$)} & \shortstack[r]{Raw\\HAC(4)} & \multicolumn{4}{c}{Holm-adjusted $p$-values by HAC lag} & \shortstack[l]{MBB-22 95\%\\interval ($\times10^{-2}$)}\\
\cmidrule(lr){5-8}
 & & & & 4 & 10 & 21 & 43 & \\ \midrule
\endhead
\bottomrule\endlastfoot
History & GARCH, direct, expanding minus rolling & -0.9286 & 0.0116 & 0.0464 & 0.0567 & 0.0780 & 0.0625 & [-1.7231, -0.2022] \\
History & GARCH, ratio, expanding minus rolling & -0.9014 & 0.0193 & 0.0464 & 0.0599 & 0.0780 & 0.0625 & [-1.7042, -0.1742] \\
History & GJR, direct, expanding minus rolling & -1.0078 & 0.0291 & 0.0464 & 0.0599 & 0.0780 & 0.0625 & [-2.0226, -0.0924] \\
History & GJR, ratio, expanding minus rolling & -1.0081 & 0.0121 & 0.0464 & 0.0599 & 0.0780 & 0.0625 & [-1.9236, -0.2458] \\
Asymmetry & rolling, direct, GJR minus GARCH & -0.3452 & 0.6136 & 1.0000 & 1.0000 & 1.0000 & 1.0000 & [-1.7924, 1.2569] \\
Asymmetry & rolling, ratio, GJR minus GARCH & -0.3548 & 0.5709 & 1.0000 & 1.0000 & 1.0000 & 1.0000 & [-1.7502, 1.1739] \\
Asymmetry & expanding, direct, GJR minus GARCH & -0.4243 & 0.4930 & 1.0000 & 1.0000 & 1.0000 & 1.0000 & [-1.8143, 1.0153] \\
Asymmetry & expanding, ratio, GJR minus GARCH & -0.4616 & 0.4462 & 1.0000 & 1.0000 & 1.0000 & 1.0000 & [-1.7861, 0.9336] \\
Ratio & GARCH, rolling, ratio vs. direct & 0.5914 & 0.0408 & 0.0817 & 0.0809 & 0.0592 & 0.0561 & [0.1106, 1.1728] \\
Ratio & GARCH, expanding, ratio vs. direct & 0.6186 & 0.0003 & 0.0012 & 0.0024 & 0.0027 & 0.0031 & [0.2856, 1.0080] \\
Ratio & GJR, rolling, ratio vs. direct & 0.5817 & 0.0635 & 0.0817 & 0.0809 & 0.0597 & 0.0590 & [-0.0041, 1.1860] \\
Ratio & GJR, expanding, ratio vs. direct & 0.5814 & 0.0036 & 0.0109 & 0.0250 & 0.0435 & 0.0561 & [0.1771, 1.0871] \\
Interaction & direct, GJR vs. GARCH history change & -0.0792 & 0.7819 & 1.0000 & 1.0000 & 1.0000 & 1.0000 & [-0.8032, 0.4391] \\
Interaction & ratio, GJR vs. GARCH history change & -0.1067 & 0.6593 & 1.0000 & 1.0000 & 1.0000 & 1.0000 & [-0.6505, 0.3499] \\
\end{longtable}
\noindent\footnotesize Mean differences and interval bounds use the scale $10^{-2}$. Each date is weighted equally after pairing across 21 instruments. Holm families contain four history, four asymmetry, four ratio, and two history-by-asymmetry tests at each lag. Pointwise percentile intervals use 5,000 synchronized circular moving-block draws and are unadjusted. The replication materials include full-precision raw and adjusted values. \normalsize
\endgroup

\FloatBarrier

\FloatBarrier
\subsection{Numerical implementation and availability records}\label{app:numerical-record}

The reconstruction uses Python 3.12.14, NumPy 2.3.5, pandas 3.0.1 and XGBoost version 3.4.1; the original run manifests record package versions and input/code hashes. XGBoost uses the native training/QuantileDMatrix interface, CPU histogram trees, 256 bins and one numerical thread. Validation quantiles reference the inner-fitting matrix. Its row sampling is uniform, growth is depthwise, and column subsampling is by tree (0.7), with by-level and by-node rates one. The L2 leaf penalty is one; L1, minimum-split-loss and maximum-delta-step penalties are zero. The initial constant is automatically estimated from the training response for each fit. A first-month original XGBoost forecast was reconstructed in an author-run replay in the recorded environment before the targeted sensitivity; its largest forecast difference was below $10^{-15}$. The software configuration example and the full fit records accompany the source bundle.

OLS and DirectQLIKE standardize with the current training population mean and standard deviation. Standard deviations no larger than $10^{-12}$ are replaced by one. Their intercept-augmented SVD rank threshold is $\sigma_{\max}\max(n,p)\epsilon_{mach}$, where $n$ is the training-row count, $p$ the supplied-column count including the intercept, $\sigma_{\max}$ the largest singular value and $\epsilon_{mach}$ double-precision machine accuracy. The 19-column OLS-18 and DirectQLIKE-18 designs have rank 18; all HAR designs have rank four. The direct optimizer works on the identifiable right-singular-vector basis.

DirectQLIKE divides nonnegative training targets by their positive training mean and initializes the reduced coefficient vector at zero. This starts the original-scale forecast at that training mean; normalization changes the objective only by a constant. The damped Newton solver uses a least-squares Hessian solve, an Armijo coefficient $10^{-4}$ and at most 45 step halvings per iteration. The stopping gradient-infinity-norm tolerance is $10^{-7}$ and the maximum iteration count is 150. A final fit is accepted only when that norm is below $10^{-6}$ and the objective is no larger than its initialization objective plus $10^{-10}$. Invalid targets, non-descent directions, unsuccessful line searches or failed final checks cause an unavailable model-month forecast, not acceptance of an unconverged fit. No comparator or preceding fit is silently substituted.

The common primary issuance mask requires all 18 supplied features to be finite, including for HAR and XGBoost-3. Training additionally requires a finite target, origins from 2004 onward, and complete maturity by refit. Scoring requires a nonnegative finite observed target and positive finite compared forecasts. All 21 instruments satisfy the common requirements on all 1,715 primary dates. The principal numerical records contain 22,260 completed fits, 4,452 per method, and zero model-fit failures, fallback forecasts or nonfinite evaluation exclusions. There are no observed post-smearing floor violations; this does not change the implemented absence of a second floor. The historical exercise additionally retains unscorable issued origins when future target sessions are missing and excludes them from scoring rather than filtering securities by future survival.

\begingroup\papertablesize\setlength{\tabcolsep}{2pt}
\begin{longtable}{@{}L{.24\linewidth}L{.20\linewidth}R{.17\linewidth}R{.15\linewidth}R{.16\linewidth}@{}}
\caption{Completed fits, failures and fallback forecasts in final reconstruction outputs}\label{app:fit-census}\\
\toprule Exercise & Method & Accepted fits & \shortstack[r]{Failed final\\fits} & Fallbacks \\ \midrule
\endfirsthead
\multicolumn{5}{c}{Table \thetable{} -- continued}\\
\toprule Exercise & Method & Accepted fits & \shortstack[r]{Failed final\\fits} & Fallbacks \\ \midrule
\endhead
\midrule\multicolumn{5}{r}{Continued on next page}\\\endfoot
\bottomrule\endlastfoot
Yahoo GK 2020--2026 & DirectQLIKE-18 & 4,452 & 0 & 0 \\
Yahoo GK 2020--2026 & HAR & 4,452 & 0 & 0 \\
Yahoo GK 2020--2026 & OLS-18 & 4,452 & 0 & 0 \\
Yahoo GK 2020--2026 & XGBoost-18 & 4,452 & 0 & 0 \\
Yahoo GK 2020--2026 & XGBoost-3 & 4,452 & 0 & 0 \\
Yahoo GK 2014--2017 & DirectQLIKE-18 & 3,024 & 0 & 0 \\
Yahoo GK 2014--2017 & HAR & 3,024 & 0 & 0 \\
Yahoo GK 2014--2017 & OLS-18 & 3,024 & 0 & 0 \\
Yahoo GK 2014--2017 & XGBoost-18 & 3,024 & 0 & 0 \\
Yahoo GK 2014--2017 & XGBoost-3 & 3,024 & 0 & 0 \\
Restricted holdings GK & DirectQLIKE-18 & 39,302 & 0 & 0 \\
Restricted holdings GK & HAR & 39,302 & 0 & 0 \\
Restricted holdings GK & OLS-18 & 39,302 & 0 & 0 \\
Restricted holdings GK & XGBoost-18 & 39,302 & 0 & 0 \\
Restricted holdings GK & XGBoost-3 & 39,302 & 0 & 0 \\
21-instrument CC & GARCH & 4,242 & 0 & 0 \\
21-instrument CC & GJR & 4,242 & 0 & 0 \\
SPY CC & DirectQLIKE-18 & 214 & 0 & 0 \\
SPY CC & GARCH & 214 & 0 & 0 \\
SPY CC & GJR & 214 & 0 & 0 \\
SPY CC & HAR & 214 & 0 & 0 \\
SPY CC & OLS-18 & 214 & 0 & 0 \\
SPY CC & XGBoost-18 & 214 & 0 & 0 \\
SPY CC & XGBoost-3 & 214 & 0 & 0 \\
\end{longtable}
\noindent\footnotesize CC denotes the one-session close-to-close target. Counts include calibration warm-up and all histories. Zero final failures do not mean that every possible security/month had enough data: 68 restricted holdings jobs have no eligible forecasts, as recorded separately. These are non-issuance cases. Native fits and the six affected Check history jobs are counted separately; the 620 verified identical-input Check aliases are not refitted or counted a second time. The GARCH optimizer-start audit contains 8,484 starts per model, all converged; the independent saved-forecast check reports no unavailable evaluation forecasts.\normalsize
\endgroup

Table~\ref{app:fit-census} counts final accepted reconstruction fits separately from optimizer starting-value attempts and from insufficient-history non-issuance. The restricted holdings records contain 196,510 completed fits and zero failed final fits. For the all-instrument GARCH exercise, both deterministic optimizer starts converge for all saved model/history refits; the separate independent forecast verification also records no unavailable positive evaluation forecasts. The SPY one-session reconstruction has 214 completed fits per method. Earlier interrupted software attempts are retained separately and are not spliced into these final outputs. The bundle retains job-level statuses, exclusions, source hashes and independent verification records rather than relying on the aggregate manifest's status field alone.

\subsection{Reconstruction specifications and analysis chronology}\label{app:analysis-chronology}

The files dated 6 November 2026 are internal reconstruction specifications created during the retrospective revision, not a preregistered study or a study date planned in advance. Earlier data, manuscripts and analyses were already available. The available message history records the choice of daily lagged-roster forecasts, the subsequent selection of 2020--2026 after an initial longer Yahoo reconstruction had started, and the later addition of 2014--2017 as a secondary historical check. The original 2020--2026 and holdings run identities link the saved forecasts to their particular specification and code hashes. Those hashes establish what code was used; they do not establish that every design decision preceded examination of all relevant losses.

In this paper, \emph{principal} identifies the fixed reporting role of the ten documented contrasts under each loss, with their unchanged original Holm families. It does not mean prospectively confirmatory. The 2020, 2021--2022 and 2023--2026 splits were recorded in that reconstruction specification and are descriptive historical decompositions. The post-2020 omission comparisons and the present feature-redundancy and source-exclusion checks were added later and are explicitly exploratory. No part of the revised analysis is described as a genuinely untouched final evaluation. The primary sample, four-lag HAC tests and original contrast families have not been replaced after inspecting the new diagnostics.

The replication code bundle includes the original specification files, frozen hashes for source and configuration files, software version records, preparation and forecast-generation scripts, and exact command instructions. The original files document the analysis chronology. Reproducing the CRSP/WRDS exercise requires institutional entitlement and the identified licensed source archives. Hashes identify the permitted derived outputs and archived Yahoo inputs.

\subsection{Historical calibration chronology}\label{app:historical-calibration}
The Yahoo 2014--2017 exercise begins simulated forecasts on 3 January 2012; its first 126-error factor is available on 10 July 2012, and 498 matured errors are available at the first scored origin, 2 January 2014. The primary Yahoo and SPY close-to-close exercises begin on 2 January 2018; their first 126-error factors are available on 10 July 2018 (3 July for the SPY one-session application), with 499 matured errors for range forecasts and 503 for the SPY application at the first scored origin in January 2020. Before 126 matured errors, log-model smearing uses multiplier one; the GARCH/GJR ratio correction is unavailable until its 126th matured ratio. In the restricted holdings panel, per-security first-issue, first-factor, floor-availability and first-scoring dates vary; the machine-readable census reports those ranges and early-origin counts. No calibration error is admitted until its own forecast target is fully mature. These historical reconstructions are not prospective or preregistered evaluations.

\paragraph{Floor source and activation census}
The fixed scale becomes available on the date of the 252nd positive source observation. The public census gives the first and 252nd dates, the separately determined feature and issuance floors, training exposures and issued forecasts for each instrument and history. For the Yahoo GK exercises the floor availability dates range from 31 December 2002 to 10 March 2008; none of the training, calibration warm-up or scored targets falls below it. Across both histories, the SPY squared-return target falls below its floor in 2,144 training-label usages, four warm-up issuances and four scored issuances. Its outcome floor becomes available on 2 January 2003. Counts of training-label usages count appearances in each history's refits; calibration counts identify the pre-scoring issuances that furnish initial feedback. The raw forecast floor is not used in the objective. No stored log-model forecast, DirectQLIKE forecast or SPY conditional-variance forecast reaches the raw floor in those public issuance stages. GARCH for the 21-instrument exercise has a separate numerical variance bound and past-ratio correction; it does not use the log-model target floor. The restricted census groups floor dates by price view and history and records zero target and issuance floor activations. The separately licensed security-level records and raw inputs remain available to authorized users.

As a bounded SPY diagnostic, we recalculated the factors with the same maturity windows using $Y_s/f_s$. In the evaluation set the largest active factor difference was $4.81\times10^{-8}$ and the largest absolute change in average QLIKE was $1.12\times10^{-9}$. The reported results retain the original operational rule.

For the SPY one-session application, final training labels mature after one session. The saved boosting implementation uses the shared, conservative five-session purge for its inner validation. Captured \texttt{first\_train} metadata are present for Yahoo range fits and absent for the five SPY learners; the authors' audit reconstructs the SPY training window from its dated panel and validates its recorded sizes and last mature labels.

The authors ran a separate replay of AAPL on 15 July 2021 and USO on 21 April 2020 under both histories, covering OLS-18 and XGBoost-18 with active smearing and DirectQLIKE-18. All 12 forecast records reconciled. The largest absolute final-forecast difference was $8.18\times10^{-14}$; the phase-by-phase values and accepted DirectQLIKE line-search steps are included in the replication records. The replay regenerates targets, predictors, maturity and purge masks and refits each selected origin. Its active factors use the archived earlier issued forecasts and reconstructed maturity windows. A separate check evaluated four raw OLS forecasts and approximated four DirectQLIKE optima with a different optimizer. The final active factors and XGBoost replays are author-run checks.

\FloatBarrier
\begingroup\small\setlength{\bibsep}{1pt}

\endgroup
\end{document}